%% file: neurips_2023.tex
\documentclass[10pt]{article}
\PassOptionsToPackage{hyphens}{url}   

\usepackage{amsmath}
\usepackage{scicom}
\usepackage[authoryear]{natbib}

\usepackage{hyperref}
\usepackage{bookmark}
\usepackage{url}
\usepackage{nicefrac}
\usepackage{microtype}
\usepackage{tabularx}
\usepackage{multirow}
\usepackage{float}
\usepackage{needspace}
\usepackage{adjustbox}
\usepackage{longtable}
\usepackage{array}
\usepackage{placeins}

\newcolumntype{L}[1]{>{\raggedright\arraybackslash}p{#1}}
\input{numbers}

\hypersetup{
pdftitle={Low Language Transfer Speech-to-Text},
pdfauthor={Husein Zolkepli},
pdfsubject={Audio and Speech Processing},
pdfkeywords={speech recognition, transfer learning, low-resource languages, fine-tuning, multilingual models},
colorlinks=true,
linkcolor=scLink,
citecolor=scLink,
urlcolor=scLink
}
\title{Low Language Transfer Speech-to-Text}
\scshorttitle{Low Language Transfer Speech-to-Text}
\scauthor{Husein Zolkepli}{husein.zolkepli@scicom.com.my}
\scaffiliation{Scicom (MSC) Berhad, Malaysia}
\scdate{November 2026}

\begin{document}

\maketitle

\begin{abstract}
\input{abstract_body}
\end{abstract}

\input{body}

\FloatBarrier
\section{Acknowledgement}

This work was carried out at Scicom (MSC) Berhad. The author thanks Scicom for the compute.

\bibliographystyle{plainnat}
\bibliography{neurips_2023}

\appendix

\input{appendix}

\end{document}

%% file: abstract_body.tex
A multilingual speech recogniser has heard much speech in its own languages. We ask how much of that it brings to a language it was never trained on, when it is fine-tuned on the little that language has. We score \val{sv/scored} open multilingual models untouched on Low-Language-TTS, 50 low-resource languages in \val{data/scripts} writing systems, and fine-tune \val{sv/tuned} of them on at most 1,000 clips a language, \val{data/k1000-hours} hours in all, with a dev set choosing every epoch. Fine-tuning lowers every model's macro character error rate (CER), by \val{ft/gain-min} to \val{ft/gain-max} points, to \val{ft/top/cer} at best, and a word language model takes three CTC models \val{lm/cer-gain-min} to \val{lm/cer-gain-max} points lower; WER is still \val{lm/best/wer} at best. On the \val{nl/n} languages that no model lists, every model starts at \val{nl/before-min} or worse and the best ends at \val{nl/best/cer}. The same architectures from random weights stay \val{scr/gap-min} points or more behind, so the gain is transfer, and an encoder left untouched already gives a Whisper or Qwen decoder most of what it needs. It is the language, not the corpus: on FLEURS recordings that ours do not share, the median model keeps \val{ooc/keep-median}\% of its gain. It costs the languages the models knew, \val{fk/loss-median} points of CER on English, Spanish, French and German in the median model. Low-rank adapters on \val{ad/lora/share-max}\% of the weights come within \val{ad/lora/cost-max} points of tuning everything, and on \val{fk/lora/less-n} of \val{fk/lora/n} models forget less. Seq2seq and language-model decoders are done within three to five epochs, while a new CTC head is still improving at twenty or thirty. Data beats epochs: 25 clips a language take Qwen3-ASR 0.6B from \val{bl/qwen3-asr-0.6b/cer} to \val{n/qwen3-asr-0.6b/25}, and at equal clips presented, ten times the data is worth \val{n/qwen3-asr-0.6b/same-gain} points. A model that first learned forty of the languages learns the other ten no faster. On a public Malay benchmark, untouched Whisper large-v3 scores \val{my/whisper-large-v3/left} when it detects the language and \val{my/whisper-large-v3/told} when told the clip is Malay, because it then translates the English that speakers mix in; trained on the Scicom telephony dataset it scores \val{my/whisper-large-v3/after}.

%% file: body.tex
\section{Introduction}

\label{sec:intro}

Speech recognisers are trained on the languages that have data, and most languages do not. A multilingual pretrained model has heard hundreds of thousands of hours of speech in its own languages. This paper asks how much of that it brings to a language it was never told about, when someone fine-tunes it on the little that language has.

We take \val{sv/n} open multilingual models and score the \val{sv/scored} that transcribe as downloaded, untouched, on Low-Language-TTS \citep{zolkepli2026lowlang}, 50 of the lowest-resource languages of a large public corpus. We then fine-tune \val{sv/tuned} of them on the same training set, at most 1,000 clips a language, and score them again. Four controls say what the gain is made of: the same architectures from random weights, each half of a model frozen in turn, each language trained alone, and the \val{nl/n} languages that no model lists, where none starts ahead (Section~\ref{sec:nolist}). Three questions from practice get their own sections: whether more epochs help (Section~\ref{sec:epochs}), how few clips are enough (Section~\ref{sec:scale}), and whether Common Voice would have done (Section~\ref{sec:cv}). Then we test the result on another corpus and on the languages the models already knew (Section~\ref{sec:ooc}), ask whether a model that has learned forty of the languages learns the next one faster (Section~\ref{sec:next}), and compare small adapters and a word language model with tuning everything (Section~\ref{sec:peft}). A last section leaves the long tail for Malay, where the same models train on telephone calls alone and are scored on speech of every other kind (Section~\ref{sec:malay}).

\section{Background}

\label{sec:background}

\textbf{Transfer in speech recognition.} Adapting a network trained on one language to another is old and cheap: a small change to an English model's weights was enough for German \citep{kunze2017transfer}. Training on many languages first helps more \citep{dalmia2018sequence}, and how similar the target is to the pretraining languages matters along several dimensions at once \citep{adams2019massively}. Meta-learning the starting point has also been tried \citep{hsu2020meta}.

\textbf{Multilingual models.} XLS-R learned speech representations from 128 languages without labels \citep{babu2021xlsr}. MMS scaled this to 1,107 languages, on readings of religious texts \citep{pratap2024scaling}, and Omnilingual ASR to over 1,600 \citep{omnilingual2025}. Whisper \citep{radford2023robust} lists 99 languages. Recent models list fewer and transcribe them better: Parakeet TDT \citep{xu2023efficient,rekesh2023fast,nvidia2025parakeetv3} and Canary \citep{nvidia2025canary} 25, Nemotron 3.5 40 \citep{nvidia2026nemotronstream,nvidia2026nemotrontranscribe}, Qwen3-ASR 30 \citep{qwen3asr2026}, Voxtral 8 \citep{voxtral2025}, Granite Speech 6 \citep{granite2025,ibm2026granite}, Cohere Transcribe 14 \citep{mack2026coheretranscribe}. Several put a language model behind the audio encoder \citep{chen2023salm}. General assistants now transcribe when asked: Gemma 4 takes audio in three of its sizes \citep{gemmateam2026gemma4}, and LFM3-Audio lists 18 languages for speech \citep{liquidAI2026LFM3}.

\textbf{Fine-tuning on one language at a time.} Most low-resource work tunes one model on one language or a few: Whisper on Amharic \citep{amharic2025}, Northern Kurdish \citep{kurdish2024} and Malasar \citep{multistage2024}, with data augmentation for Gronings and West Frisian \citep{bartelds2023making}. Studies of data size find most of the gain early, on African languages \citep{kinyarwanda2025,african2025benchmark}, and pretrained encoders differ in how well they hold up with little data \citep{yang2023investigating}. The nearest comparison is the benchmark of Whisper, XLS-R, MMS and w2v-BERT on 13 African languages \citep{african2025benchmark}.

\textbf{Adapters, forgetting and decoding.} A network trained on something new loses some of what it knew \citep{kirkpatrick2016overcoming}. Low-rank adapters train a few added weights and leave the rest as it was \citep{hu2021lora}, MMS adds each language through a small adapter of its own \citep{pratap2024scaling}, and adapters have been used to keep a speech recogniser from forgetting \citep{vandereeckt2022adapters}. A CTC model's output can also be searched with a word language model \citep{heafield2011kenlm,kensho2021pyctcdecode}.

\textbf{What this paper adds.} One held-out test set of 50 languages in \val{data/scripts} writing systems. Every family of current decoders on the same training data, including models whose vocabulary cannot spell the language at all. Random-weight controls that measure the transfer itself rather than the final error. A check of the gain on another corpus, of what it costs the languages the models knew, and of what adapters and a word language model change.

\needspace{8\baselineskip}
\section{Data}

\label{sec:data}

\subsection{The test set}

\begin{figure}[H]
\centering
\includegraphics[width=0.78\linewidth]{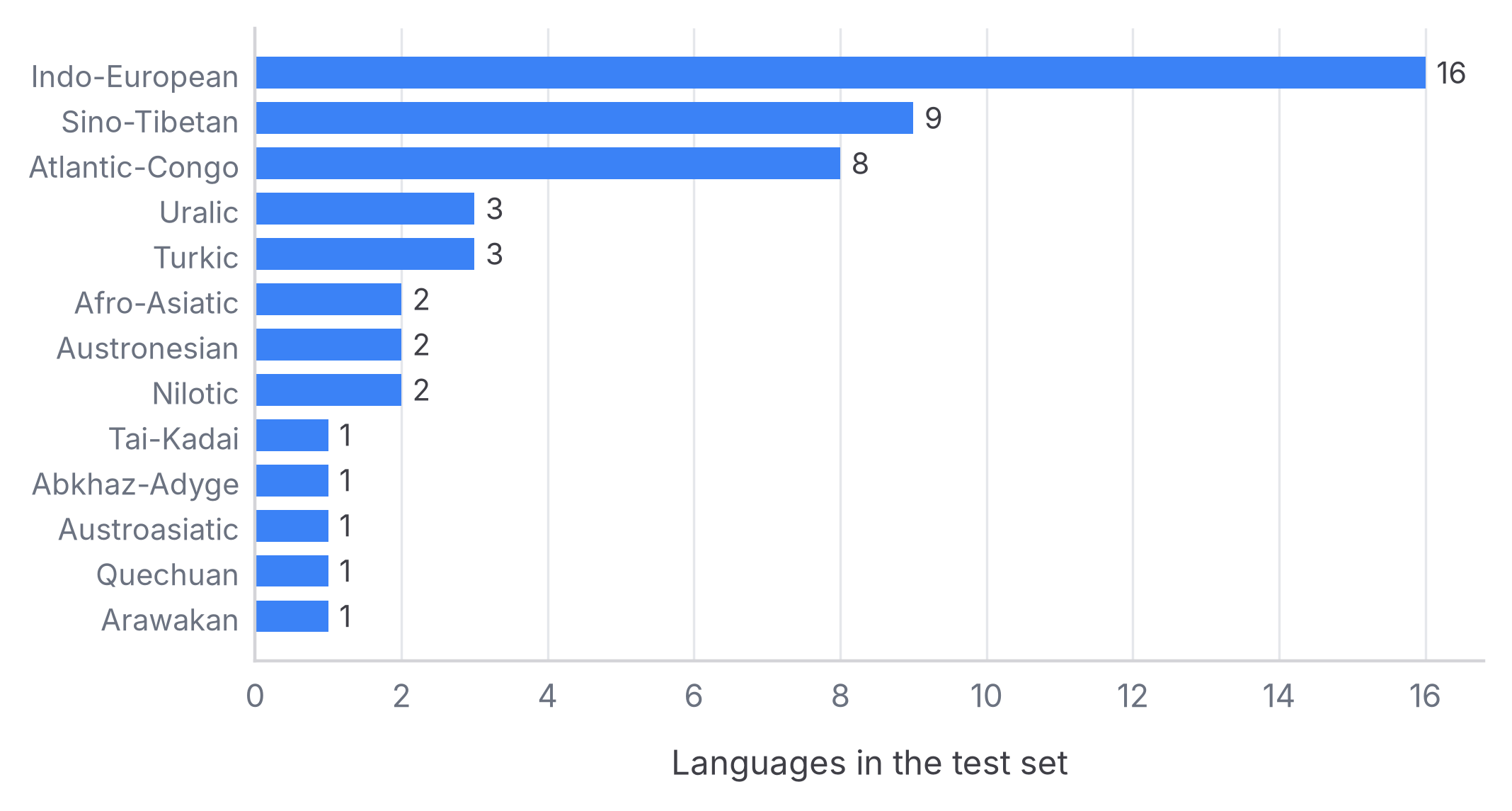}
\caption{The test set's languages by language family.}
\label{fig:families}
\end{figure}

\begin{figure}[H]
\centering
\includegraphics[width=0.78\linewidth]{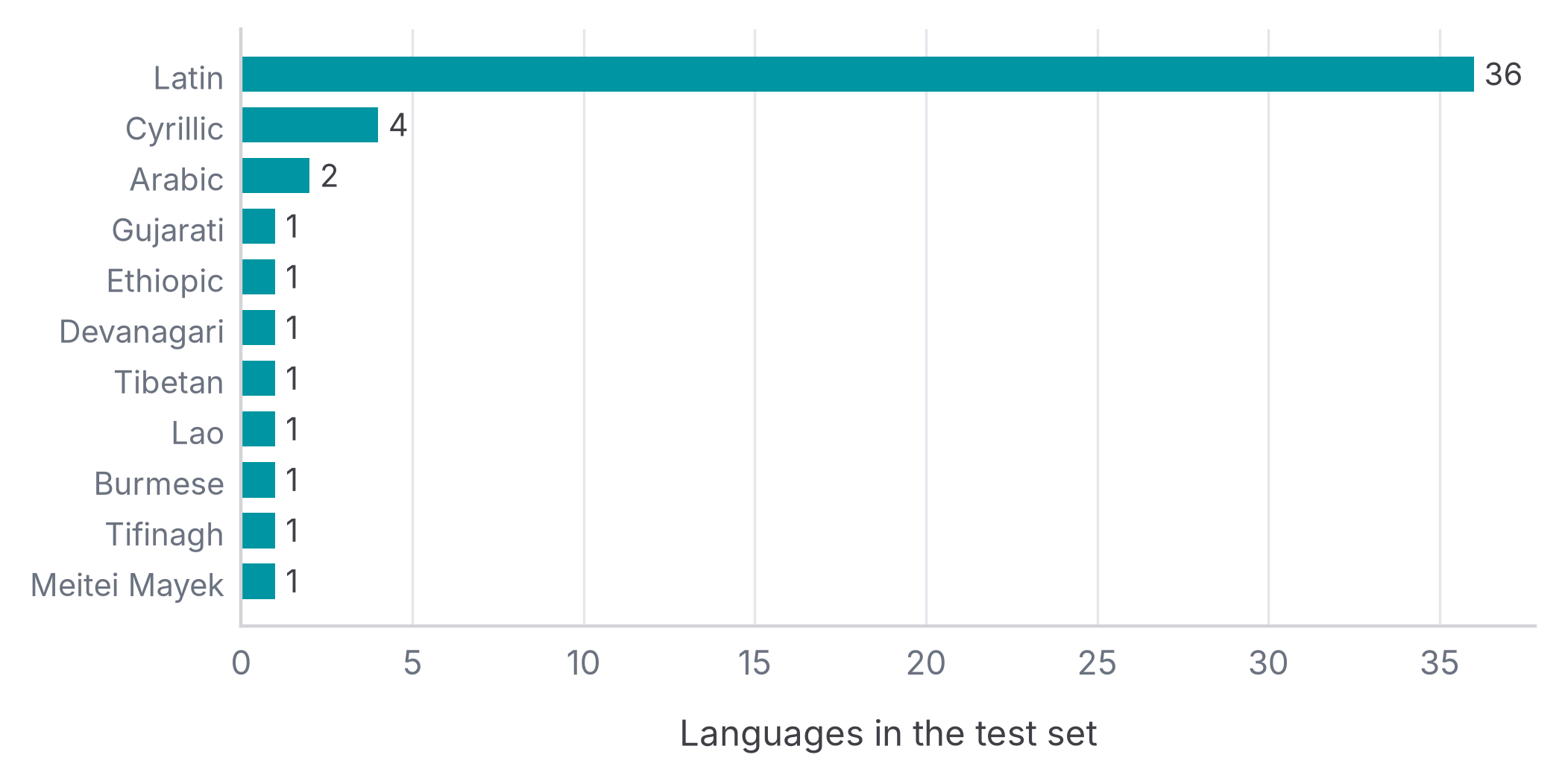}
\caption{The test set's languages by writing system.}
\label{fig:scripts}
\end{figure}

We score on Low-Language-TTS \citep{zolkepli2026lowlang}: \val{data/test-clips} clips and \val{data/test-hours} hours from \val{data/test-speakers} speakers, 25 clips for each of 50 of the lowest-resource languages of a 121.8M-row corpus \citep{zolkepli2026mtts}. A language counts only if some subset of the corpus was collected for it, because the rare end of a language detector's histogram is noise \citep{kargaran2023glotlid}. Figures~\ref{fig:families} and~\ref{fig:scripts} show what that leaves: \val{data/families} language families, and \val{data/nonlatin} languages outside the Latin script. The study never chooses anything on this set. Some of the corpus's subsets repackage FLEURS \citep{conneau2022fleurs}: \val{data/fleurs-test-clips} test clips in \val{data/fleurs-test-langs} languages are FLEURS recordings, among them every test clip of \val{data/fleurs-whole-names}, and \val{data/fleurs-test-from-train} of them come from FLEURS's training split, so a model trained on it has heard them (Section~\ref{sec:ooc}).

\subsection{The training set}

\begin{figure}[H]
\centering
\includegraphics[width=\linewidth]{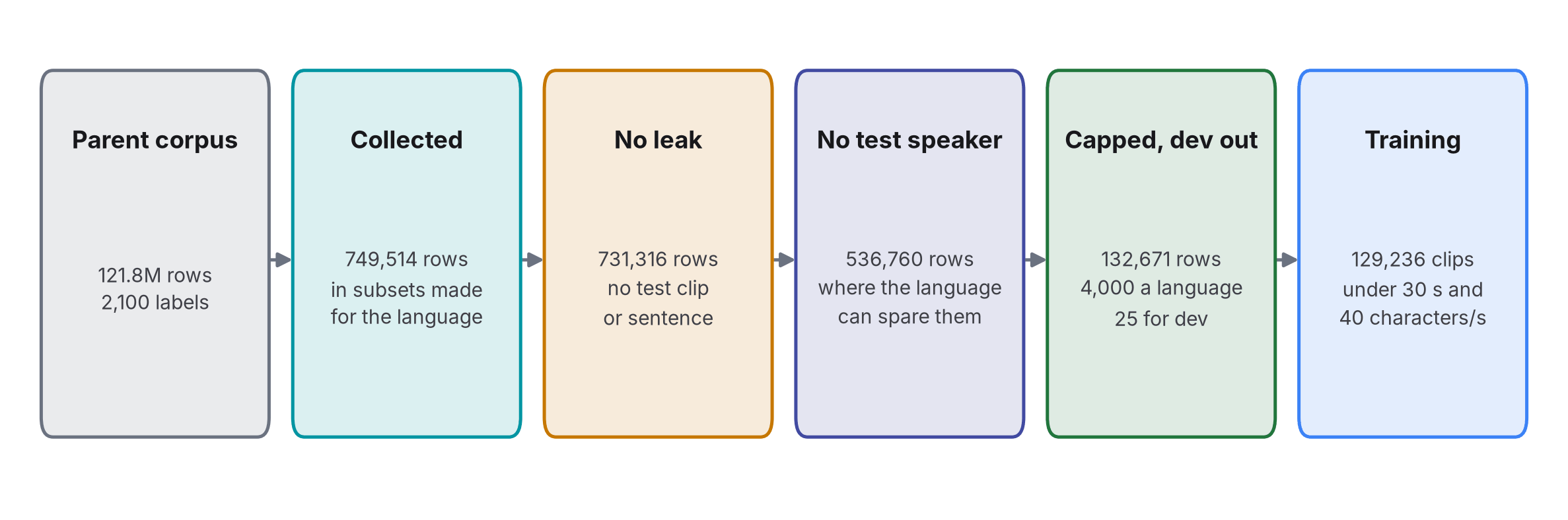}
\caption{From the parent corpus to the training set. A test or dev clip's text leaves training in every subset, since Bible corpora and augmented copies repeat their sentences. No test clip, test sentence or identical decoded audio remains in training or dev; a script checks all three.}
\label{fig:split}
\end{figure}

Common Voice 22 \citep{ardila2020common,mozilla2024cv} was the first candidate, and it does not cover the study: of its \val{cv/locales} locales only \val{cv/matched} are among our 50 languages, so \val{cv/absent} have nothing there. Training data therefore comes from the corpus the test set was cut from, under the rules of Figure~\ref{fig:split}; Section~\ref{sec:cv} adds Common Voice as a comparison.

\begin{figure}[H]
\centering
\includegraphics[width=0.92\linewidth]{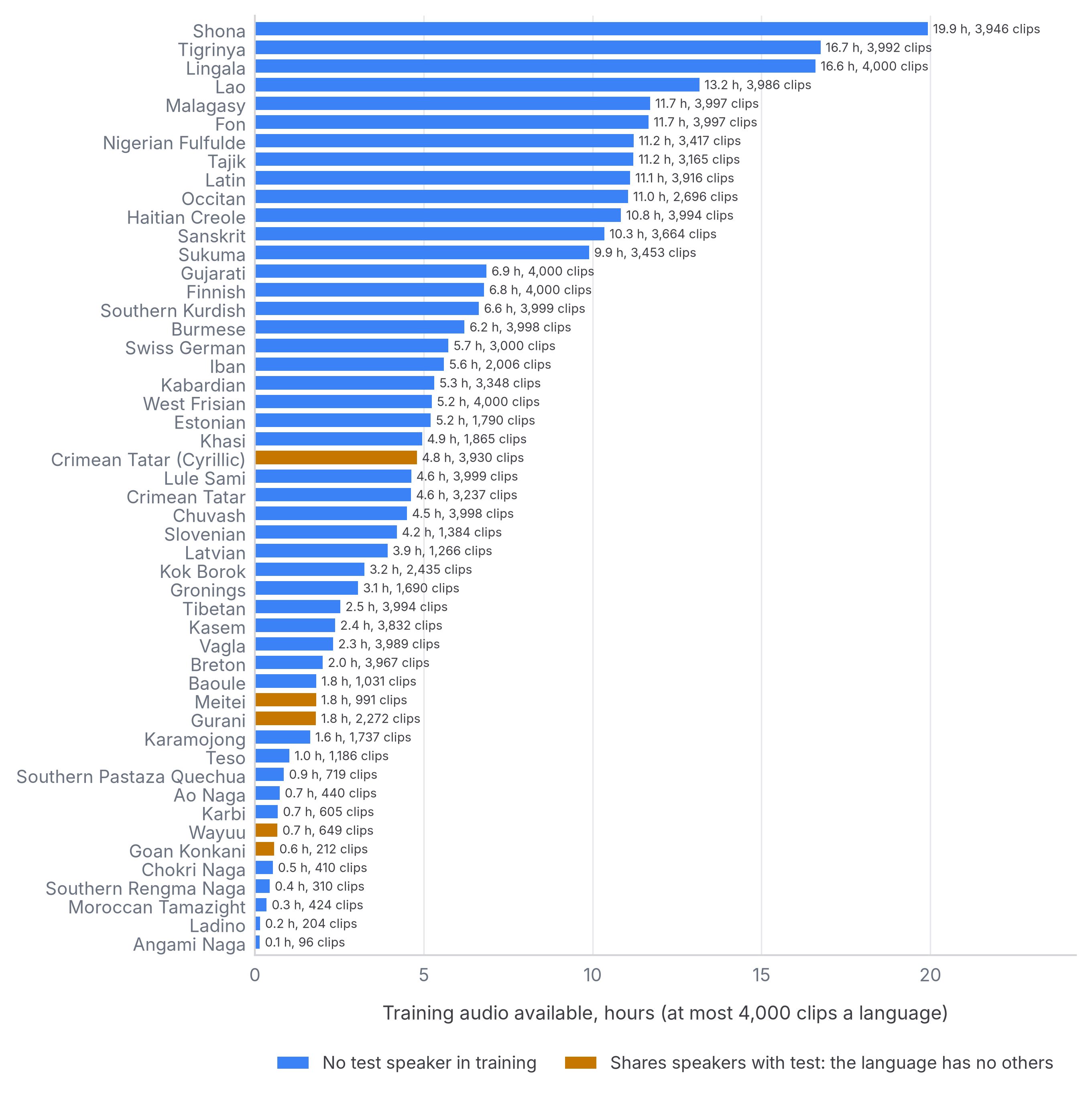}
\caption{Training audio per language, after every rule of Figure~\ref{fig:split}.}
\label{fig:languages}
\end{figure}

That leaves \val{data/train-clips} clips and \val{data/train-hours} hours from \val{data/train-speakers} speakers (Figure~\ref{fig:languages}). In \val{data/disjoint} languages no test speaker is in training. The other \val{data/shared} (\val{data/shared-names}) have no other speakers to train on. Dev has \val{data/dev-clips} clips, 25 a language, gated on audio length as the test set was: some corpora pair a long sentence with a second of audio. It takes each language's smallest speakers first, because they cost training least, which makes it harder than test (Section~\ref{sec:limits}). Within a language, speakers take turns in the order of the clips, so a smaller training set is always a prefix of a larger one. The main runs use at most 1,000 clips a language: \val{data/k1000-clips} clips and \val{data/k1000-hours} hours in all. Fetching the test clips again through the code that fetched the training audio returns the same samples for all \val{data/fetch-n}, and a training clip's audio and its speech tokens, which the parent corpus keeps in separate archives, agree on length to within 20\% on all but \val{data/pairs-off} of the \val{data/pairs-n} clips that have tokens.

\subsection{The metric}

\label{sec:metric}

The headline is the macro character error rate: the CER of each language over its 25 clips, then the mean over languages, so that each language counts once. Text is normalised the same way for every model (NFKC, lower case, punctuation and symbols removed, combining marks kept). Characters are the fair unit because \val{data/scripts} writing systems do not agree on what a word is; WER is reported beside it.

We cap each clip's errors at its own length. A decoder stuck in a loop then counts as a complete failure of that clip, not as twenty failures. Without the cap, a few loops moved a language by tens of points from one epoch to the next and hid everything else (Appendix~\ref{app:baseline}). Intervals resample speakers within each language, because one speaker's clips are not independent; they take the 50 languages as given, and Section~\ref{sec:order} draws the languages again.

\section{Models}

\label{sec:models}

The study's one criterion for a model is that it is multilingual by its own card. We read the \val{sv/top} most-downloaded speech recognition repositories on Hugging Face on the day of the study, removed conversions, quantisations, third-party fine-tunes and single-language models, and kept what runs in current Transformers or its own open package (Table~\ref{tab:models}). Between them they cover the families of decoder in use today. Two general assistants are not on that list, because the Hub files them under other tasks, and were added for comparison: Gemma 4, in the three sizes that take audio, and LFM3-Audio.

\begin{table}[H]
\centering
\caption{The models. Languages listed is what each card says it transcribes; w2v-BERT 2.0 is a self-supervised encoder pretrained on that many languages, without a transcription head. MMS as downloaded swaps in an adapter and a head for the language it is told; fine-tuned here, it has one head for all 50.\ifval{bl/gemma/cer-span}{ Gemma 4's parameters are the stored ones; its card counts E2B and E4B as 2.3B and 4.5B effective, without the token tables each layer looks up.}{}\late{none-pending}{}{ \val{sv/pending-names} joined the study last.}}
\label{tab:models}
\scriptsize
\input{table_models}
\end{table}

\begin{figure}[H]
\centering
\includegraphics[width=0.84\linewidth]{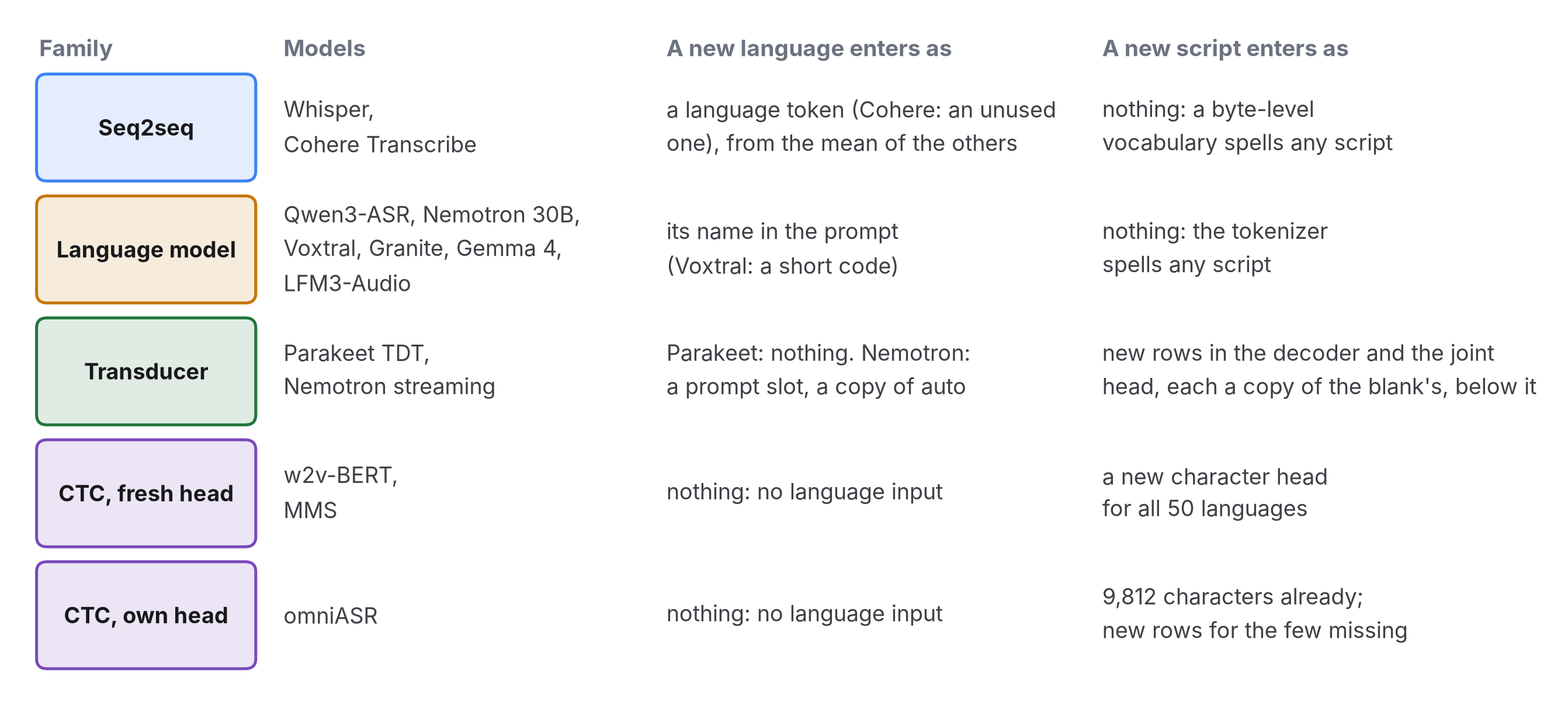}
\caption{What each family needs before it can fine-tune on a language it does not know. Every pretrained weight is kept in every family, except MMS's CTC heads, which are replaced; w2v-BERT comes without one.}
\label{fig:recipe}
\end{figure}

A model cannot fine-tune on a language until it can accept the language and spell it, and each family needs something different (Figure~\ref{fig:recipe}). Models with a language model or a byte-level vocabulary can write any script already. The transducers cannot: \val{data/uncovered/parakeet-tdt-v3} of our languages are mostly outside Parakeet's vocabulary and \val{data/uncovered/nemotron-stream} outside Nemotron's. We add a row per missing character and keep every pretrained weight \citep{nvidia2026finetune}. MMS as downloaded is different again: a small adapter in every layer and a head for each of its languages, swapped in for the language the user names. We load its English adapter once, replace the heads with one head for all 50 languages and train every weight, so the fine-tuned MMS, like the other CTC models, has one head and is never told the language.

A new row has to start silent, so that the untrained extension is still the pretrained model, and that took two attempts. Zero weights and a bias below every pretrained row is the obvious start, and it is not silent: this head's logits sink far below its biases, so the constant wins most frames. Untrained, Parakeet repeated its first new character on \val{sr/parakeet-tdt-v3/stuck} of the \val{sr/n} test clips and Nemotron on \val{sr/nemotron-stream/stuck}. A row that starts as a copy of the blank's, ten below it, cannot win a frame, because it is below the blank whatever the frame holds. With that start the untrained model's output is the pretrained model's on \val{sr/parakeet-tdt-v3/identical} of those clips for Parakeet and \val{sr/nemotron-stream/identical} for Nemotron. Section~\ref{sec:start} follows both starts through training.

\section{Untouched}

\label{sec:untouched}

As downloaded, a language on a model's own list is forced; any other is left to the model, as a user would have to, since it cannot be told about a language it was never trained on (Canary and Cohere Transcribe have no detection, so they get English).\ifval{bl/gemma/cer-span}{ Gemma 4 is the exception: its card lists no languages for speech and the prompt it gives for transcription takes any language's name, so it is told every language.}{}

Models that list up to a hundred languages sit at a macro CER of \val{bl/whisper-large-v3/cer} to \val{bl/vibevoice/cer} (Figure~\ref{fig:baseline}): they write something, often in another script (Section~\ref{sec:transfer}).\ifval{bl/gemma/cer-span}{ Gemma 4's \val{bl/gemma/sizes} with an audio encoder are lower, at \val{bl/gemma/cer-span}\ifval{bl/gemma/alone-span}{, and the name is part of that: with no language in the prompt they score \val{bl/gemma/alone-span}, \val{bl/gemma/told-gain-span} points worse}{}.}{}\ifval{bl/gemma4-12b/alone}{ Its 12B, which reads the waveform with no encoder, is at \val{bl/gemma4-12b/cer}, and at \val{bl/gemma4-12b/alone} with no language in the prompt: on English, Spanish, French and German it scores \val{bl/gemma4-12b/known}, but a language it has not heard it writes as one it has, Finnish as Arabic or Spanish, Gujarati in Hindi's script.}{} The two systems built for the long tail are far ahead: omniASR at \val{bl/omni-llm-1b/cer} to \val{bl/omni-ctc-300m/cer} on all 50, and MMS at \val{bl/mms-1b-all/cer} on the \val{bl/mms-1b-all/n} languages it has a head for. On the rest, MMS cannot detect a language any more than Canary can, so it gets English's adapter and head, and over all 50 it scores \val{bl/mms-1b/cer}. MMS was trained on readings of religious texts \citep{pratap2024scaling}, and some of our languages' clips are the same kind of reading. Appendix~\ref{app:baseline} shows why the cap is there and has every score with and without it.

\begin{figure}[H]
\centering
\includegraphics[width=0.95\linewidth]{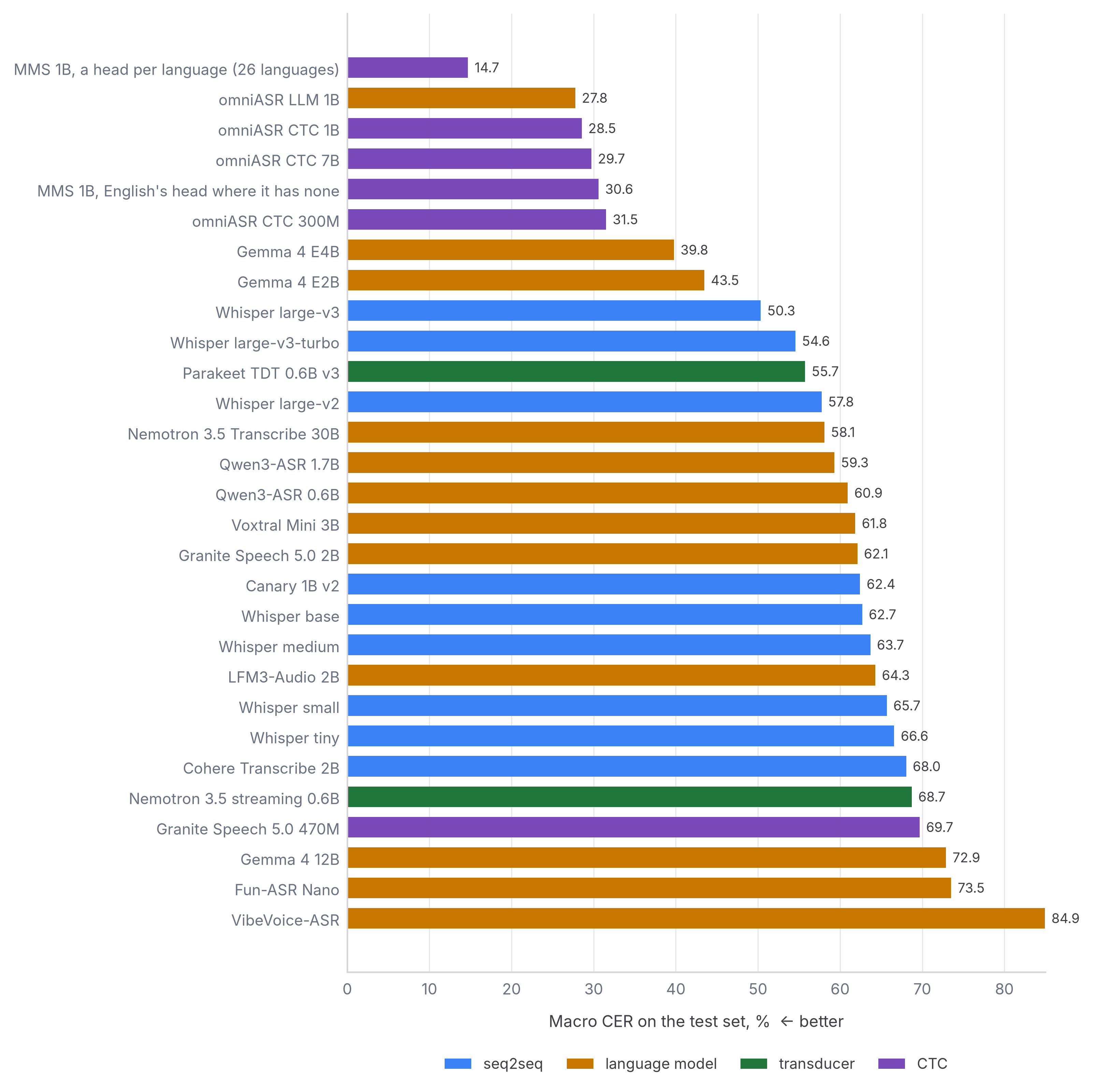}
\caption{Every model as it was downloaded, before any fine-tuning, on the test set. Mean over 50 languages unless stated; the fine-tuned scores are in Section~\ref{sec:before-after}.\ifval{bl/gemma/cer-span}{ Gemma 4 is told every language's name, the others only the languages they list.}{}}
\label{fig:baseline}
\end{figure}

\input{results_sections}

%% file: table_models.tex
\begin{adjustbox}{max width=\linewidth}
\begin{tabular}{llrlrl}
\toprule
\schead \scth{Model} & \scth{Decoder} & \scth{Parameters} & \scth{Languages listed} & \scth{Fine-tuned here} & \scth{Licence} \\
\midrule
Whisper large-v3 & seq2seq & 1.54B & 99 & yes & Apache-2.0 \\
Whisper large-v3-turbo & seq2seq & 0.81B & 99 & yes & MIT \\
Whisper large-v2 & seq2seq & 1.54B & 99 & yes & Apache-2.0 \\
Whisper medium & seq2seq & 0.76B & 99 & yes & Apache-2.0 \\
Whisper small & seq2seq & 0.24B & 99 & yes & Apache-2.0 \\
Whisper base & seq2seq & 0.07B & 99 & yes & Apache-2.0 \\
Whisper tiny & seq2seq & 0.04B & 99 & yes & Apache-2.0 \\
Qwen3-ASR 1.7B & language model & 2.04B & 30 & yes & Apache-2.0 \\
Qwen3-ASR 0.6B & language model & 0.78B & 30 & yes & Apache-2.0 \\
Nemotron 3.5 Transcribe 30B & language model & 32.36B & 40 & yes & OpenMDW-1.1 \\
Parakeet TDT 0.6B v3 & transducer & 0.63B & 25 & yes & CC-BY-4.0 \\
Nemotron 3.5 streaming 0.6B & transducer & 0.64B & 40 & yes & OpenMDW-1.1 \\
Canary 1B v2 & seq2seq & 0.96B & 25 & no & CC-BY-4.0 \\
w2v-BERT 2.0 & CTC & 0.58B & 143 & yes & MIT \\
MMS 1B, a head per language & CTC & 0.96B & 1,198 & yes & CC-BY-NC-4.0 \\
omniASR CTC 300M & CTC & 0.33B & 1,600+ & yes & Apache-2.0 \\
omniASR CTC 1B & CTC & 0.98B & 1,600+ & yes & Apache-2.0 \\
omniASR CTC 7B & CTC & 6.50B & 1,600+ & yes & Apache-2.0 \\
omniASR LLM 1B & language model & 2.28B & 1,600+ & no & Apache-2.0 \\
Voxtral Mini 3B & language model & 4.68B & 8 & yes & Apache-2.0 \\
Granite Speech 5.0 2B & language model & 2.31B & 6 & yes & Apache-2.0 \\
Granite Speech 5.0 470M & CTC & 0.47B & 7 & no & Apache-2.0 \\
Fun-ASR Nano & language model & 0.83B & 4 & no & Apache-2.0 \\
VibeVoice-ASR & language model & 8.33B & 50+ & no & MIT \\
Cohere Transcribe 2B & seq2seq & 2.07B & 14 & yes & Apache-2.0 \\
LFM3-Audio 2B & language model & 1.98B & 18 & yes & LFM Open v1.0 \\
Gemma 4 E2B & language model & 5.10B & 35+ & yes & Apache-2.0 \\
Gemma 4 E4B & language model & 7.94B & 35+ & yes & Apache-2.0 \\
Gemma 4 12B & language model & 11.96B & 35+ & yes & Apache-2.0 \\
\bottomrule
\end{tabular}
\end{adjustbox}

%% file: results_sections.tex
\section{Fine-tuning}

\label{sec:finetune}

\subsection{The recipe}

\begin{figure}[H]
\centering
\includegraphics[width=0.89\linewidth]{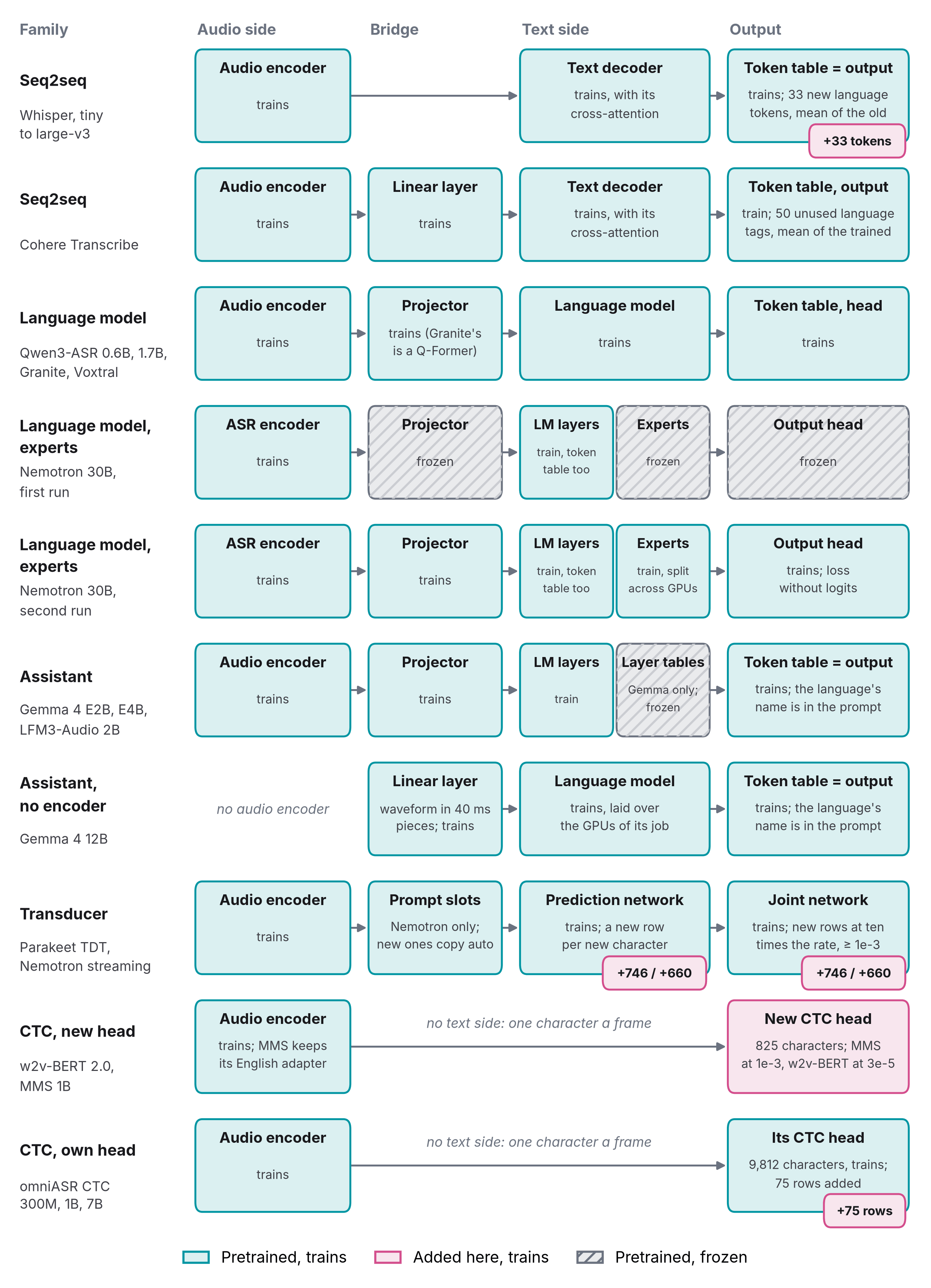}
\caption{What the recipe does to each part of each family. Nemotron 30B ran twice: first with its experts, projector and output head frozen, then with every part training. A pink badge is what fine-tuning adds to a pretrained part; on the transducers it is Parakeet's count of new rows, then Nemotron streaming's.}
\label{fig:arch}
\end{figure}

Every model trains the same way: AdamW with weight decay 0.01, its pretrained weights in 32-bit floats under bf16 autocast, a learning rate chosen for it on dev (Appendix~\ref{app:lr}), a warm-up of 200 steps and then a constant rate, gradient clipping at 1, and greedy decoding. All 50 languages train together on at most 1,000 clips each, \val{data/k1000-hours} hours in all, for ten epochs, except \val{rec/fewer-epochs}. After every epoch we score dev and test; dev alone chooses the epoch. The targets are the corpus's transcripts with their case and punctuation, except for the CTC models below. Every pretrained weight a transcript runs is kept and trains, with three exceptions: Nemotron 30B's first run freezes most of the model, Gemma 4's token tables for each layer stay frozen, and MMS's heads are replaced. Figure~\ref{fig:arch} shows what happens to each part of each family, and the paragraphs below go through them.

\textbf{Whisper.} Encoder and decoder train together, cross-attention included, and nothing is frozen. Its byte-level vocabulary already spells every script, so the one addition is a language token for each of the \val{rec/whisper/new} languages Whisper has none for, started at the mean of its language tokens; the other \val{rec/whisper/own} keep their own. The token goes where Whisper expects one, after the start-of-transcript token, and the loss counts only the transcript and the end token. Whisper's output layer is its token table, so a new token is also a new output.

\textbf{Cohere Transcribe.} The other seq2seq model is a Conformer encoder of \val{rec/cohere/enc}B parameters under a small Transformer decoder, trained from scratch on 14 languages, none of them ours \citep{mack2026coheretranscribe}. It trains whole, and its SentencePiece vocabulary falls back to bytes, so it spells every script and nothing is added. Its prompt names the language with a tag, and its table holds tags that were never trained: \val{rec/cohere/tags} of our languages take the tag of their two-letter code and the other \val{rec/cohere/spare} a spare token, each started at the mean of the 14 trained tags, as a new Whisper token is. Its output layer is not tied to the token table, so the two train side by side.

\begin{figure}[H]
\centering
\includegraphics[width=0.95\linewidth]{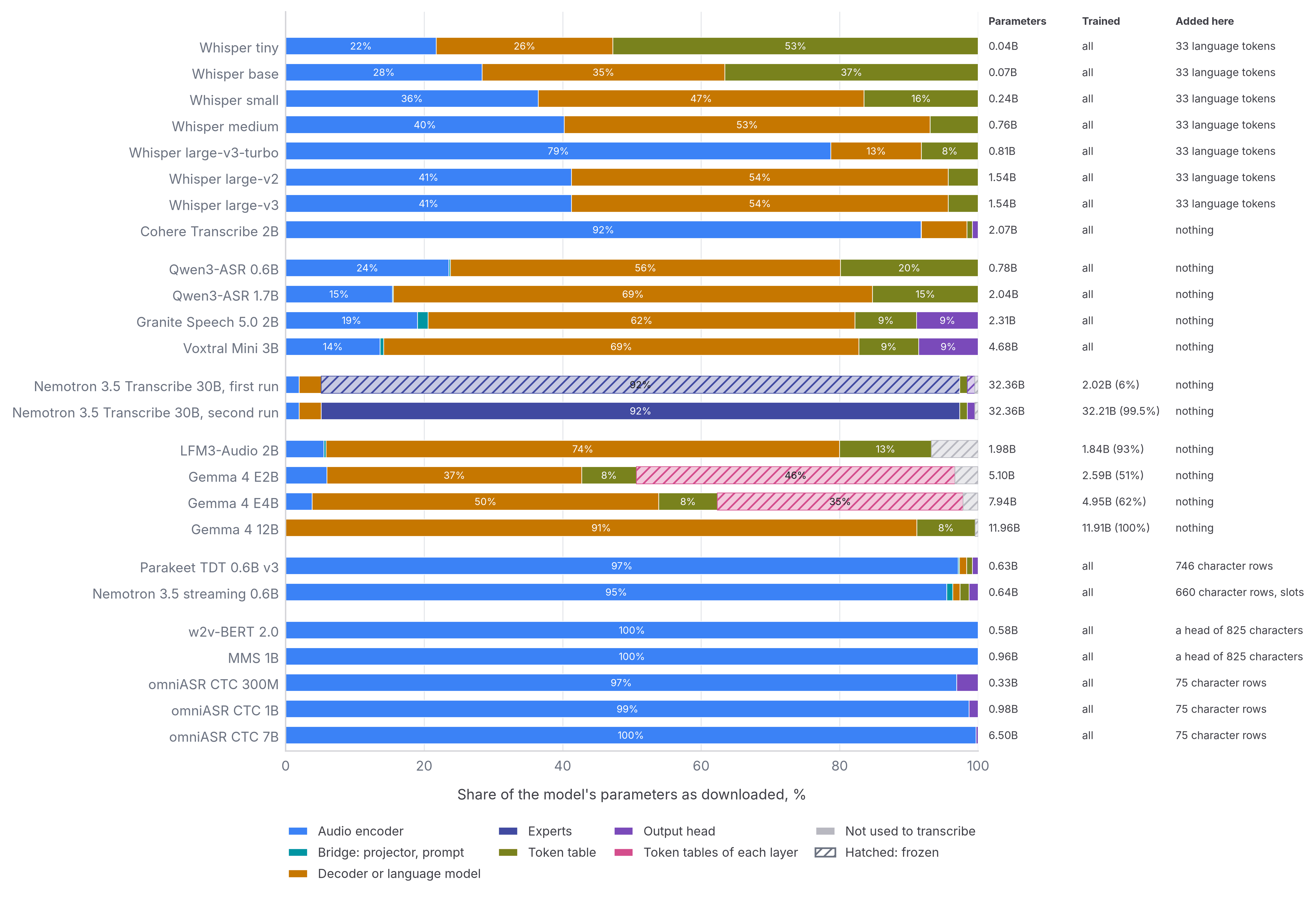}
\caption{Every fine-tuned model's parameters by part, as a share of the model as downloaded; hatched parts stay frozen. Nemotron 30B's second run trains every part but those transcription never runs. Trained is what fine-tuning updates; added here is what it adds.}
\label{fig:parts}
\end{figure}

\begin{figure}[H]
\centering
\includegraphics[width=0.88\linewidth]{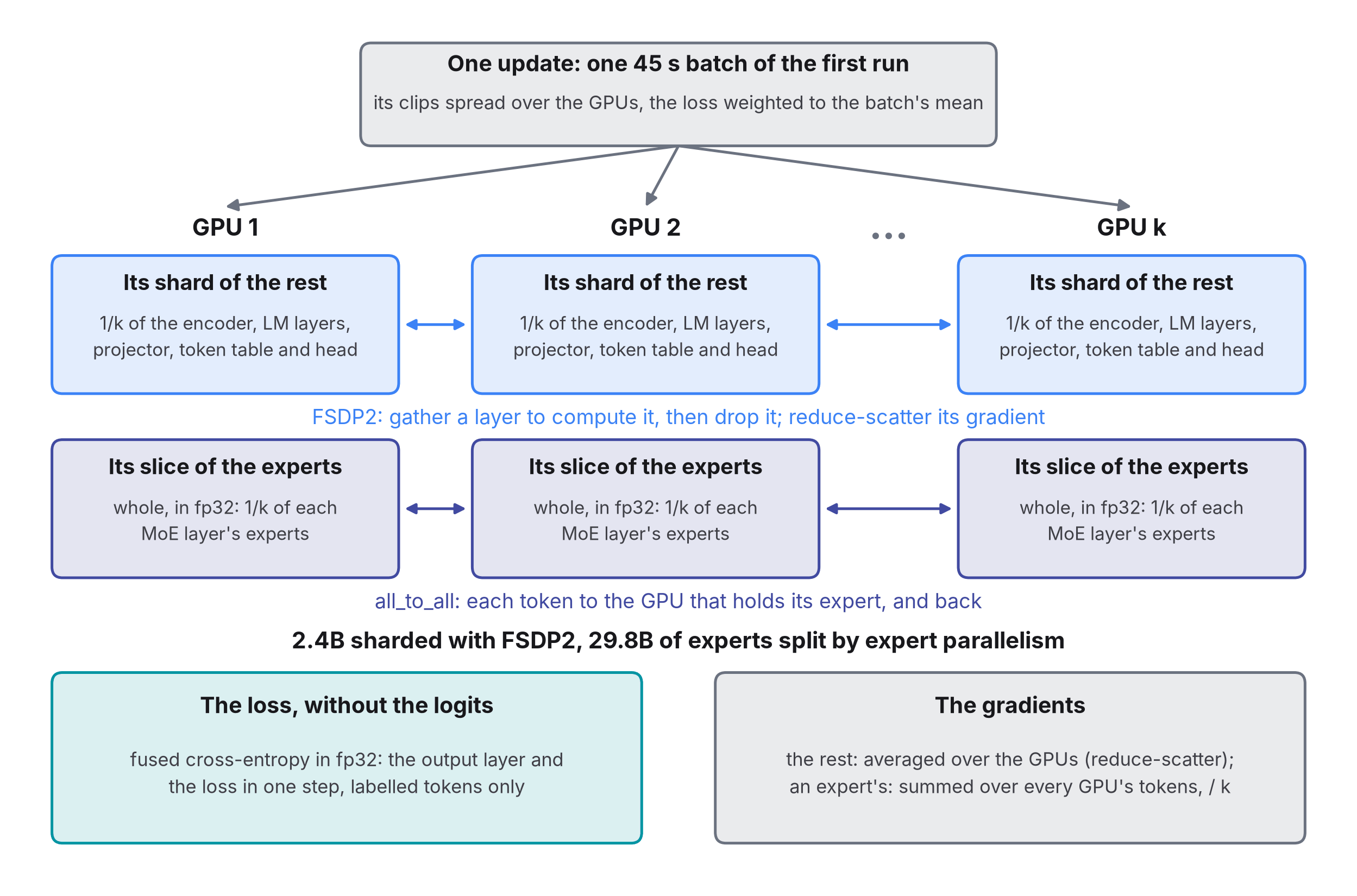}
\caption{Nemotron 30B's second run on the GPUs of one job: the experts split by expert parallelism, everything else sharded by FSDP2, and the loss computed without the logits \citep{scicom2026crossentropy}. Each update is the first run's: the same batch, its clips spread over the GPUs.}
\label{fig:parallel}
\end{figure}

\textbf{Language-model decoders.} Qwen3-ASR, Granite and Voxtral train whole: the audio encoder, the projector that carries its output into the language model (a Q-Former in Granite), the language model and its output head. Their tokenizers spell every script, so nothing is added. The language enters as text in the prompt, its name for Qwen3-ASR and Granite and a short code for Voxtral, and the loss counts only the transcript. Nemotron 30B is a mixture of experts, and its first run trains only part of it: its \val{rec/nemo30/experts}B expert weights stay frozen in bf16, with the output head, the parts that transcription never runs (speaker diarisation, word timings) and the \val{rec/nemo30/proj}M-parameter projector. Its ASR encoder and every language-model weight outside the experts train in fp32: \val{rec/nemo30/trained}B parameters, \val{rec/nemo30/share}\% of the model (Figure~\ref{fig:parts}). Our freezing rule matched names, and it was meant to train the projector and freeze the token table; it did the reverse for both. A second run trains every weight that transcription uses, \val{rec/nemo30/all}B: the experts split across GPUs by expert parallelism, everything else sharded with FSDP2, and the loss computed without the logits by a fused cross-entropy in fp32 \citep{scicom2026crossentropy} after Cut Cross-Entropy \citep{wijmans2024cut} (Figure~\ref{fig:parallel}). Dev prefers it, so it is Nemotron 30B's main run from here on: at epoch \val{rec/nemo30/all-epoch} of four it scores \val{rec/nemo30/all-cer} on the test set, \val{rec/nemo30/all-gain} points below the first run at its epoch \val{rec/nemo30/first-epoch} (\val{rec/nemo30/all-gain-ci}). Section~\ref{sec:nemo-abl} trains its parts one at a time. Voxtral and Nemotron 30B's first run train with gradient checkpointing.

\textbf{General assistants.} Gemma 4 and LFM3-Audio are chat models that transcribe when asked \citep{gemmateam2026gemma4,liquidAI2026LFM3}. Each gets the prompt its card gives for speech recognition, with the language's name in it, and nothing is added: their tokenizers spell every script. The loss counts the transcript and the token that ends the turn. The parts a transcript never runs stay frozen, Gemma's vision parts and LFM3-Audio's speech output. Gemma E2B and E4B also keep a token table for every layer, \val{rec/gemma4-e2b/lookup}B and \val{rec/gemma4-e4b/lookup}B parameters, \val{rec/gemma4-e2b/lookup-share}\% and \val{rec/gemma4-e4b/lookup-share}\% of what is stored. They are lookups, Google counts the models without them, and we leave them frozen, so \val{rec/gemma4-e2b/trained}B and \val{rec/gemma4-e4b/trained}B train (Figure~\ref{fig:parts}). Gemma 12B has no audio encoder: a linear layer hands the waveform to the language model in 40 ms pieces, and all \val{rec/gemma4-12b/trained}B of it train, laid layer by layer over the GPUs of its job. It and E4B train with gradient checkpointing.

\textbf{Transducers.} Parakeet and Nemotron streaming train whole. Their SentencePiece vocabularies come from their own languages, so every character the training transcripts need and the vocabulary lacks gets a new row in the prediction network's embedding and in the joint network's output: \val{rec/parakeet/new} for Parakeet, \val{rec/nemo-stream/new} for Nemotron streaming. The rows start silent (Section~\ref{sec:models}) and train without weight decay at ten times the model's rate and at least 1e-3, \val{rec/parakeet/row-rate} for Parakeet and \val{rec/nemo-stream/row-rate} for Nemotron streaming. Parakeet has no language input. Nemotron streaming takes the language as a one-hot prompt slot, and a language it does not list gets an unused slot that starts as a copy of its ``auto'' slot. They train on the TDT and the RNN-T loss.

\textbf{CTC models.} A CTC model has no decoder: its head writes a character or a blank for every frame of the encoder's output. w2v-BERT 2.0 comes without a head and MMS's heads belong to its own languages (Section~\ref{sec:models}), so each gets a new linear head over the \val{rec/ctc/chars} characters of the training transcripts. For these two the transcripts are lower-cased and stripped of punctuation, as is usual for a CTC head with no language model to put them back. The encoder trains with the head, every weight of it, MMS's English adapter included. MMS's head trains at \val{rec/mms/head-rate}. w2v-BERT's sits behind a LayerNorm, because the encoder's output is so small that a plain new head stayed on the blank, and trains at the encoder's rate. omniASR keeps its own head of \val{rec/omni/vocab} characters and adds a row for each of the \val{rec/omni/new} it lacks, all at the model's rate. Every model is scored on the same normalised text (Section~\ref{sec:metric}), so case and punctuation never count.

We add no augmentation of our own. Each model keeps the dropout and layer drop its own configuration switches on in training, and MMS's configuration also masks \val{rec/mms/mask}\% of the encoder's frames in spans of \val{rec/mms/mask-len}, a form of SpecAugment (Table~\ref{tab:recipe}). A batch holds a fixed amount of padded audio, or a fixed number of clips for Whisper and Voxtral, whose encoders take 30-second windows.

\begin{table}[H]
\centering
\caption{What each fine-tuned model trains with. Rate is the model's own, chosen on dev (Appendix~\ref{app:lr}). Added is what the training set needed and the model lacked; it trains at the model's rate unless another is given. Batch is clips, or seconds of padded audio. The last column is the dropout, layer drop and masking the model's own configuration switches on in training; we add nothing.}
\label{tab:recipe}
\scriptsize
\input{table_recipe}
\end{table}

\subsection{Before and after}

\label{sec:before-after}

\begin{figure}[H]
\centering
\includegraphics[width=0.95\linewidth]{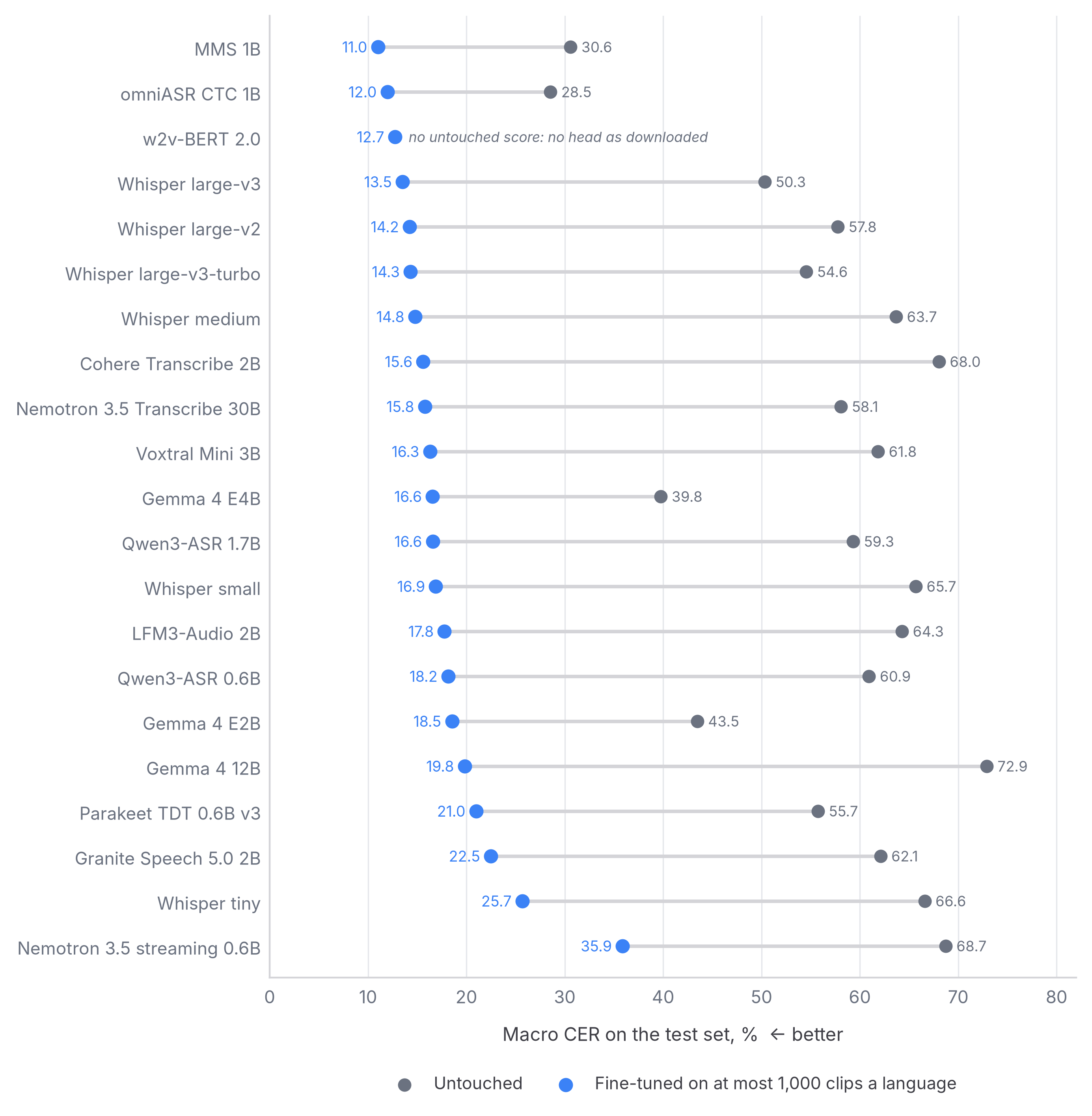}
\caption{Every model before and after fine-tuning on at most 1,000 clips a language, 50 languages together. The epoch is the one dev chose. MMS untouched uses its own adapter and head for each language that has one and English's for the rest (Section~\ref{sec:untouched}); w2v-BERT has no head as downloaded, so no untouched score.}
\label{fig:main}
\end{figure}

Table~\ref{tab:main} and Figure~\ref{fig:main} are the paper's main result, the \val{ft/n} models fine-tuned with the main recipe.

\begin{table}[H]
\centering
\caption{The main runs, best first. CER and WER on the test set at the epoch dev chose, capped per clip, macro over 50 languages. Hours are for one training epoch. Dev also chose how the transducers' new rows start (Section~\ref{sec:start}): for Parakeet, \val{ft/parakeet-tdt-v3/start}; for Nemotron streaming, \val{ft/nemotron-stream/start}.}
\label{tab:main}
\scriptsize
\input{table_main}
\end{table}

\begin{figure}[H]
\centering
\includegraphics[width=\linewidth]{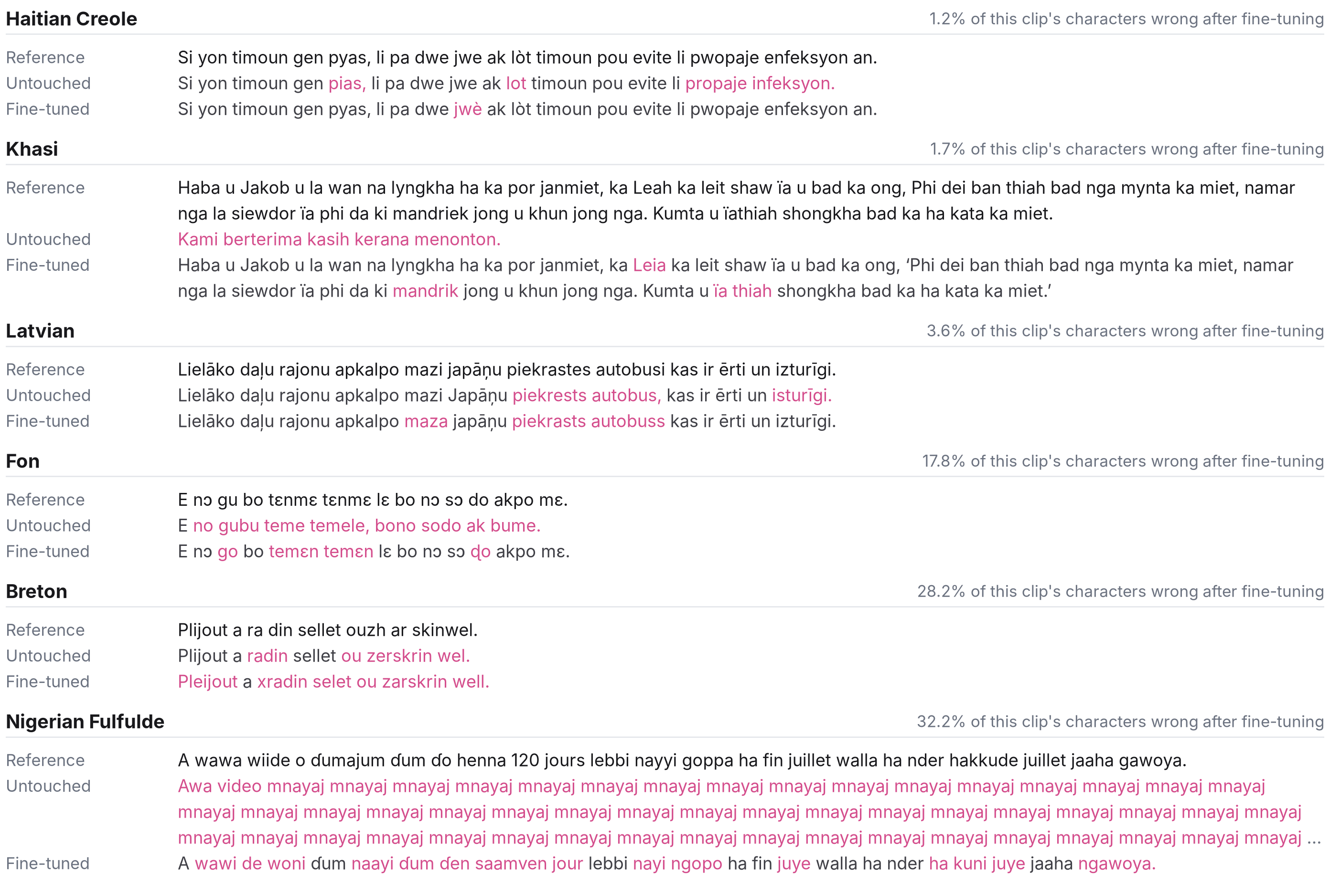}
\caption{Whisper large-v3, untouched and fine-tuned, on one clip of each of six Latin-script languages of the public test set: the clip at the median of the language's errors after fine-tuning, so nobody chose a flattering one. Words that differ from the reference are in pink, punctuation and case aside. Long clips are cut at three lines.}
\label{fig:examples}
\end{figure}

The rest of Figure~\ref{fig:baseline} is \val{sv/untuned-n} models we only scored as downloaded (\val{sv/untuned-names}) and the three extra sizes trained for Section~\ref{sec:size}, Whisper base and omniASR CTC 300M and 7B.\late{none-pending}{}{ \val{sv/pending-names} \val{sv/pending-was} still training when this was written: a model marked * is shown, here and in every table and figure, at the best of the epochs it had finished.} Every model gains, by \val{ft/gain-min} to \val{ft/gain-max} points of macro CER, and the best reaches \val{ft/best/cer} (\val{ft/best/name}); omniASR CTC 7B, trained with the same recipe for Section~\ref{sec:size}, goes lower still, to \val{size/omni-ctc-7b/after}. Another seed moves little: the \val{seedall/n-models} models we trained again, \val{seedall/n-draws} times in all, land within \val{seedall/max-diff} points of their scores here (some draws are the first ten epochs of the long runs in Section~\ref{sec:epochs}). The first three keep their order on every seed: over \val{seeds/mms-1b/n} seeds each, MMS scores \val{seeds/mms-1b/min} to \val{seeds/mms-1b/max}, omniASR CTC 1B \val{seeds/omni-ctc-1b/min} to \val{seeds/omni-ctc-1b/max} and w2v-BERT \val{seeds/w2v-bert/min} to \val{seeds/w2v-bert/max}. A language has about \val{data/k1000-minutes} minutes of speech on average, trained alongside 49 others, and that takes a recogniser that could not write it to one that gets \val{ft/best/right}\% of its characters right. Words are harder: the lowest WER among them is \val{ft/best-wer/wer} (\val{ft/best-wer/name}) and omniASR CTC 7B's \val{size/omni-ctc-7b/wer}, so this is a draft to correct, not a transcript. Figure~\ref{fig:examples} shows what such numbers look like on the page: for six languages, from easy to hard, the clip at the median of the language's errors.

The yardstick is MMS as downloaded, with an adapter and a head for each language it knows, built for the long tail and trained on far more of it. On the \val{an/mms/n} of our languages it has a head for it scores \val{an/mms/own} as downloaded. On those same languages, fine-tuned on this small set, \val{an/mms/beat-n} models do better: \val{an/mms/beat-list}. MMS 1B in that list is the same model with its heads replaced by one trained on this set and never told the language, so on this test set a little speech from the test's own corpora does more for MMS than its own training data, mostly readings of religious texts. Fine-tuning a general model on a little of a language does as well as a model built for that language. Section~\ref{sec:nolist} takes the other side of the same question, the \val{nl/n} languages that neither MMS nor any other model here lists.

\subsection{By decoder}

\label{sec:decoder}

Grouped by decoder, the same runs fall in a clear order (Figure~\ref{fig:decoder}): CTC at a median of \val{dec/CTC/median}, seq2seq at \val{dec/seq2seq/median}, the language-model decoders at \val{dec/language model/median} and the transducers at \val{dec/transducer/median}. Every CTC model is ahead of every other model; the closest pair, \val{dec/gap-ctc} and \val{dec/gap-other}, is \val{dec/gap} points apart (\val{dec/gap-ci} on resampled speakers; Section~\ref{sec:order} draws the languages again too). Size does not explain the order: it is the same among the models of \val{dec/band-lo}B to \val{dec/band-hi}B parameters, where every decoder has one.\ifval{dec/assist/min}{ The general assistants, Gemma 4 and LFM3-Audio, were not built to transcribe, and they end at \val{dec/assist/min} to \val{dec/assist/max}, inside the \val{dec/speechlm/min} to \val{dec/speechlm/max} of the language-model decoders made for speech.}{}

\begin{figure}[H]
\centering
\includegraphics[width=0.95\linewidth]{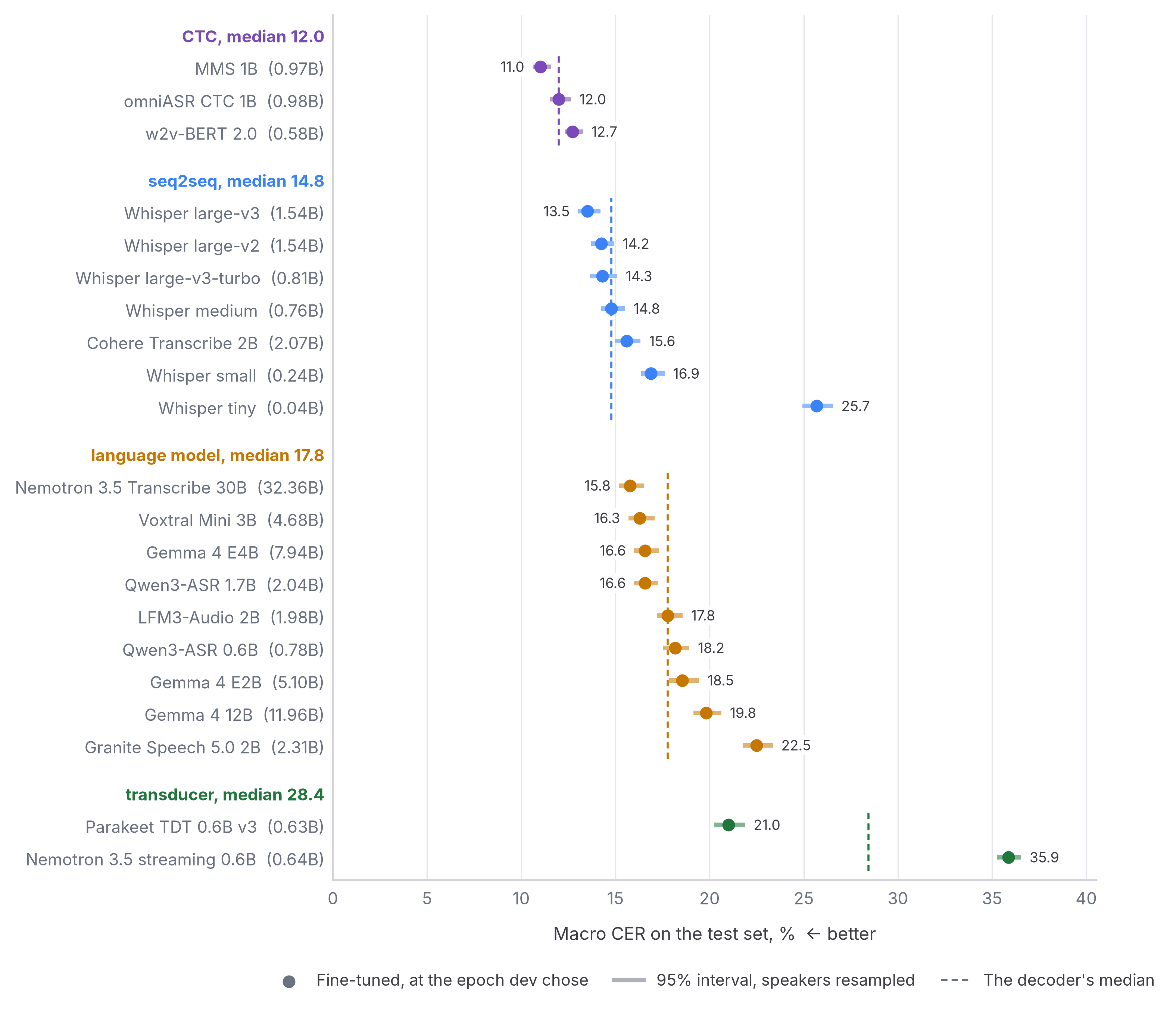}
\caption{The main runs of Table~\ref{tab:main} grouped by decoder, best first within each: macro CER on the test set at the epoch dev chose, with its 95\% interval. Parameters are beside each name.}
\label{fig:decoder}
\end{figure}

The decoder is not all that differs, so this ranks the models as they come rather than testing the decoders. The CTC models are also the three whose cards list the most languages, \val{dec/ctc-langs-min} to \val{dec/ctc-langs-max} against at most \val{dec/other-langs-max} for the rest, though they still lead, by less, on the \val{nl/n} languages that none of them lists (Section~\ref{sec:nolist}). The transducers trail on the Latin script too and fall furthest behind outside it (Section~\ref{sec:script}), where the rows we add have no pretrained letter behind them. The one pair built on the same pretrained encoder with different decoders, omniASR's CTC 1B and LLM 1B, scores \val{bl/omni-ctc-1b/cer} and \val{bl/omni-llm-1b/cer} untouched, which suggests the encoder sets the order more than the decoder does; only the CTC variant was fine-tuned here, so the same test after training is still open.

\subsection{How firm the order is}

\label{sec:order}

Every interval so far draws the speakers again and keeps the 50 languages, so it says how firm a difference is on these languages. Whether it would hold on another fifty is a second question, and Figure~\ref{fig:order} asks both. It draws the test set again \val{order/draws} times, the speakers alone and then the languages with their speakers, and counts how often the better model of each pair stays ahead.

\begin{figure}[H]
\centering
\includegraphics[width=0.92\linewidth]{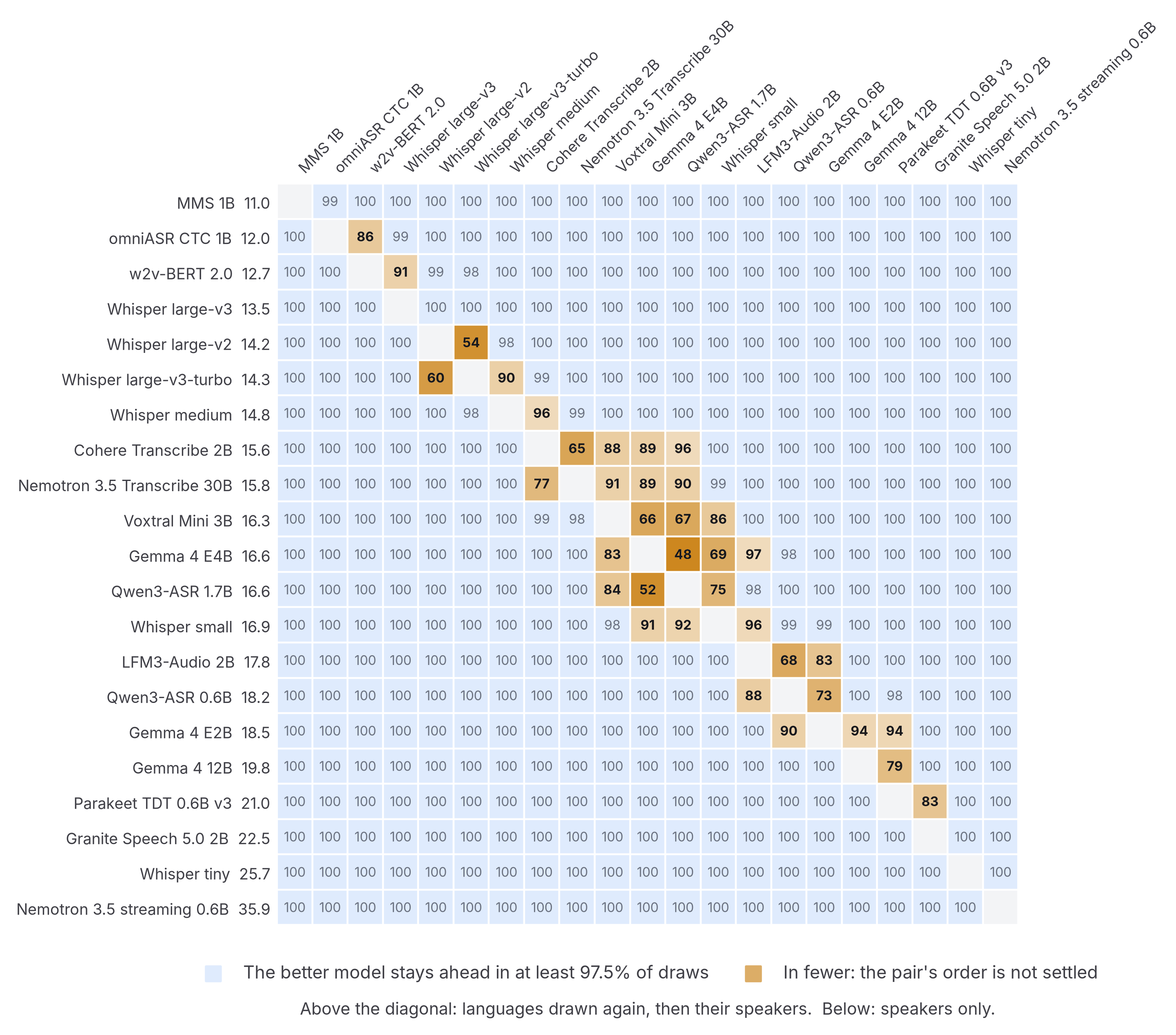}
\caption{The main runs of Table~\ref{tab:main}, best first, with their macro CER. Each cell is a pair of models: the percentage of \val{order/draws} draws of the test set in which the one ahead on the test set stays ahead. Below the diagonal a draw resamples the speakers inside each language, as the paper's intervals do; above it, the 50 languages are resampled first.}
\label{fig:order}
\end{figure}

With the speakers alone, \val{order/spk/held} of the \val{order/pairs} pairs keep their order in at least 97.5\% of the draws; with the languages drawn too, \val{order/both/held}. The pairs that come loose are never more than \val{order/both/loose-reach} places apart, and \val{order/top/name} is first in \val{order/both/top-first}\% of the draws. One of them is the pair Section~\ref{sec:decoder} rests on: \val{dec/gap-ctc} stays ahead of \val{dec/gap-other} in \val{order/both/dec-share}\% of the draws (a difference of \val{order/both/dec-ci}), so on another fifty languages a Whisper may well pass the last CTC model.

\subsection{Told the language}

Before any training a model can already be told what it is listening to, and Table~\ref{tab:zero} shows how little that does: the middle column has no training at all. The name of a language in Qwen3-ASR 0.6B's prompt is worth \val{z/qwen3-asr-0.6b/gain} points, mostly the script: told the language, it writes the Latin-script languages in Latin on \val{sc/qwen3-asr-0.6b/told/latin}\% of clips instead of \val{sc/qwen3-asr-0.6b/untouched/latin}\%, while the other scripts barely move.\ifval{z/dagger}{ The name is worth \val{z/lfm3-audio/gain} points to LFM3-Audio and \val{bl/gemma/told-gain-span} to Gemma 4's \val{bl/gemma/sizes} with an audio encoder.}{} A new Whisper language token, the mean of the others, has little to say yet, and Cohere Transcribe's untrained tag, the mean of its fourteen, does harm: \val{z/cohere-transcribe} against \val{bl/cohere-transcribe/cer} for the model told English. Telling is not teaching: the right-hand column is where the gain is.

\begin{table}[H]
\centering
\caption{The test set's macro CER with nothing trained, and after. The middle column is the first evaluation of each main run: the language's name in the prompt, or a new language token for Whisper and Cohere Transcribe.\ifval{z/dagger}{ $^\dagger$Gemma 4 is told every language as downloaded (Section~\ref{sec:untouched}), so its middle column is its untouched score and its first the model with no language in the prompt.}{}}
\label{tab:zero}
\scriptsize
\input{table_zero}
\end{table}

\subsection{Language by language}

Some languages are almost solved (Figures~\ref{fig:perlang} and~\ref{fig:heatmap}), the five lowest across any model being \val{an/easy-names}. Others resist every model, the five highest being \val{an/hard-names}; the numbers are each language's best CER across all models. When every model fails on a language in the same way, the cause is more likely to be that language's data than the models.

\begin{figure}[H]
\centering
\includegraphics[width=0.95\linewidth]{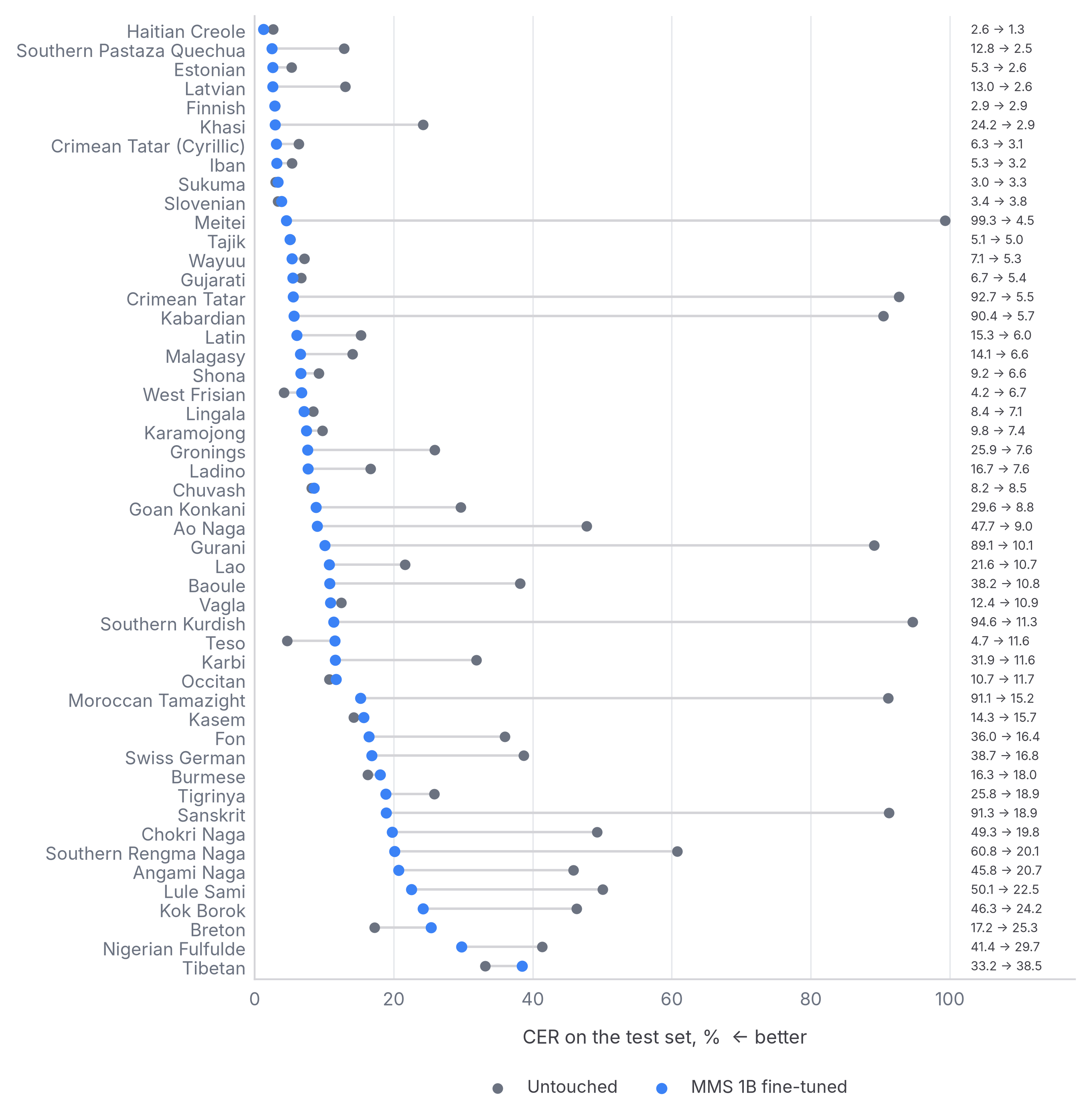}
\caption{The strongest model of the main result, \val{ft/best/name}, language by language, before and after.}
\label{fig:perlang}
\end{figure}

\begin{figure}[H]
\centering
\includegraphics[width=0.95\linewidth]{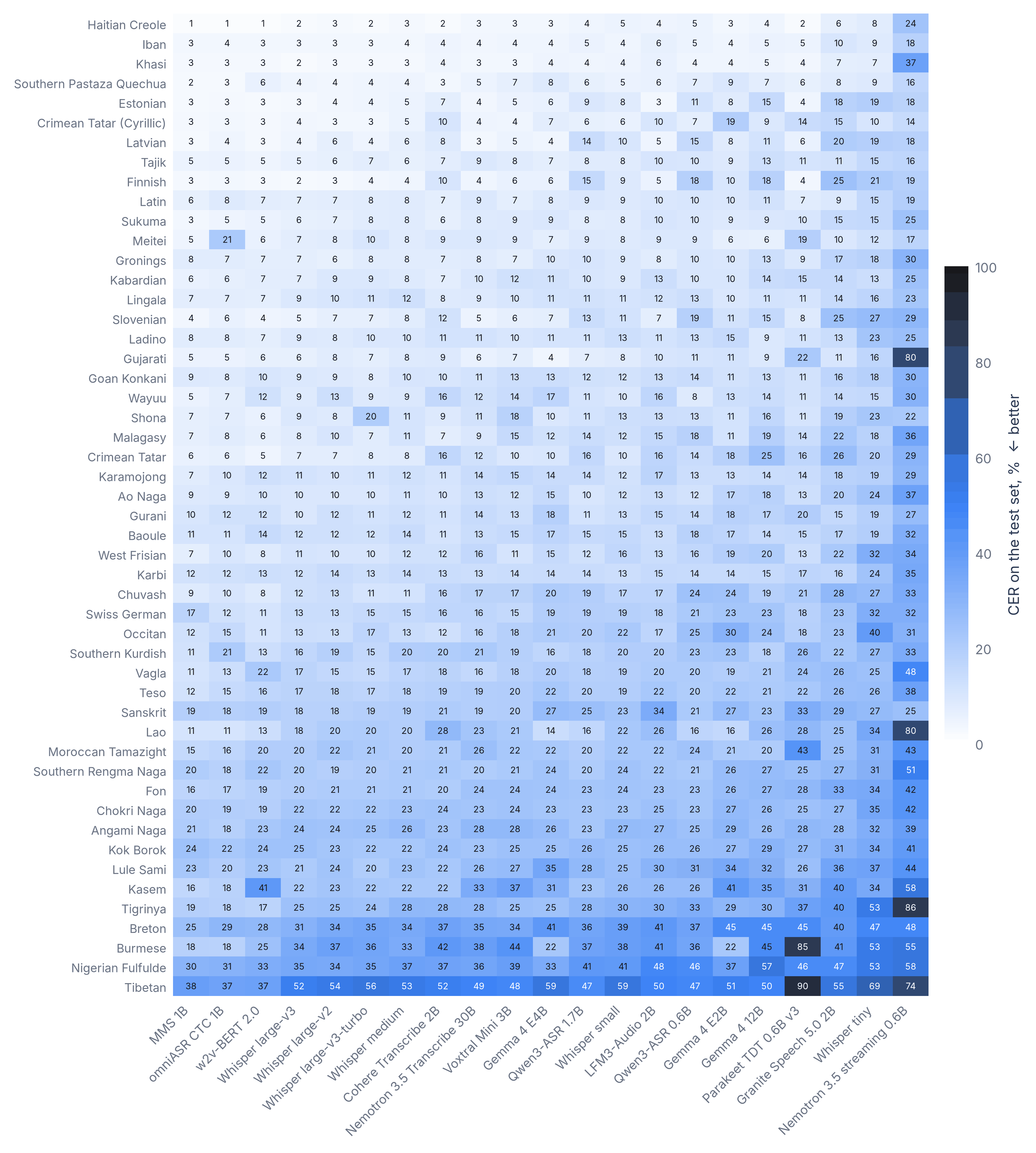}
\caption{Every fine-tuned model on every language. A row that stays dark is a language no model learns; a column that stays dark is a model that cannot write that language's script.}
\label{fig:heatmap}
\end{figure}

What goes with a hard language is how it was recorded more than how much of it there is (Figure~\ref{fig:hard}). By each language's CER in the median model, the half of the languages with more speakers in their training clips is \val{hard/speakers/diff} points harder than the half with fewer (\val{hard/speakers/ci}, languages resampled), the half with shorter clips \val{hard/length/diff} harder than the half with longer ones (\val{hard/length/ci}), and the \val{hard/shared/nb} languages whose test speakers are new \val{hard/shared/diff} harder than the \val{hard/shared/na} whose test speakers also train (\val{hard/shared/ci}). A language read by a few voices in long clips is easy; one gathered from many speakers a few seconds at a time is hard. The number of clips says nothing (Spearman's $\rho$ \val{hard/clips/rho}), and neither the script nor being on no model's list separates the languages by more than its interval. Together the six properties leave \val{hard/unexplained}\% of the variance between languages unexplained.

\begin{figure}[H]
\centering
\includegraphics[width=0.95\linewidth]{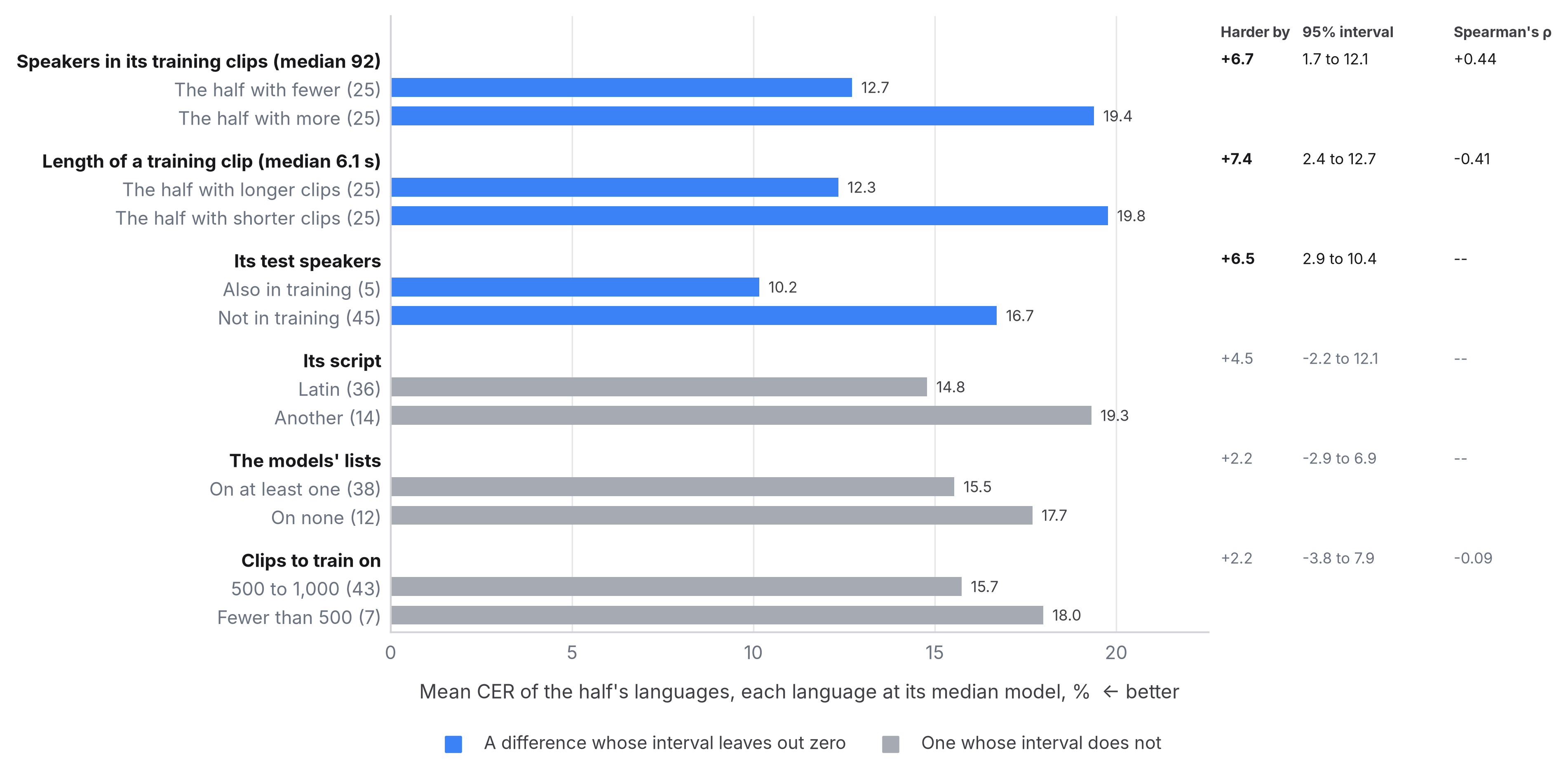}
\caption{The 50 languages split in two by one property at a time, the harder half second: the mean test CER of each half's languages, each language at its median over the main models. Harder by is the difference, with an interval from resampling the languages; Spearman's $\rho$ is between the property, where it is a number, and the language's CER.}
\label{fig:hard}
\end{figure}

\subsection{Model size}

\label{sec:size}

A larger model transfers more within a family of one design (Figure~\ref{fig:size}). Whisper large-v3 ends \val{size/step/whisper-tiny-large} points below Whisper tiny, \val{size/step/whisper-tiny-small} of them already at small, and Qwen3-ASR 1.7B is \val{size/step/qwen} below the 0.6B. omniASR CTC 1B is \val{size/step/omni-300m-1b} below the 300M, and the 7B \val{size/step/omni-1b-7b} below the 1B (\val{size/step/omni-1b-7b-ci}): at \val{size/omni-ctc-7b/after} it is the best model we trained, \val{size/7b-vs-main} points below \val{size/7b-vs-main-name} (\val{size/7b-vs-main-ci}). Size counts for more when data is short: on dev, each omniASR size leads the one below it by \val{size/sweep-gap/omni-300m-1b} and \val{size/sweep-gap/omni-1b-7b} points on the sweep's budget of three epochs on 200 clips a language (Appendix~\ref{app:lr}), and by \val{size/dev-gap/omni-300m-1b} and \val{size/dev-gap/omni-1b-7b} after the full recipe.\ifval{size/gemma/behind}{ A change of design can undo it: Gemma 4's 12B, which reads the waveform with no audio encoder, ends at \val{size/gemma4-12b/after}, behind both smaller sizes, which have one (\val{size/gemma4-e2b/after} and \val{size/gemma4-e4b/after}).}{}

\begin{figure}[H]
\centering
\includegraphics[width=0.85\linewidth]{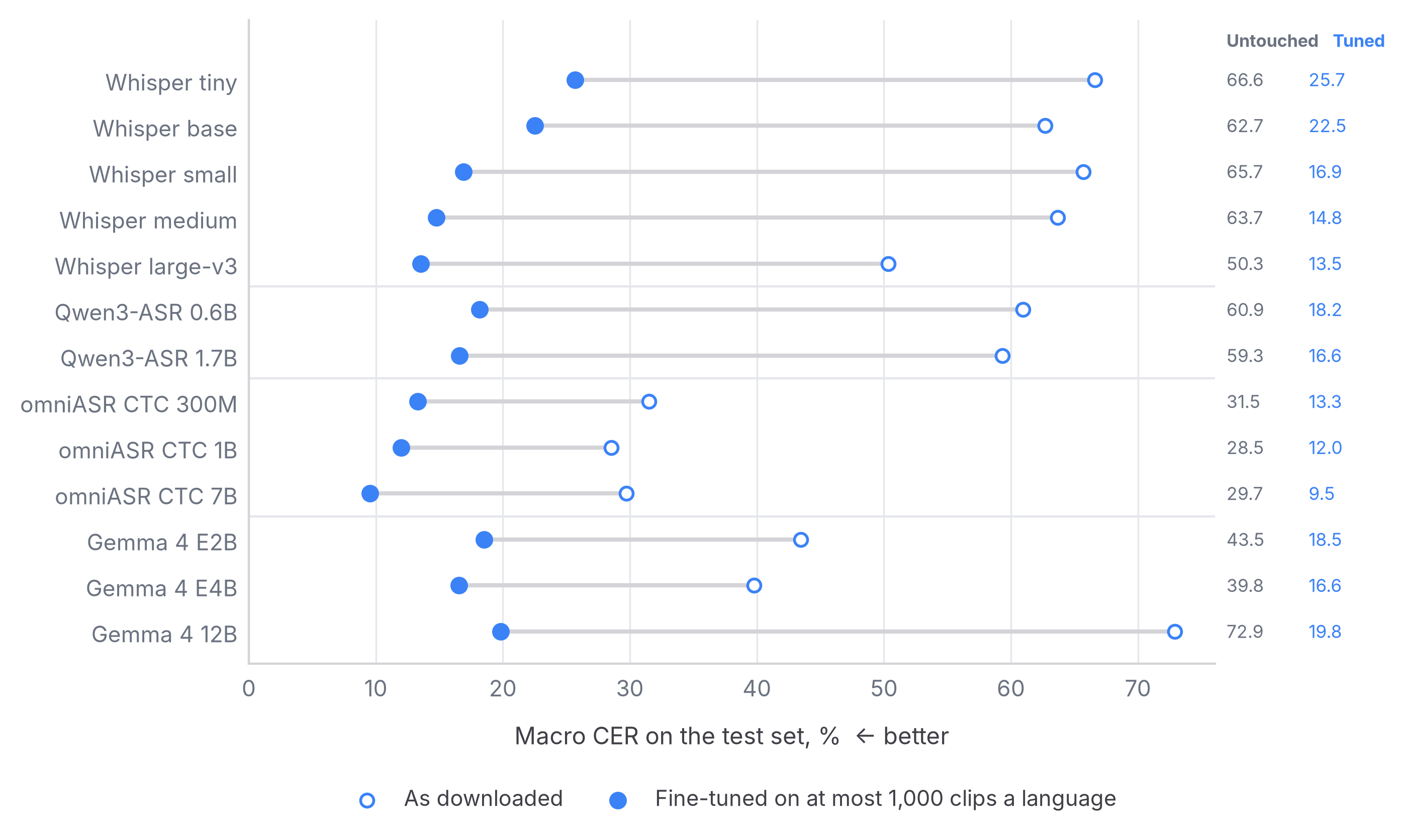}
\caption{Four families by size, as downloaded and fine-tuned with the same recipe. Macro CER on the test set at the epoch dev chose. Gemma 4's 12B is its one size with no audio encoder.}
\label{fig:size}
\end{figure}

\subsection{Which weights train}

\label{sec:nemo-abl}

Nemotron 3.5 Transcribe 30B is the one mixture of experts here, and its routed experts are \val{nabl/experts-extra}B of the \val{nabl/full/trained}B weights its second run trains. To see which parts are worth training, we trained it \val{nabl/n-more} more ways beside its two runs for \val{nabl/upto} epochs, the second run's recipe with some parts frozen, on another seed, at another rate for all of it or for the encoder alone, or with the rate falling to zero over the two epochs, each scored at the epoch dev chooses (Figure~\ref{fig:nemo-abl}).

The audio encoder has to train: freezing it costs \val{nabl/cost/enc} points with the routed experts training (\val{nabl/cost/enc-ci}) and \val{nabl/cost/enc-frz} with them frozen (\val{nabl/cost/enc-frz-ci}). So does the language model, since the audio side alone ends \val{nabl/cost/lm} points behind (\val{nabl/cost/lm-ci}). The projector does not matter: freezing it moves a run by at most \val{nabl/projector-max} points, while two seeds end \val{nabl/seed-diff} apart with everything training and \val{nabl/seed-diff-frz} with the routed experts frozen. Nor do the routed experts much: freezing them costs \val{nabl/cost/experts} points (\val{nabl/cost/experts-ci}) and leaves \val{nabl/frz-experts/trained}B weights to train, which need no expert parallelism. The first run, which also froze the shared experts and the output head, ends \val{nabl/first/vs-full} behind everything training (\val{nabl/first/vs-full-ci}), and the shared experts account for it: freezing them with the routed ones costs \val{nabl/cost/shared} (\val{nabl/cost/shared-ci}) and ends level with the first run (\val{nabl/frz-experts-shared/vs-first-ci}). A lower rate for everything, 3e-6, ends \val{nabl/full-lr3e-6/ahead} points below 1e-5 (\val{nabl/full-lr3e-6/ahead-ci}), about as far as two seeds end apart; with the routed experts frozen, 3e-6 ends \val{nabl/frz-experts-lr3e-6/vs-rate} behind 1e-5 (\val{nabl/frz-experts-lr3e-6/vs-rate-ci}) and 3e-5 \val{nabl/frz-experts-lr3e-5/vs-rate} behind (\val{nabl/frz-experts-lr3e-5/vs-rate-ci}). The encoder wants no faster a rate than the rest: at 3e-5 it ends \val{nabl/frz-experts-enc3e-5/vs-rate} behind (\val{nabl/frz-experts-enc3e-5/vs-rate-ci}), at 1e-4 \val{nabl/frz-experts-enc1e-4/vs-rate}.

\begin{figure}[H]
\centering
\includegraphics[width=0.9\linewidth]{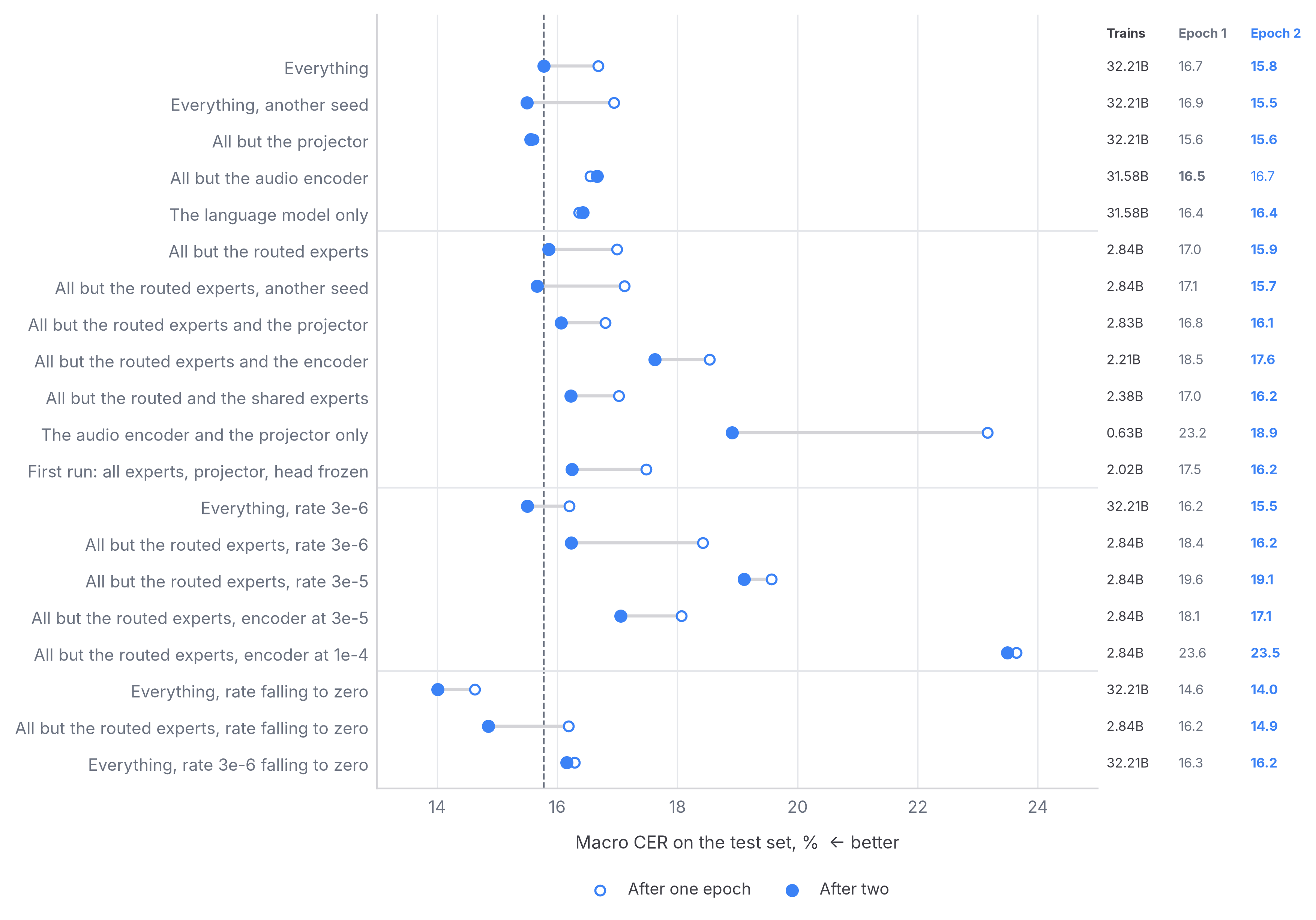}
\caption{Nemotron 3.5 Transcribe 30B trained \val{nabl/upto} epochs, part by part: test macro CER after one epoch (hollow) and two (filled), the weights each variant trains, and in bold the epoch dev chose. The rules separate the variants that train the routed experts, those that keep them as downloaded, those that change a rate, and those whose rate falls to zero over the two epochs. The dashed line is everything training, after two epochs.}
\label{fig:nemo-abl}
\end{figure}

The schedule gives what the parts do not: with the rate falling to zero over the two epochs, as in Section~\ref{sec:epochs}, everything training ends \val{nabl/sched/full-gain} points lower than at a constant rate (\val{nabl/sched/full-gain-ci}), at \val{nabl/full-linear/test}, and the run with the routed experts frozen \val{nabl/sched/frz-gain} lower (\val{nabl/sched/frz-gain-ci}), at \val{nabl/frz-experts-linear/test}: both below every variant at a constant rate (the best of them \val{nabl/const-best-test}). The routed experts, worth \val{nabl/cost/experts} at a constant rate, are worth \val{nabl/cost/experts-linear} once the rate falls (\val{nabl/cost/experts-linear-ci}). The fall needs a high start: from 3e-6, everything training ends \val{nabl/full-lr3e-6-linear/vs-linear} behind the fall from 1e-5 (\val{nabl/full-lr3e-6-linear/vs-linear-ci}) and \val{nabl/full-lr3e-6-linear/vs-rate} behind 3e-6 held constant (\val{nabl/full-lr3e-6-linear/vs-rate-ci}).

A longer fall does not help. Planned for four epochs instead of two, everything training ends at \val{nabl/e4/test} at the epoch dev chooses, its last, \val{nabl/e4/behind-two} behind the two-epoch run (\val{nabl/e4/behind-two-ci}) and \val{nabl/e4/vs-main} below the main run, which held the rate constant over its four (\val{nabl/e4/vs-main-ci}). Two epochs with the rate falling to zero are the best recipe we found for this model, at \val{nabl/full-linear/test}; the main result keeps the constant rate that every other model trained at.

\section{What transfers}

\label{sec:transfer}

Studies that need many runs use a small model in place of a large one: Whisper small, which trains an epoch in \val{ft/whisper-small/h-epoch} hours against \val{ft/whisper-large-v3-turbo/h-epoch} for large-v3-turbo, and Qwen3-ASR 0.6B rather than 1.7B. Whisper large-v3-turbo repeats the studies of the next language (Section~\ref{sec:next}), of training longer (Section~\ref{sec:epochs}) and of data size (Section~\ref{sec:scale}) to check that a larger model answers the same way.

\subsection{Pretrained against random weights}

The direct measure of transfer is the same architecture, on the same data, without its pretraining (Figure~\ref{fig:scratch}). From random weights every model ends \val{scr/gap-min} to \val{scr/gap-max} points behind its pretrained twin, and Parakeet never stops looping: after \val{scr/parakeet-tdt-v3/done} epochs, \val{scr/parakeet-tdt-v3/loop} of the \val{data/test-clips} test clips still loop. These runs had an untuned rate of 1e-4 and, but for Parakeet's, twice the main runs' epochs, so a better recipe would close some of the gap, though hardly much of it: \val{data/k1000-hours} hours is far too little to learn speech from nothing. Nearly everything a fine-tuned model gets right here, it owes to pretraining on other languages.

\begin{figure}[H]
\centering
\includegraphics[width=0.95\linewidth]{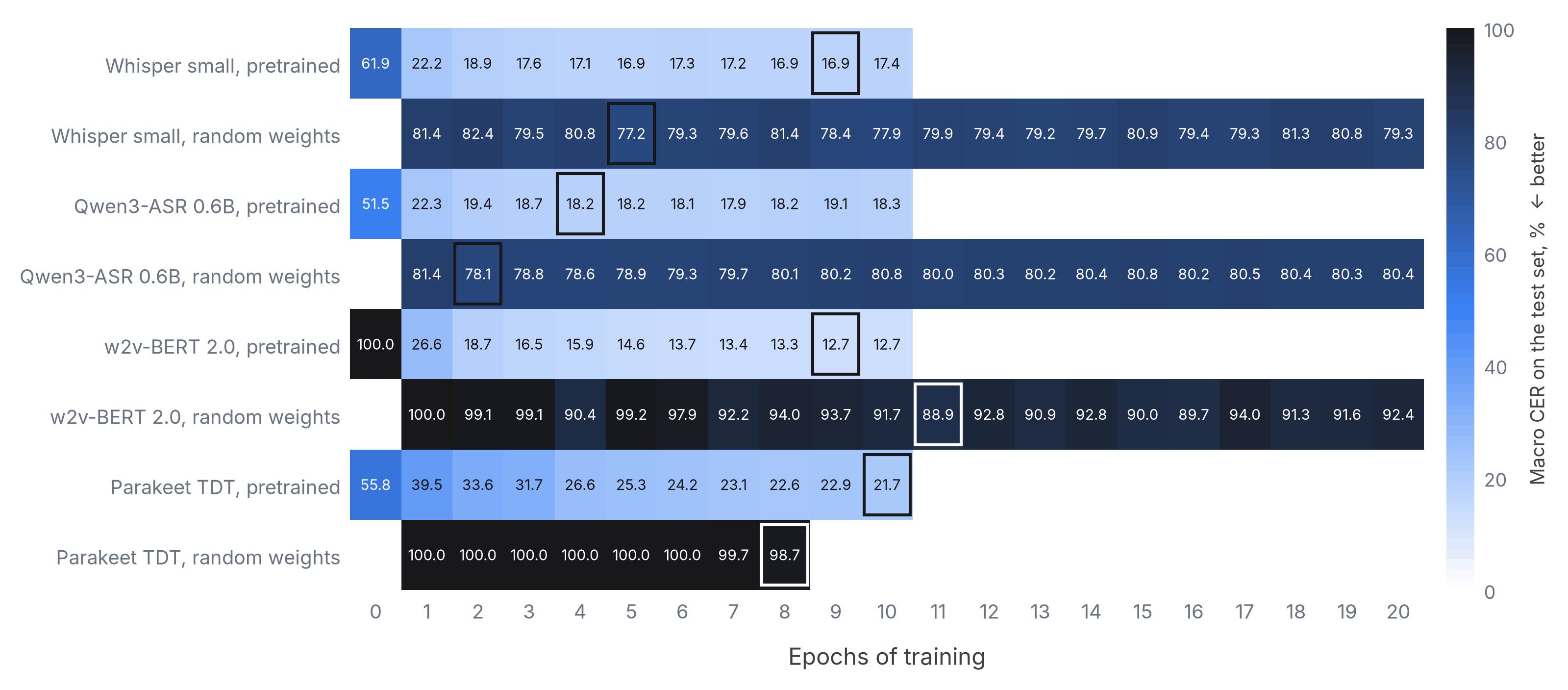}
\caption{The same architecture and the same data, from pretrained and from random weights, on the test set after every epoch. Epoch 0 is the pretrained model before training, told the language where it can be (Table~\ref{tab:zero}); w2v-BERT's head is new, so it starts at 100. A box marks the epoch dev chose; blank cells were not run.}
\label{fig:scratch}
\end{figure}

\subsection{Encoder or decoder}

To see which half carries the transfer, we freeze each half in turn for six epochs, leaving only the rows a new character needs (Figure~\ref{fig:freeze}). On a model whose decoder attends to all of the audio, either half alone learns most of a new language: Qwen3-ASR 0.6B reaches \val{fz/qwen3-asr-0.6b/all} with everything training, \val{fz/qwen3-asr-0.6b/fze} with its encoder frozen and \val{fz/qwen3-asr-0.6b/fzd} with its decoder frozen, and Whisper small behaves the same. So an encoder that has never heard these languages already gives such a decoder most of what it needs. Without such a decoder it is not enough. w2v-BERT's new head on its frozen encoder stays at \val{fz/w2v-bert/fze}, and Parakeet with its encoder frozen at \val{fz/parakeet-tdt-v3/fze}, hardly below the untouched model: a linear head cannot read a new language's letters straight out of a self-supervised representation, and a transducer's decoder sees the audio only one frame at a time. In these the encoder has to learn the new language.

\begin{figure}[H]
\centering
\includegraphics[width=0.85\linewidth]{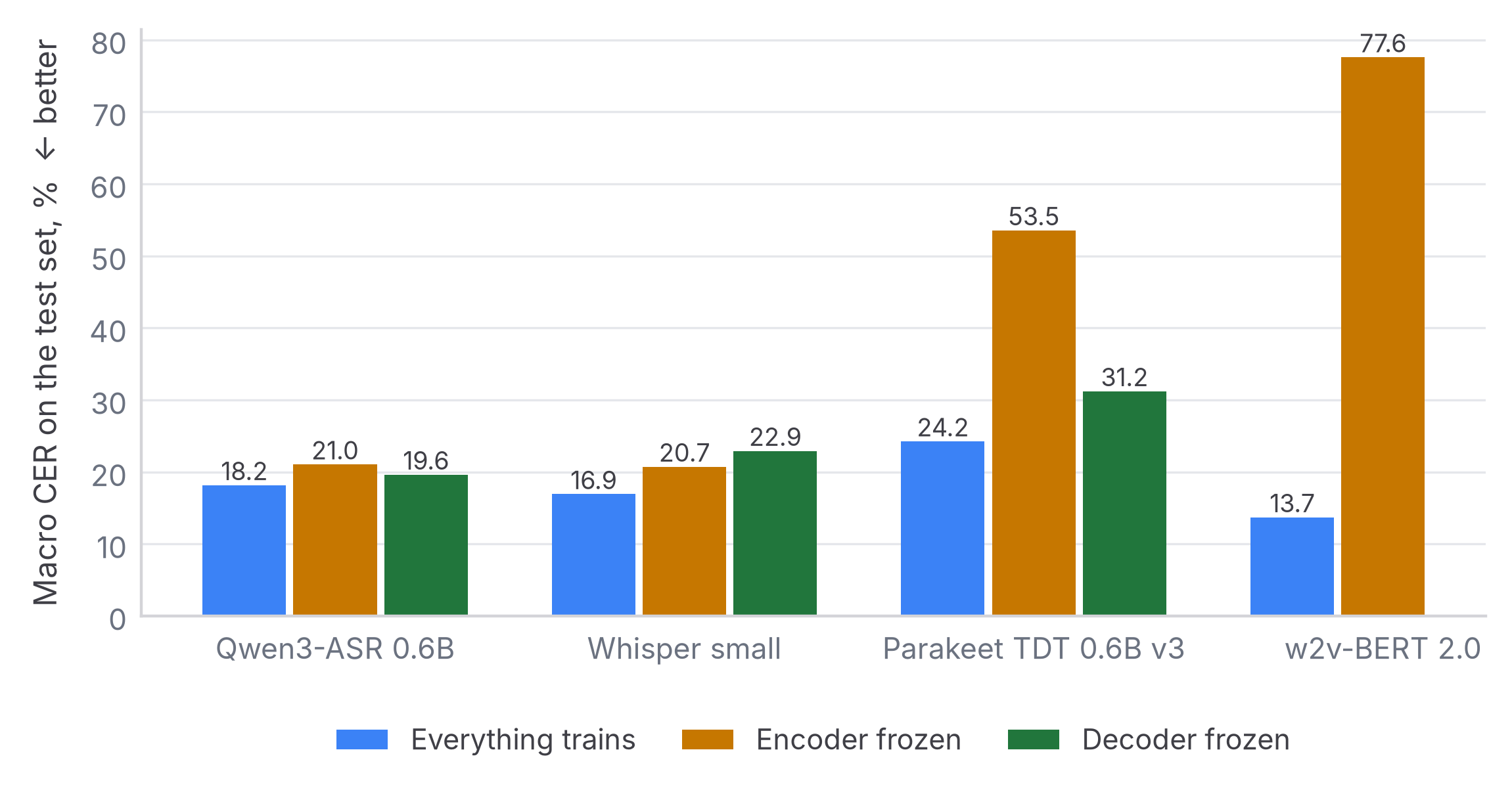}
\caption{Which half has to learn, at the epoch dev chose among the first six. The frozen half keeps its pretrained weights, apart from new rows for new characters. Parakeet's decoder is its prediction network and joint; w2v-BERT has no decoder to freeze.}
\label{fig:freeze}
\end{figure}

\subsection{The script}

\label{sec:script}

The models that can already spell any script end close together on the Latin script and on the others (Figure~\ref{fig:script}), while the transducers end at \val{an/script/transducer/other} outside Latin against \val{an/script/transducer/latin} on it: their vocabularies were built from their own languages, and the rows we add for a new script have no pretrained letter behind them (Section~\ref{sec:start}). The script is also the first thing that moves (Figure~\ref{fig:scriptshare}). Untouched, a model writes the reference's script on \val{sc/untouched-min} to \val{sc/untouched-max}\% of the clips outside Latin; after a single epoch \val{sc/full-after-one} of the models write it on every clip.

\begin{figure}[H]
\centering
\includegraphics[width=0.95\linewidth]{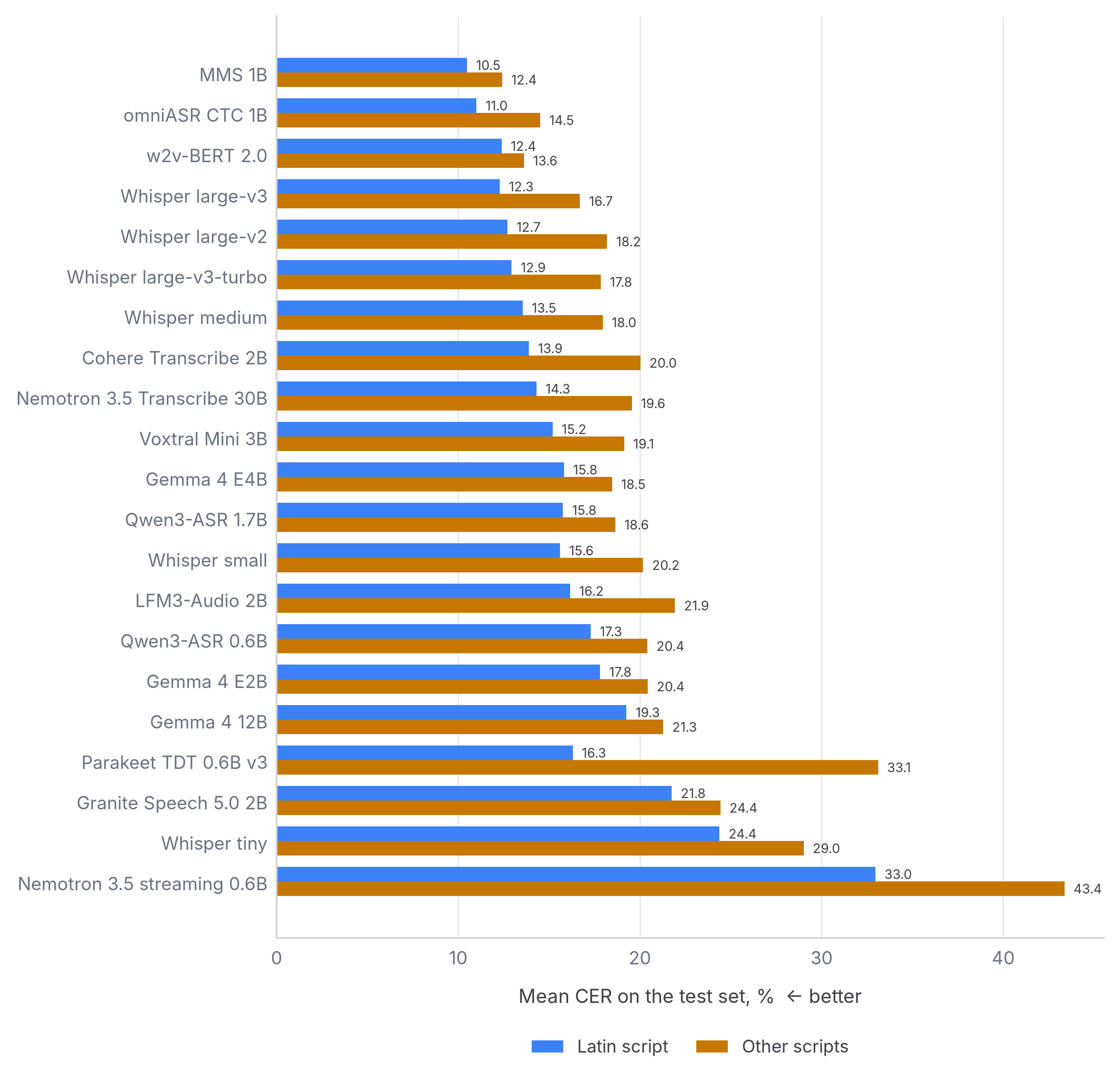}
\caption{Fine-tuned models on the Latin-script languages and on the rest.}
\label{fig:script}
\end{figure}

\begin{figure}[H]
\centering
\includegraphics[width=0.86\linewidth]{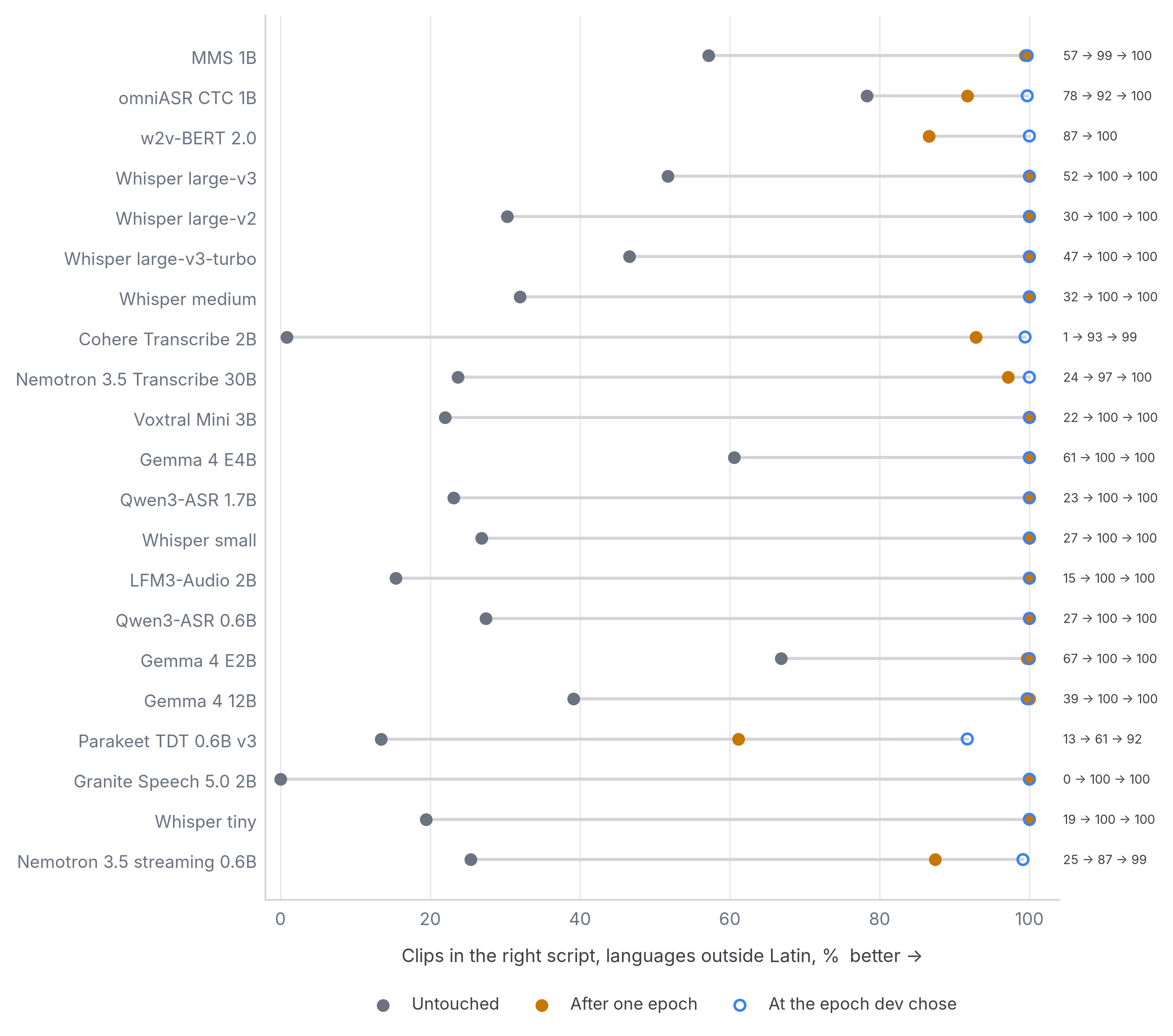}
\caption{Whether the model writes in the language's script, by the Unicode script of most of a clip's letters. Only the languages outside Latin. w2v-BERT has no head as downloaded, so no untouched score; MMS untouched uses its own heads where it has them and English's elsewhere.}
\label{fig:scriptshare}
\end{figure}

\subsection{How a new row starts}

\label{sec:start}

Section~\ref{sec:models} gave the transducers' new rows a silent start in place of a first start that was not silent. A start that changes nothing need not learn best, so both ran the full length, and the two models disagree (Figure~\ref{fig:start}): on Parakeet the silent start is ahead from the first epoch on, on Nemotron it is far behind, \val{start/nemotron-stream/ten/e1} after one epoch against \val{start/nemotron-stream/first/e1}.

\begin{figure}[H]
\centering
\includegraphics[width=0.95\linewidth]{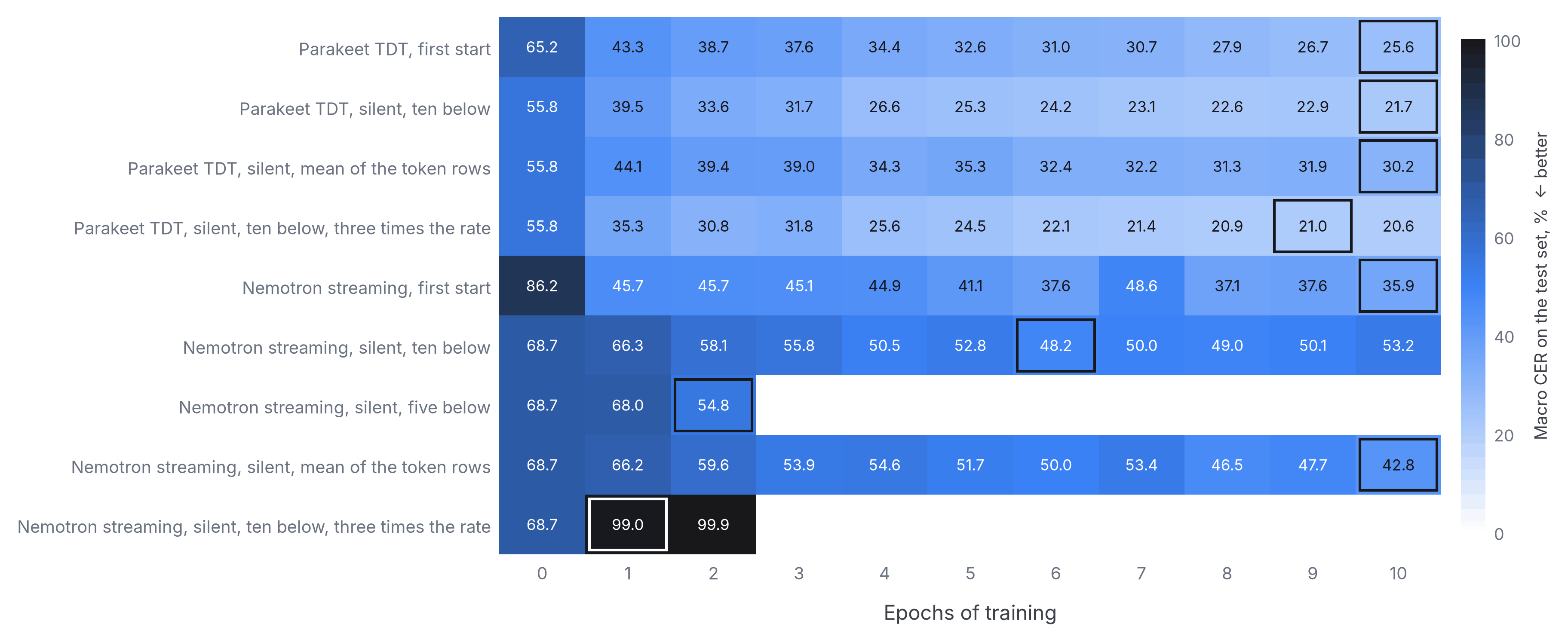}
\caption{Ways to start the transducers' new rows, on the test set after every epoch. Epoch 0 is the extended model before any training: a silent start reads the untouched score there, the first start does not. A box marks the epoch dev chose; blank cells were not run.}
\label{fig:start}
\end{figure}

The silent Nemotron writes too little, and not only where it needs new rows: after one epoch its output on the Latin-script languages is \val{start/nemotron-stream/ten/len/e1/latin}\% as long as the reference, against \val{start/nemotron-stream/first/len/e1/latin}\% from the first start, and its new rows have hardly left the blank, at a median cosine of \val{rows/nemotron-stream/ten/cos} with the blank's row after training. A start silent for another reason, each new row the mean of the pretrained token rows a point below it, leaves almost no clip empty and still learns slowly, as does five below the blank instead of ten; three times the rate makes the model fall silent, with \val{start/nemotron-stream/hi/len/last/empty} of the \val{sr/n} test clips empty after \val{start/nemotron-stream/hi/len/last-epoch} epochs. On Parakeet the mean start ends behind even the first, while the copy of the blank's row reaches \val{start/parakeet-tdt-v3/ten/cer}, and at three times the rate \val{start/parakeet-tdt-v3/hi/cer}, Parakeet's main run (Appendix~\ref{app:lr}). Leaving the pretrained model intact at the start is not the same as starting well, and Nemotron's main run is the first start, because dev chose it.

\subsection{Languages on a model's list}

\label{sec:listed}

A model that already lists a language starts ahead: as downloaded, Whisper large-v3 is at \val{an/whisper-large-v3/listed-before} on the \val{an/whisper-large-v3/listed-n} languages on its list and \val{an/whisper-large-v3/other-before} on the rest. Part of that is the languages, so Figure~\ref{fig:listed} fits the CER of every fine-tuned model on every language as the model's level plus the language's difficulty, less a lead where the language is on the model's own list. Most models do not list a given language, so its difficulty does not rest on the ones that do, and the fit keeps to the \val{ld/n-latin} Latin-script languages, so that the script cannot decide it. A list is the card's for most models. For MMS and omniASR it is the model's own table of languages, and for w2v-BERT 2.0, which never saw a transcript, the languages of its unlabelled audio.

As downloaded a listed language is \val{ld/e0-min} to \val{ld/e0-max} points ahead. On Whisper that lead is gone after \val{ld/whisper/latin/gone-epoch}, and with it the \val{sc/whisper-large-v3/untouched/latin-wrong}\% of Latin-script clips it wrote in another script as downloaded; omniASR's is gone after \val{ld/omni/latin/gone-epoch}. The other lists keep some of theirs to the epoch dev chose: MMS \val{ld/mms/latin/best} points (\val{ld/mms/latin/best-ci}, languages resampled), w2v-BERT \val{ld/w2v/latin/best} (\val{ld/w2v/latin/best-ci}), and Parakeet and the two Nemotrons \val{ld/short/latin/best} (\val{ld/short/latin/best-ci}) on the \val{ld/short/latin/n} they list, Estonian, Finnish, Latvian and Slovenian. Being on the list can even cost something: Finnish, the only language Qwen3-ASR lists, goes from \val{an/qwen3-asr-1.7b/listed-before} to \val{an/qwen3-asr-1.7b/listed-after} on the larger model when all 50 train together. Nor does its list save Whisper outside the Latin script: on \val{lh/behind/names}, all three on its list, large-v3 ends \val{lh/behind/min} to \val{lh/behind/max} points behind the median of \val{lh/ref-names}, and twice the decoding budget wins back at most \val{lh/behind/redecode-max} of that (Section~\ref{sec:limits}).

\begin{figure}[H]
\centering
\includegraphics[width=0.84\linewidth]{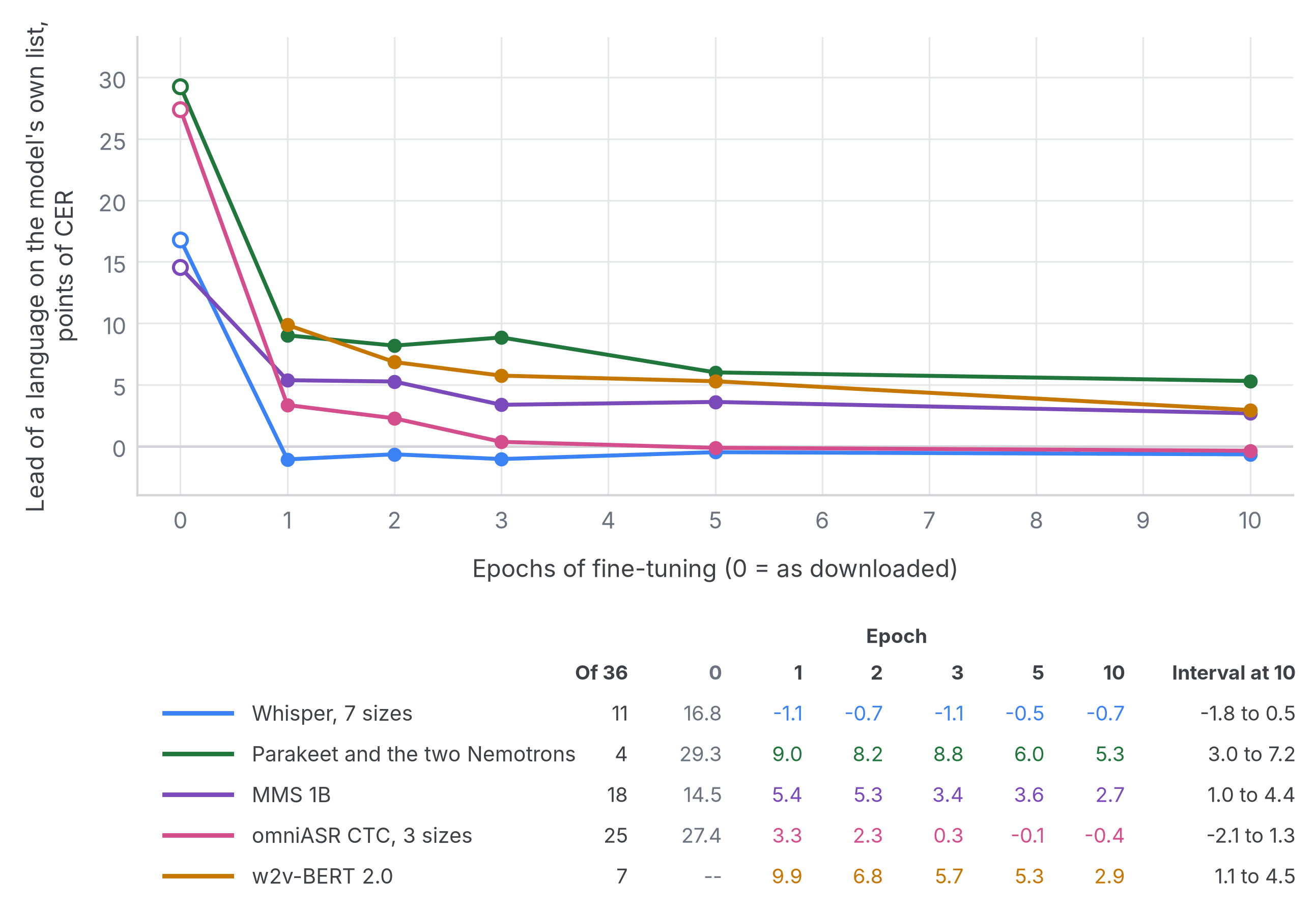}
\caption{What a language on a model's own list is still worth, epoch by epoch, on the \val{ld/n-latin} Latin-script languages. The test CER of all \val{ld/n-models} models fine-tuned with the main recipe is fitted as a model's level plus a language's difficulty, less a lead where the language is on the model's own list; the models that share a list share a lead. Epoch 0 (hollow) is the models as downloaded, and a run that stopped before an epoch stands at its last. Of \val{ld/n-latin}: the languages on the list. The interval resamples the languages.}
\label{fig:listed}
\end{figure}

To see whether a relative on the list helps, Table~\ref{tab:kin} keeps to Latin-script languages and splits them by whether the model lists the language, another of its family, or neither. Before training a relative seems to help the models with short lists, Nemotron 3.5 Transcribe being at \val{kin/nemotron-30b/relative/untouched} with one and \val{kin/nemotron-30b/none/untouched} without, but training leaves \val{kin/nemotron-30b/gap} points of that there and \val{kin/qwen3-asr-1.7b/gap} on Qwen3-ASR 1.7B, and on Whisper the \val{kin/whisper-large-v3/none/n} languages with no relative on its list even end ahead. Family is a weak guide to what transfers.

\begin{table}[H]
\centering
\caption{Mean CER of the Latin-script languages by what the model's card says about them. A relative is another language of the same top-level family.}
\label{tab:kin}
\scriptsize
\input{table_kin}
\end{table}

\subsection{Languages no model lists}

\label{sec:nolist}

For \val{nl/n} of the 50, no model in the study starts with anything: \val{nl/names} are on no card and in neither MMS's nor omniASR's table, in any script. Nothing a model does there is mixed with what it knew of the language, so Figure~\ref{fig:nolist} repeats the main result on them. As downloaded every model is at \val{nl/before-min} to \val{nl/before-max}, the two systems built for the long tail included: omniASR CTC 1B, at \val{nl/omni-ctc-1b/rest-before} on the other \val{nl/rest-n} languages, is at \val{nl/omni-ctc-1b/before} on these. Fine-tuned, the models gain \val{nl/gain-min} to \val{nl/gain-max} points and end in nearly the order they end in on the other \val{nl/rest-n} (Spearman's $\rho$ \val{nl/rho}), the best at \val{nl/best/cer} (\val{nl/best/name}).

\begin{figure}[H]
\centering
\includegraphics[width=0.95\linewidth]{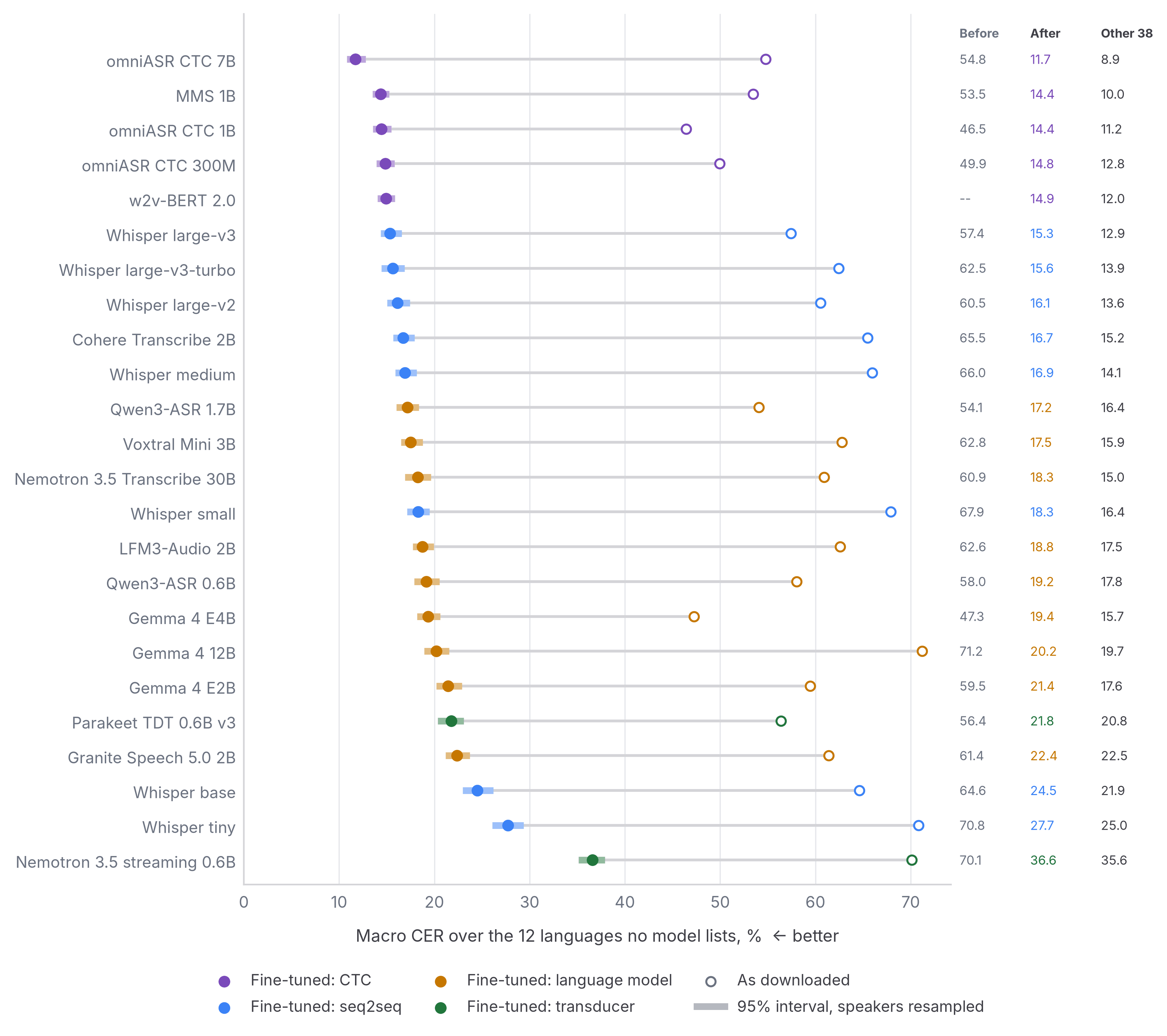}
\caption{The \val{nl/n} languages on no model's list, in any script: all \val{nl/n-models} models fine-tuned with the main recipe, as downloaded and at the epoch dev chose, macro CER over the \val{nl/n}. Other \val{nl/rest-n}: the fine-tuned model on the rest of the test set. MMS as downloaded has no head for these and uses English's; w2v-BERT has none at all.}
\label{fig:nolist}
\end{figure}

So the order of Section~\ref{sec:decoder} is not the lists' doing, though some of its margins are. Every CTC model is still ahead of every other model, but \val{dec/gap-ctc} and \val{dec/gap-other} are \val{nl/dec-gap} apart here (\val{nl/dec-gap-ci}), and MMS's lead of \val{nl/pair/omni-ctc-1b/mms-1b-all} over omniASR CTC 1B on all 50 is \val{nl/pair/omni-ctc-1b/mms-1b} (\val{nl/pair/omni-ctc-1b/mms-1b-ci}). The twelve are no easier than the rest: \val{nl/harder-n} of the \val{nl/n-models} models score worse on them, by a median of \val{nl/harder-median} points.

\subsection{One language or fifty}

\label{sec:single}

Most low-resource work trains one language at a time. Figure~\ref{fig:single} trains Qwen3-ASR 0.6B on each language alone, with the same clips and recipe: alone is better by \val{sl/diff} points on average over the \val{sl/n} languages, and only \val{sl/better-joint} do better with the other 49. Part of that is each language choosing its own epoch on its own 25 dev clips, but at the last epoch of both, where nothing was chosen, alone is still ahead, \val{sl/alone-last} against \val{sl/joint-last}, and the \val{sl/scarce-n} languages with fewer than 500 clips also do better alone, so the scarce ones do not lean on the others. A model of 0.6B parameters shared by fifty languages may have less of itself for each. Fifty models are fifty times the storage, though, and the joint model is one.

\begin{figure}[H]
\centering
\includegraphics[width=0.95\linewidth]{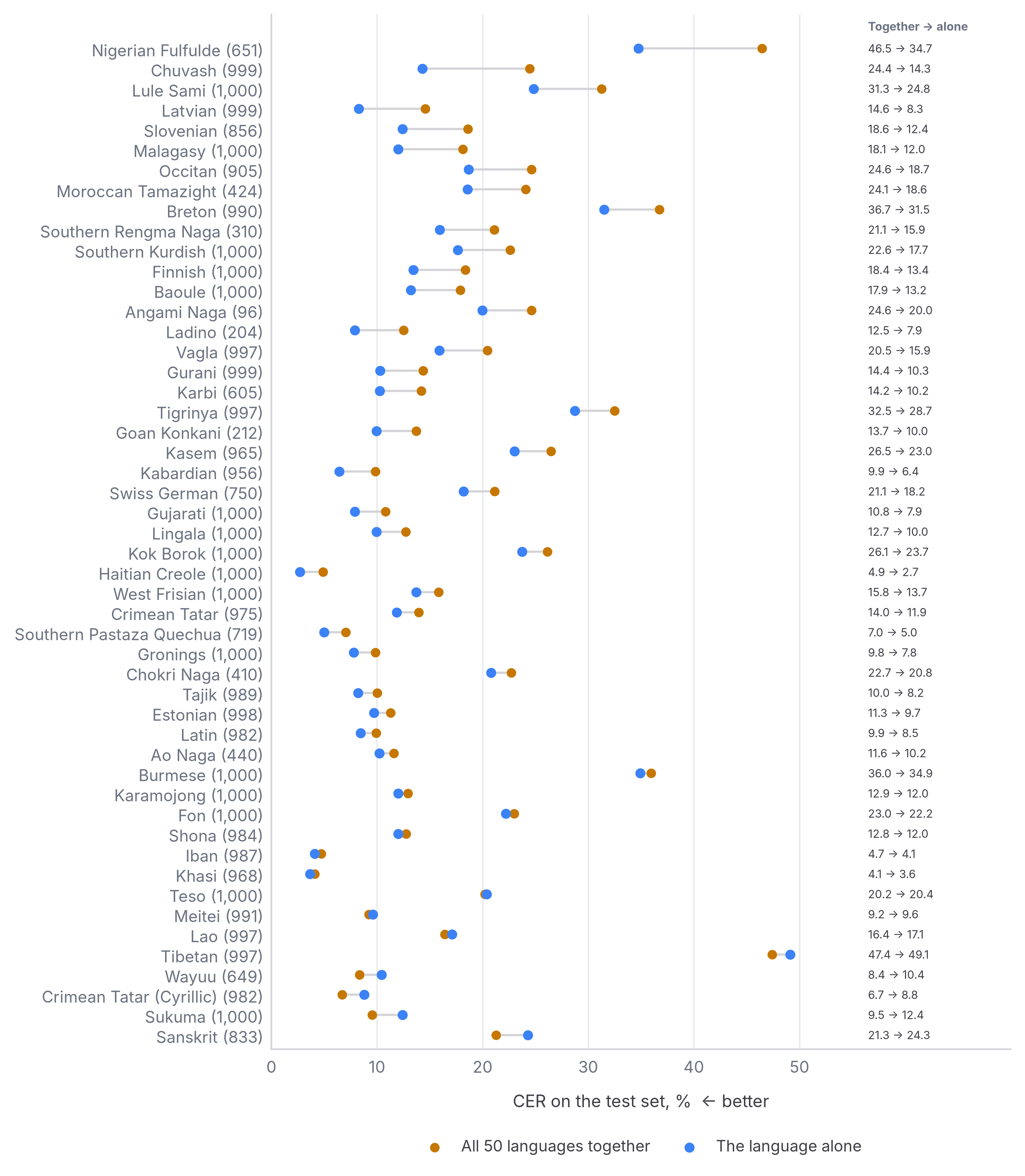}
\caption{Each language trained alone against all 50 trained together, Qwen3-ASR 0.6B, ten epochs, each at the epoch its own dev chose. In brackets, the language's training clips. Sorted by the difference: at the bottom, the languages that did better with the other 49.}
\label{fig:single}
\end{figure}

\subsection{The next language}

\label{sec:next}

Does a model that has learned forty low-resource languages learn a forty-first faster than the model it started from? We drew \val{ho/n-held} of the 50 at random, trained Qwen3-ASR 0.6B, Whisper small, Whisper large-v3-turbo and MMS on the other forty with the main recipe, and then trained each on the held-out ten with 25, 100 and 1,000 clips a language, from that first stage and from the pretrained model (Figure~\ref{fig:holdout}).

\begin{figure}[H]
\centering
\includegraphics[width=0.85\linewidth]{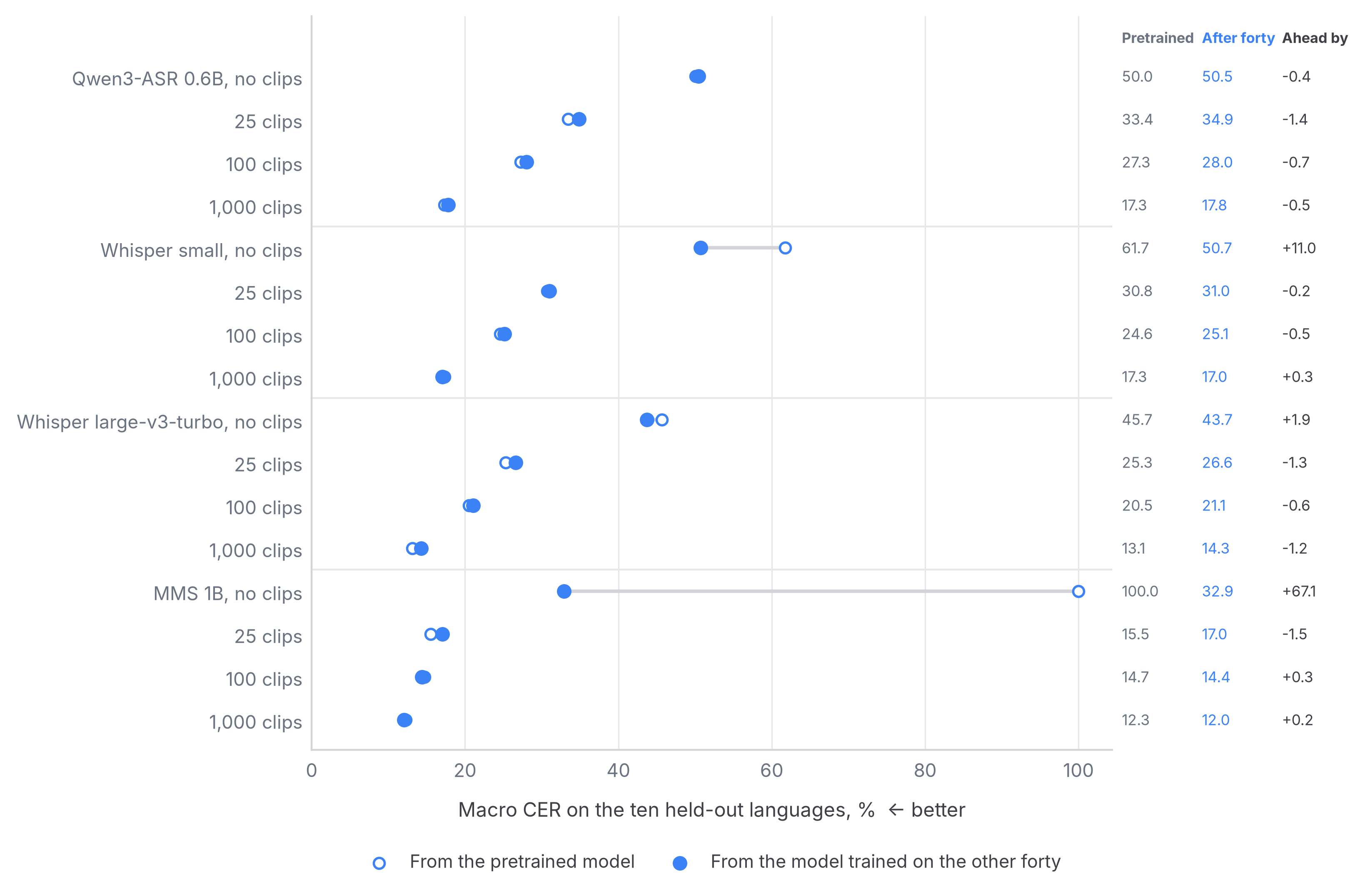}
\caption{Ten languages held out of a first stage on the other forty: the second stage from the pretrained model (hollow) and from the first stage (filled), by how many clips of each held-out language it trains on. No clips is the model before the second stage. Test macro CER on the ten at the epoch dev chose; the last column is how far the first stage puts the second ahead. One training seed; the text adds a second.}
\label{fig:holdout}
\end{figure}

Once there are clips to learn from, it does not. Before the second stage, the first puts Whisper small \val{ho/whisper-small/gain/0} points ahead on ten languages it has never heard and large-v3-turbo \val{ho/whisper-large-v3-turbo/gain/0}, and takes MMS, whose new head writes nothing until it is trained, from nothing to \val{ho/mms-1b/from40/0}; Qwen3-ASR does not move. With clips of the ten, the head start is gone, and with very few it turns into a small cost. Over two training seeds on the same clips, the first stage leaves the model \val{ho/all/k25-clause}; at 100 and 1,000 clips the \val{ho/all/more-n} differences go both ways, none larger than \val{ho/all/more-absmax} points, \val{ho/all/more-sig-behind-n} of them behind and \val{ho/all/more-sig-ahead-n} ahead with an interval that leaves out zero. Training the ten beside the other forty does not help either: Whisper small's main run, on all fifty at once, scores \val{ho/whisper-small/main} on the ten, and the ten alone at 1,000 clips \val{ho/whisper-small/pre/1000}; large-v3-turbo's, \val{ho/whisper-large-v3-turbo/main} and \val{ho/whisper-large-v3-turbo/pre/1000}.

\section{Epochs}

\label{sec:epochs}

Whether training longer helps depends on the decoder, and for most of them the answer comes early. Figure~\ref{fig:epochs} has dev after every epoch at a constant rate, and Table~\ref{tab:epochs} has the same runs on the test set.

On seq2seq and language-model decoders, the first epoch has \val{ep/first-share}\% of the whole gain in the median model and the third \val{ep/three-share}\%, and dev is within a point of its best by epoch \val{ep/within-median}. After that the last column of Table~\ref{tab:epochs} is flat for most of them, and \val{ep/d3-seq-sig-names} gain no more than \val{ep/d3-seq-sig-max} points. The CTC models and the transducers, which have new rows to learn, keep going: from epoch 3 to the last, w2v-BERT improves by \val{ep/w2v-bert/d3-last} points, MMS by \val{ep/mms-1b/d3-last}, omniASR CTC 1B by \val{ep/omni-ctc-1b/d3-last}, Parakeet by \val{ep/parakeet-tdt-v3/d3-last} and Nemotron streaming by \val{ep/nemotron-stream/d3-last}.

\begin{figure}[H]
\centering
\includegraphics[width=0.84\linewidth]{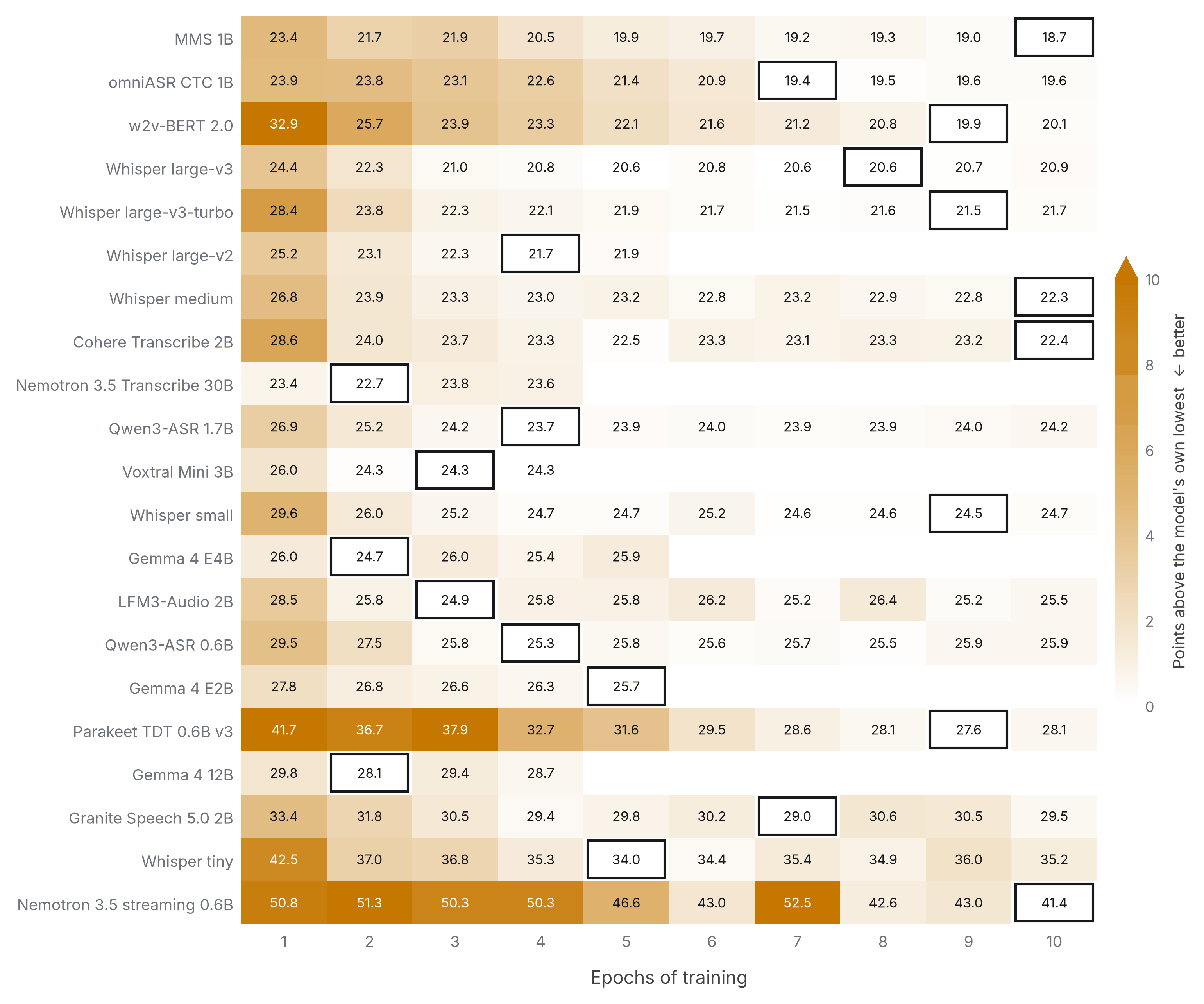}
\caption{Dev after every epoch at a constant learning rate, one row a model. Each cell is the dev macro CER; its colour is how far it is above that model's own lowest. A box marks the epoch dev chose; blank cells were not run.}
\label{fig:epochs}
\end{figure}

\begin{table}[H]
\centering
\caption{Test macro CER after 1, 2, 3, 5 and 10 epochs of the same runs, shown for the record: no epoch was chosen on it. Best: the epoch dev chose. Within 1 pt: the first epoch within one point of that best on dev. The last column is how much the run's last saved epoch improves on epoch 3 on the test set, with an interval from resampling the same speakers on both sides.}
\label{tab:epochs}
\scriptsize
\input{table_epochs}
\end{table}

To find where that stops we ran the longest curves again on another seed (Figure~\ref{fig:epochslong}). Thirty epochs of Whisper small, Whisper large-v3-turbo and Qwen3-ASR 0.6B are no better than the first ten (on turbo dev chooses epoch \val{ep30/whisper-large-v3-turbo/epoch}), though Whisper's training loss falls from \val{loss/ep30_whisper-small/3} at epoch 3 to \val{loss/ep30_whisper-small/20} at epoch 20: it is learning its training set by heart. In the CTC models' longer runs, w2v-BERT, whose head is new, goes on improving to epoch \val{ep30/w2v-bert/epoch} of \val{ep30/w2v-bert/last-epoch}, \val{ep30/w2v-bert/after10} points below its epoch 10, while omniASR CTC 1B, whose head came with the model, has flattened out, dev choosing epoch \val{ep20/omni-ctc-1b/epoch} by only \val{ep20/omni-ctc-1b/dev-after10} points over epoch 10. MMS, whose head is also new, is still improving at epoch \val{ep20/mms-1b/last-epoch}, the last it ran: dev chooses it, \val{ep20/mms-1b/dev-after10} points below epoch 10. Parakeet at 1e-4, copied at epoch 7 and carried on to twenty, ends \val{ep20/parakeet-tdt-v3/after10} points below its own epoch 10.

\begin{figure}[H]
\centering
\includegraphics[width=\linewidth]{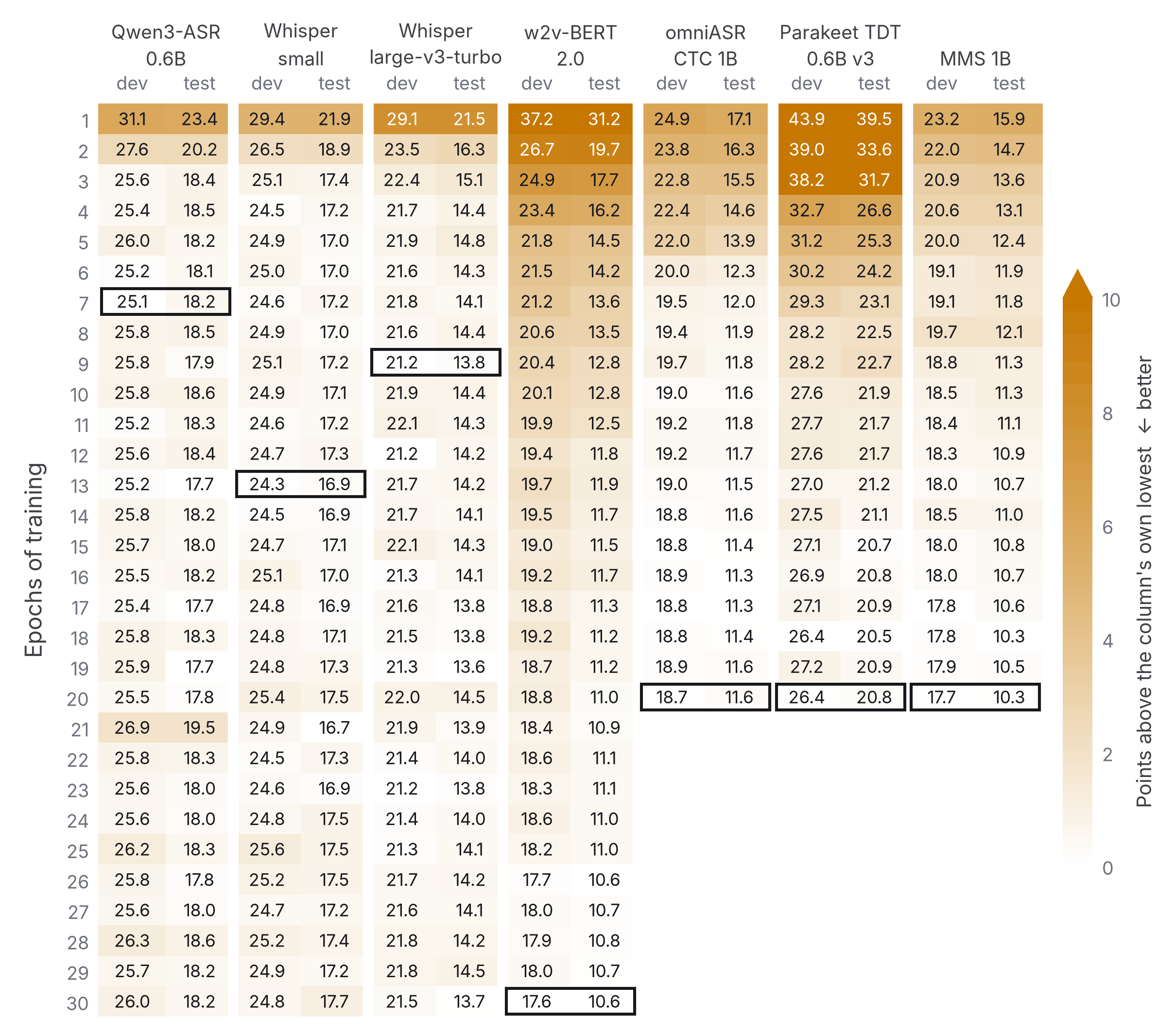}
\caption{The longest runs at a constant rate, on another seed, except Parakeet's, which carries its run at 1e-4 on from epoch 7. Dev and test macro CER after every epoch; the test columns are for the record. The colour is how far a cell is above its column's lowest. A box marks the epoch dev chose; blank cells were not run.}
\label{fig:epochslong}
\end{figure}

Scarce languages might need more passes, since they see fewer clips in each. Figure~\ref{fig:epochsgroups} says not many more: on dev the \val{gr/scarce/n} languages with under 500 clips come within a point of their best at epoch \val{gr/scarce/within}, the other two groups at \val{gr/mid/within} and \val{gr/full/within}.

\begin{figure}[H]
\centering
\includegraphics[width=0.75\linewidth]{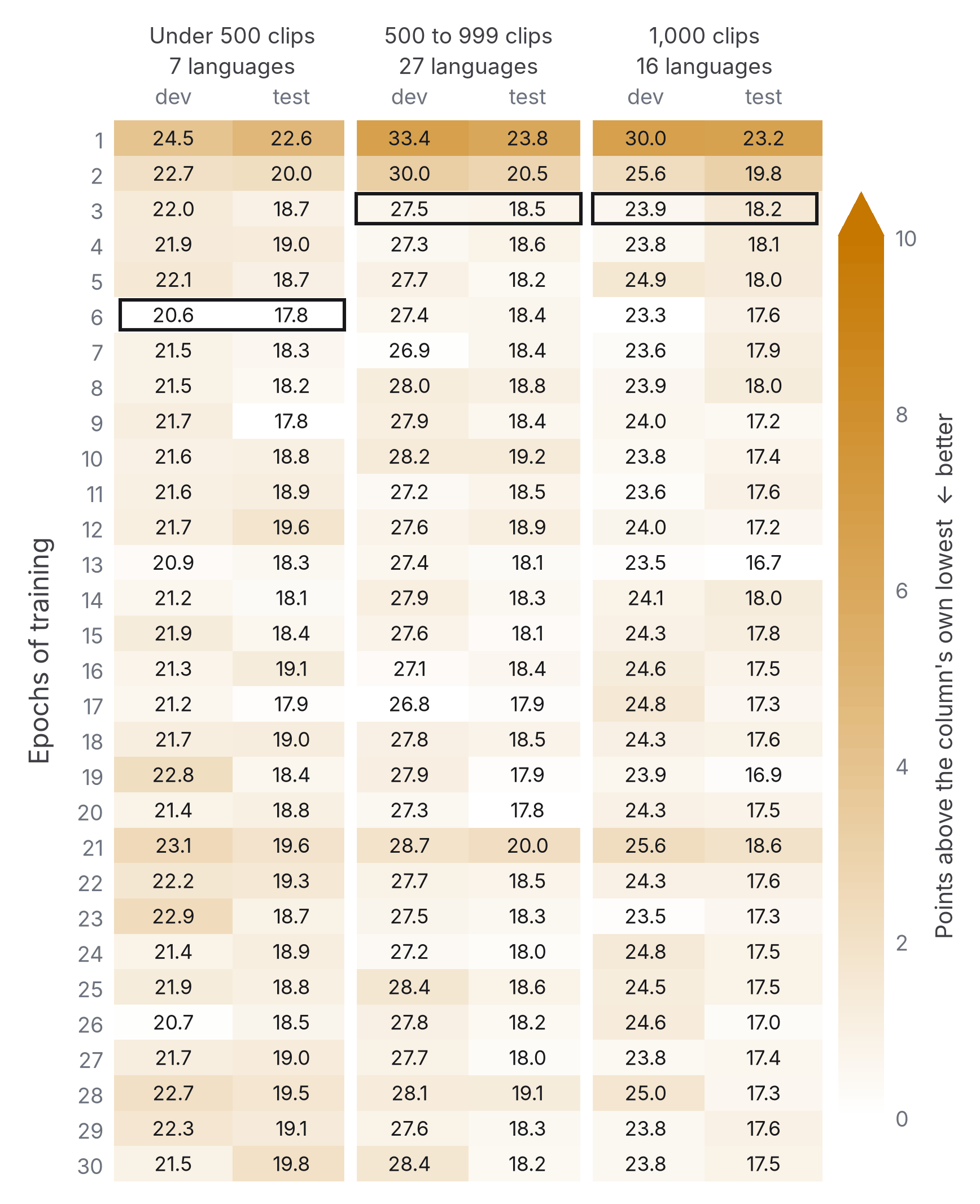}
\caption{The long run of Qwen3-ASR 0.6B, by how many clips a language has to train on: the mean dev and test CER of each group's languages after every epoch. The colour is how far a cell is above its column's lowest. A box marks the first epoch within a point of the group's lowest on dev.}
\label{fig:epochsgroups}
\end{figure}

A curve through one long run is not a training planned for that length, so we also ran the usual recipe at each length, the rate falling to zero at the last epoch (Figure~\ref{fig:schedule}). From three epochs on it beats the constant rate by \val{sch/qwen3-asr-0.6b/gain-min3} to \val{sch/qwen3-asr-0.6b/gain-max3} points on Qwen3-ASR 0.6B and \val{sch/whisper-small/gain-min3} to \val{sch/whisper-small/gain-max3} on Whisper small, where two seeds of the constant recipe lie up to \val{seed/qwen3-asr-0.6b/diff-max3} and \val{seed/whisper-small/diff-max3} points apart, and it hardly moves the answer about epochs: Qwen planned for three epochs scores \val{sch/qwen3-asr-0.6b/3/best} and planned for ten \val{sch/qwen3-asr-0.6b/10/best}. The two models still learning at ten epochs lose from the falling rate instead: Parakeet at every length, by more than its seeds differ except at three, and w2v-BERT until ten, where the cost is gone. That ten-epoch w2v-BERT run is on seed 1, because on seed 0 its new head stalled, writing the same output for every clip for \val{stall/stall_sch10_w2v-bert/stuck} epochs until we stopped it; the same command run again broke through within its first epoch, so the stall was chance, not the seed.

\begin{figure}[H]
\centering
\includegraphics[width=0.9\linewidth]{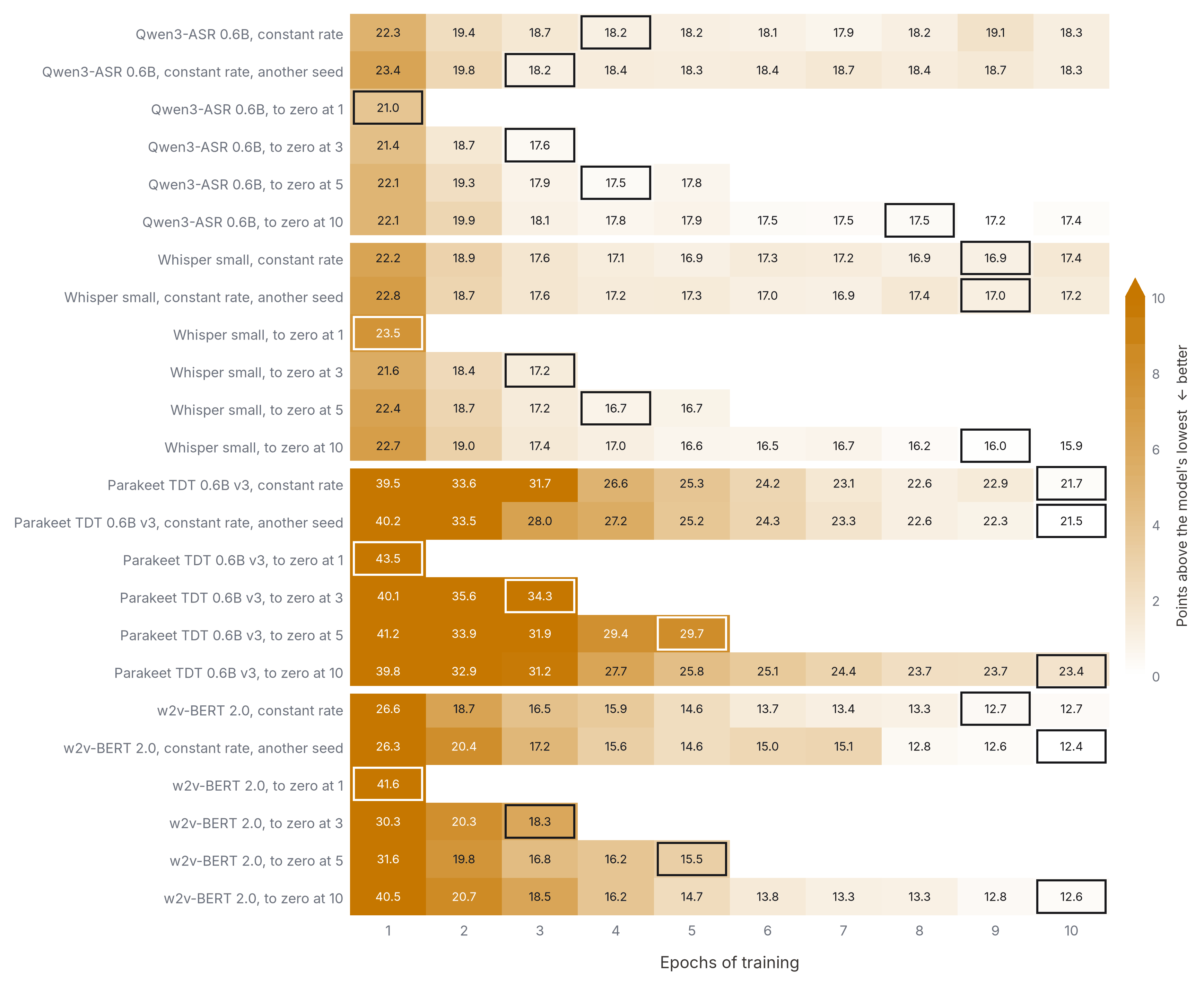}
\caption{Each model's main run at a constant rate, the same recipe on another seed, and separate runs planned for 1, 3, 5 and 10 epochs whose rate falls linearly to zero at their last. Test macro CER after every epoch, for the record; a box marks the epoch dev chose. The colour is how far a cell is above the lowest of that model's rows; blank cells were not run. w2v-BERT's ten-epoch run is on seed 1, after a stall on seed 0 (see text).}
\label{fig:schedule}
\end{figure}

\section{Data size}

\label{sec:scale}

A few clips already transfer (Table~\ref{tab:scale}, Figure~\ref{fig:scale}): twenty-five clips a language, about \val{data/k25-hours} hours over all 50, take Qwen3-ASR 0.6B from \val{bl/qwen3-asr-0.6b/cer} to \val{n/qwen3-asr-0.6b/25}, Whisper small from \val{bl/whisper-small/cer} to \val{n/whisper-small/25} and Whisper large-v3-turbo from \val{bl/whisper-large-v3-turbo/cer} to \val{n/whisper-large-v3-turbo/25}, and every doubling after that buys less. Parakeet TDT 0.6B v3 gets less from the first few clips, \val{n/parakeet-tdt-v3/25} after 25.

\begin{table}[H]
\centering
\caption{The data-size runs. Clips presented is clips times epochs. Macro CER on the test set at the epoch dev chose. A language with fewer clips than a row's size uses all it has.}
\label{tab:scale}
\scriptsize
\input{table_scale}
\end{table}

Data also beats epochs at the same cost. On Qwen3-ASR 0.6B a thousand clips a language for \val{n/qwen3-asr-0.6b/large-same-epoch} epochs are \val{n/qwen3-asr-0.6b/same-gain} points better than a hundred for \val{n/qwen3-asr-0.6b/small-epochs}, the same number of clips presented; Whisper small gains \val{n/whisper-small/same-gain} the same way, large-v3-turbo \val{n/whisper-large-v3-turbo/same-gain}, and Parakeet, which keeps improving for longer, \val{n/parakeet-tdt-v3/same-gain}. At the top of Table~\ref{tab:scale}, four thousand clips a language for three epochs present fewer clips than a thousand for ten and still beat them, by \val{n/qwen3-asr-0.6b/top-gain}, \val{n/whisper-small/top-gain}, \val{n/whisper-large-v3-turbo/top-gain} and \val{n/parakeet-tdt-v3/top-gain} points in the same order, each interval above zero. New clips are worth more than further passes over the old ones, and on every model but Parakeet time past the third epoch would be better spent collecting clips.

\begin{figure}[H]
\centering
\includegraphics[width=0.95\linewidth]{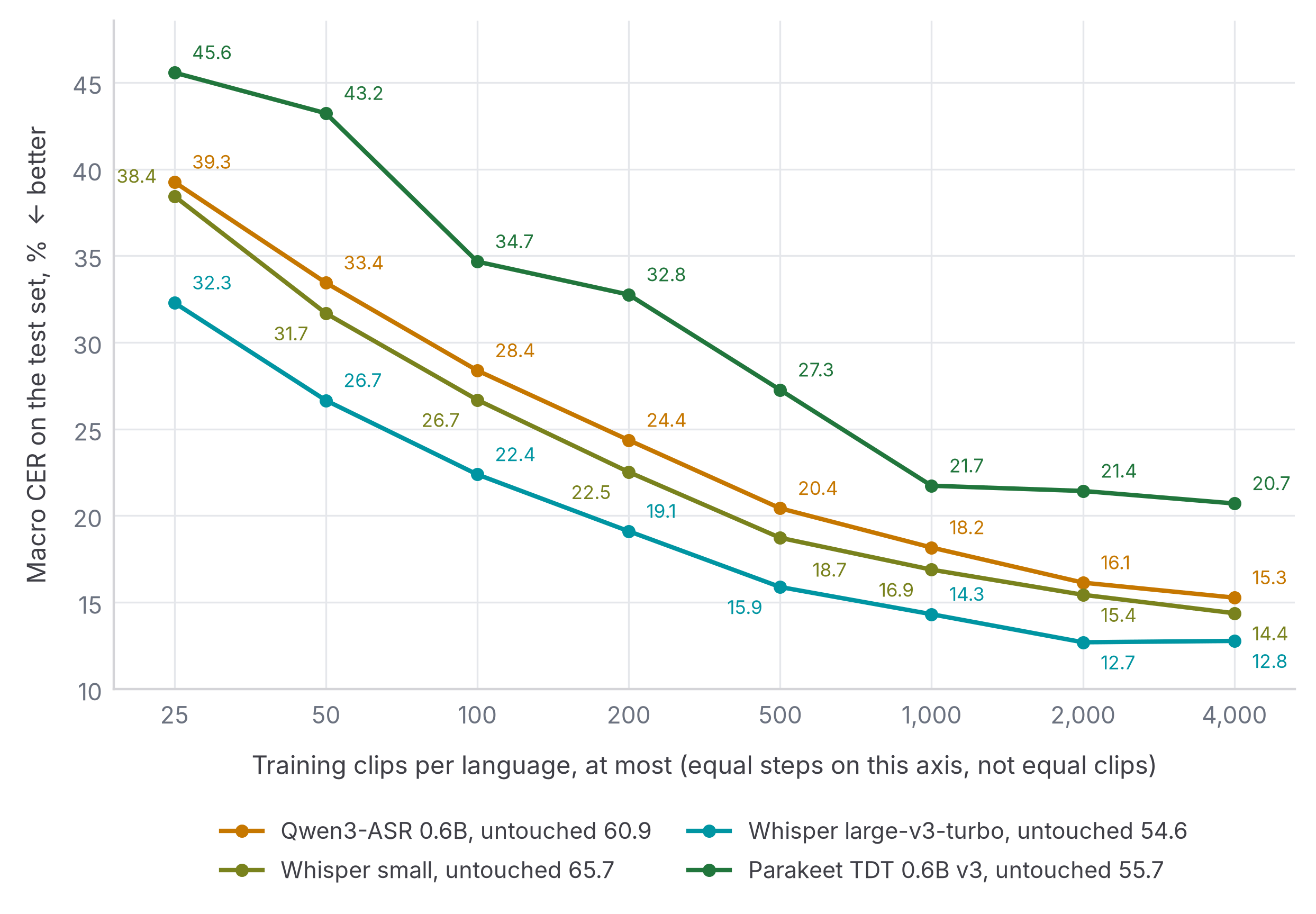}
\caption{Macro CER against the number of clips per language, at the epoch dev chose; the legend gives each model untouched. Smaller sets train for more epochs, and the two largest for fewer (Table~\ref{tab:scale}).}
\label{fig:scale}
\end{figure}

\section{Common Voice}

\label{sec:cv}

Common Voice 22 has at least 100 clips of \val{cv/n} of our languages, and Figure~\ref{fig:cv} trains on them three ways: on this corpus, on Common Voice, and on both.

We took at most \val{cv/cap} a language, leaving out any sentence in our dev or test set, \val{cv/cv-clips} clips and \val{cv/cv-hours} hours against this corpus's \val{cv/main-clips} and \val{cv/main-hours} of the same languages, and trained Qwen3-ASR 0.6B with the usual recipe. On Common Voice alone, with three times the clips, it ends \val{cv/only-loss} points behind this corpus alone, partly a change of kind, since our test clips come from the same corpus as our training clips: Common Voice alone is better on \val{cv/big-only-better} of the \val{cv/big-n} languages where it has more clips than this corpus and far worse where it has fewer. Together they beat either, \val{cv/both-gain} points below this corpus alone (\val{cv/both-gain-ci}), and Whisper small and Parakeet TDT 0.6B v3, trained the same three ways, agree. Where Common Voice has a language, it belongs in the training set.

The run on this corpus alone also checks Section~\ref{sec:single} at a larger scale: trained on these \val{cv/n} languages only, Qwen reaches \val{cv/cv_main/cer} on them, \val{cv/ft50-gain} points better than the main run trained on all 50 (\val{cv/ft50-gain-ci}; Figure~\ref{fig:cv}, purple), and Whisper small does \val{cv/whisper-small/ft50-gain} better. On Parakeet, still learning at ten epochs (Section~\ref{sec:epochs}), there is no difference. Fewer languages sharing a model again do better on each.

\begin{figure}[H]
\centering
\includegraphics[width=0.95\linewidth]{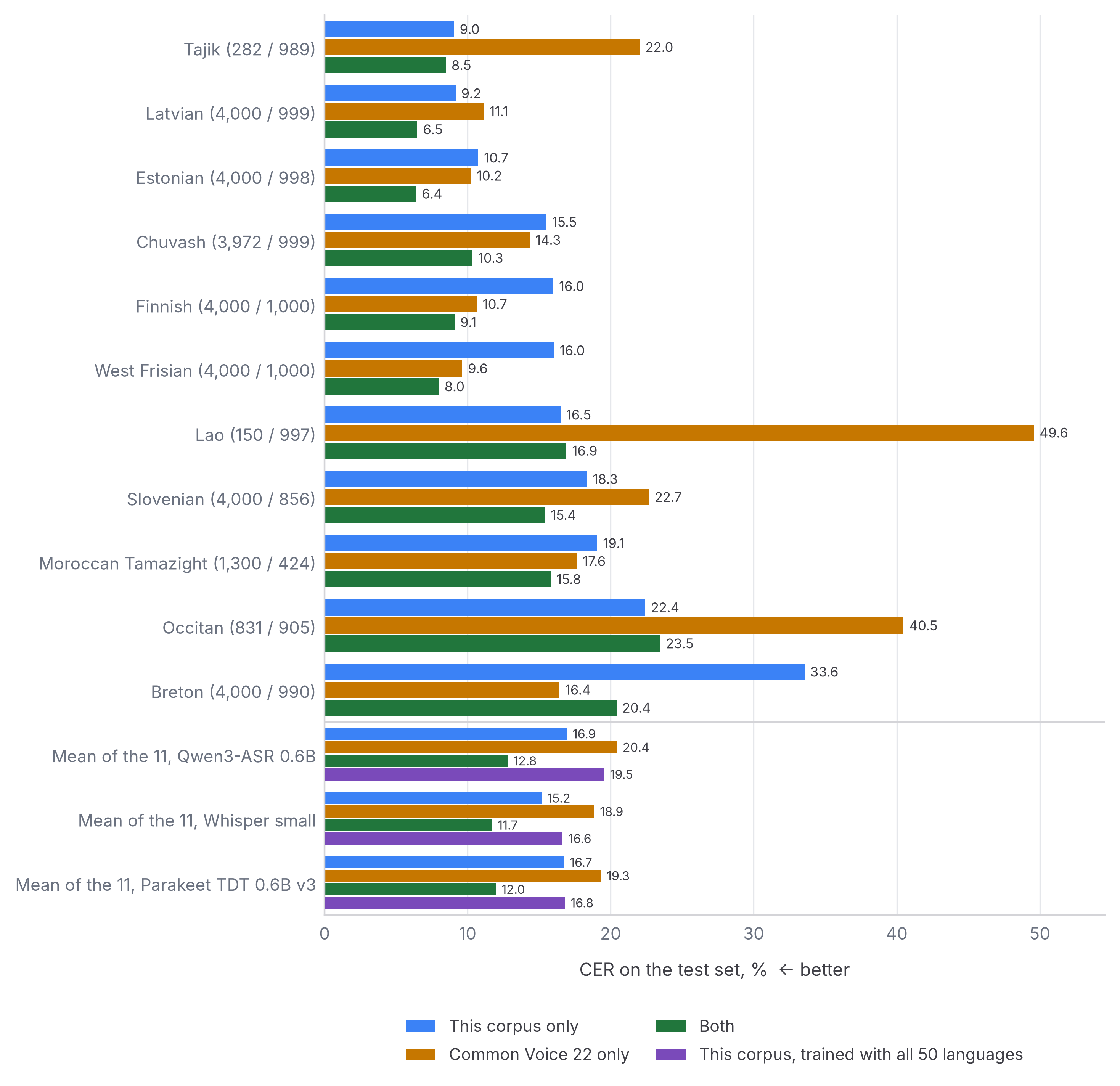}
\caption{The languages Common Voice 22 has at least 100 clips of, trained on this study's corpus, on Common Voice, and on both. Test CER at the epoch dev chose: each language on Qwen3-ASR 0.6B, then the mean of the same three runs on each model, beside the model trained on all 50 languages with this corpus's clips and scored on the same \val{cv/n} (purple). In brackets, the language's clips in Common Voice (at most \val{cv/cap} taken) and in this corpus (at most 1,000).}
\label{fig:cv}
\end{figure}

\section{Another corpus}

\label{sec:ooc}

Every test clip so far comes from the corpus that also gave the training clips, so a model could learn the corpus rather than the language. FLEURS \citep{conneau2022fleurs} records \val{ooc/n-shared} of our languages, but for \val{ooc/n-from-fleurs} of them the corpus took clips from FLEURS itself: each of our clips that reads a FLEURS sentence is as long as a FLEURS recording of that sentence, to within 0.02 seconds, and \val{ooc/fleurs-test-in-train-max} of \val{ooc/per-lang} FLEURS test clips of \val{ooc/fleurs-test-in-train-max-name} read a sentence our training clips read. That leaves \val{ooc/clean-names}, where FLEURS is another corpus: other texts, other speakers, other microphones. We score every model there on \val{ooc/per-lang} FLEURS test clips a language, as downloaded and from the epoch dev chose in its main run, beside the same checkpoint on our own test clips of those languages (Figure~\ref{fig:ooc}).

Most of the gain is the language. On FLEURS the median model keeps \val{ooc/keep-median}\% of what fine-tuning gained it on our clips, \val{ooc/keep-min}\% at least (among the \val{ooc/keep-n} that gained \val{ooc/keep-min-gain} points or more there), and FLEURS ranks the fine-tuned models as our test does (Spearman's $\rho$ \val{ooc/rank-rho}). The \val{ooc/worse-word} that end worse on FLEURS than they started, \val{ooc/worse-names}, are the ones that knew these languages already, at \val{ooc/knew-before-max} or better on our clips as downloaded: with little to gain there, fine-tuning on fifty languages trades some of it away.

\begin{figure}[H]
\centering
\includegraphics[width=0.86\linewidth]{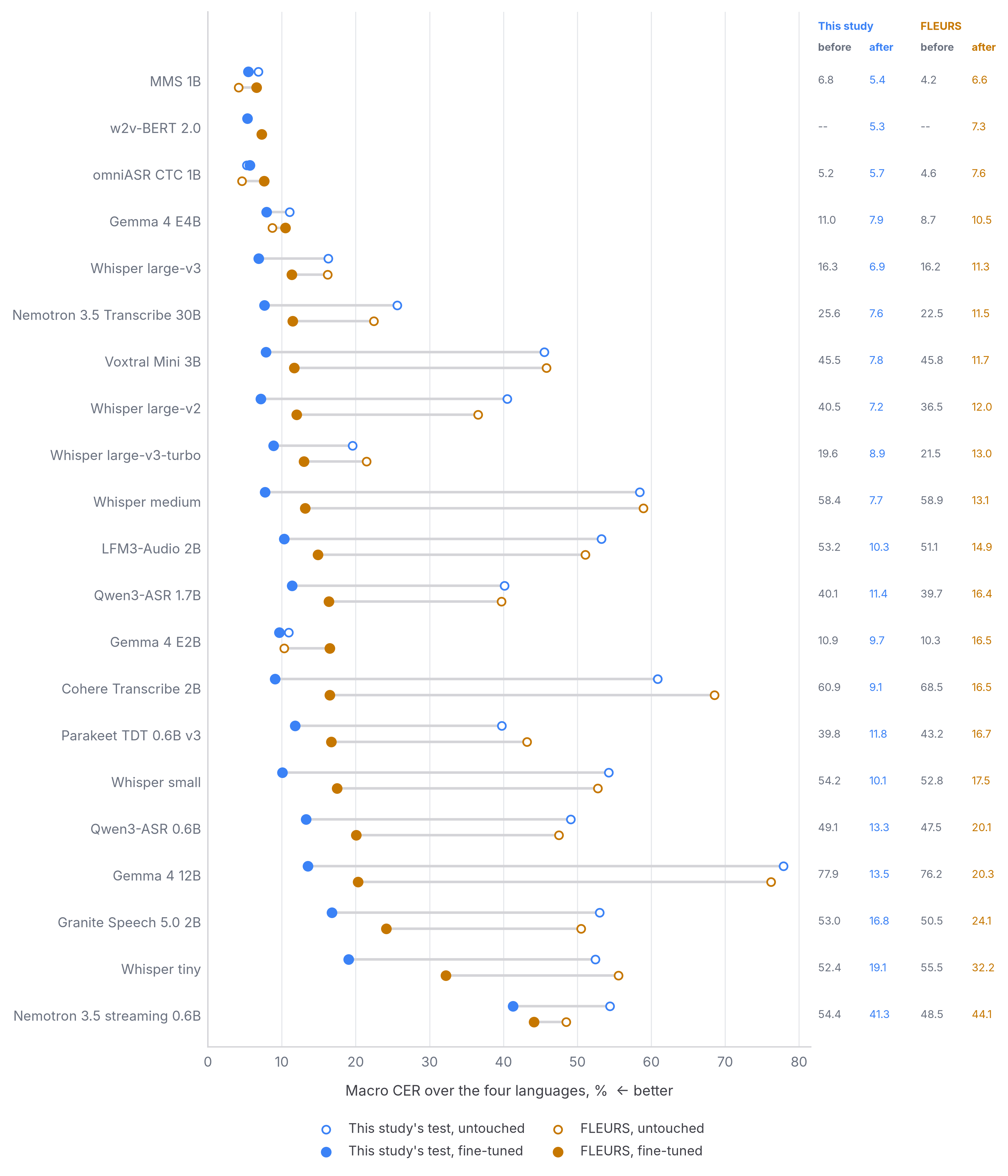}
\caption{\val{ooc/clean-names}, the languages FLEURS shares with this study whose clips here are not FLEURS recordings: as downloaded and fine-tuned, on this study's test clips and on FLEURS's. Each fine-tuned model is one checkpoint, scored on both; w2v-BERT has no head as downloaded.}
\label{fig:ooc}
\end{figure}

\subsection{The languages they knew}

\label{sec:forget}

Learning fifty new languages costs a model some of the ones it knew (Figure~\ref{fig:forget}). On English, Spanish, French and German from FLEURS, the median model's CER rises by \val{fk/loss-median} points, and by \val{fk/loss-max} for \val{fk/worst/name}, which goes from \val{fk/nemotron-stream/before} to \val{fk/nemotron-stream/after}. All \val{fk/all-worse-n} end worse, \val{fk/least-worse/name} the least, by \val{fk/least-worse/loss} (\val{fk/least-worse/loss-ci}); its first run, which kept the experts as downloaded, even ended better, from \val{fk/nemotron-30b-first/before} to \val{fk/nemotron-30b-first/after}. Adapters, which leave the pretrained weights as they were, lose less on \val{fk/lora/less-n} of the \val{fk/lora/n} models we tried them on (Section~\ref{sec:peft}).

\begin{figure}[H]
\centering
\includegraphics[width=0.9\linewidth]{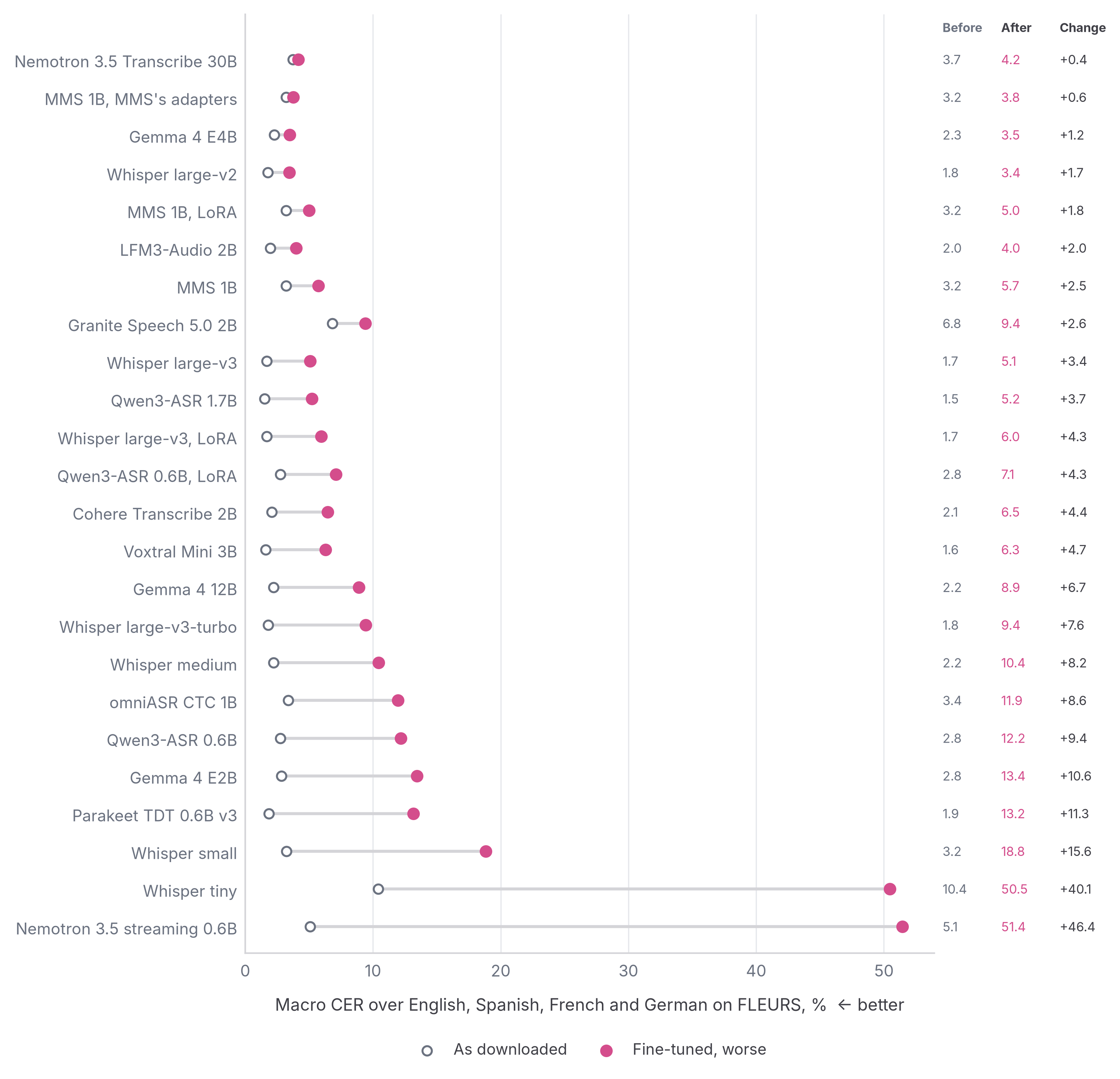}
\caption{English, Spanish, French and German on FLEURS, as downloaded and after fine-tuning on the fifty languages, with the adapters of Section~\ref{sec:peft} where they ran.}
\label{fig:forget}
\end{figure}

\subsection{Switching languages}

\label{sec:cs}

Every clip so far is in one language, and speech often mixes two. CS-FLEURS \citep{yan2025csfleurs} rewrites FLEURS sentences with English words and phrases mixed in and voices each language's stretch with MMS's TTS; five of its pairs are languages of ours with English. We score all \val{cs/n} models on 200 clips of each, told the language that is not English, beside the same five languages spoken alone on FLEURS (Figure~\ref{fig:cs}). The voices are synthetic and the switches are joins, so this asks whether a model follows a switch, not whether it hears one speaker make it, and the Latvian, Tajik and Burmese sentences come from FLEURS, which our corpus partly repackages (Section~\ref{sec:ooc}).

\begin{figure}[H]
\centering
\includegraphics[width=\linewidth]{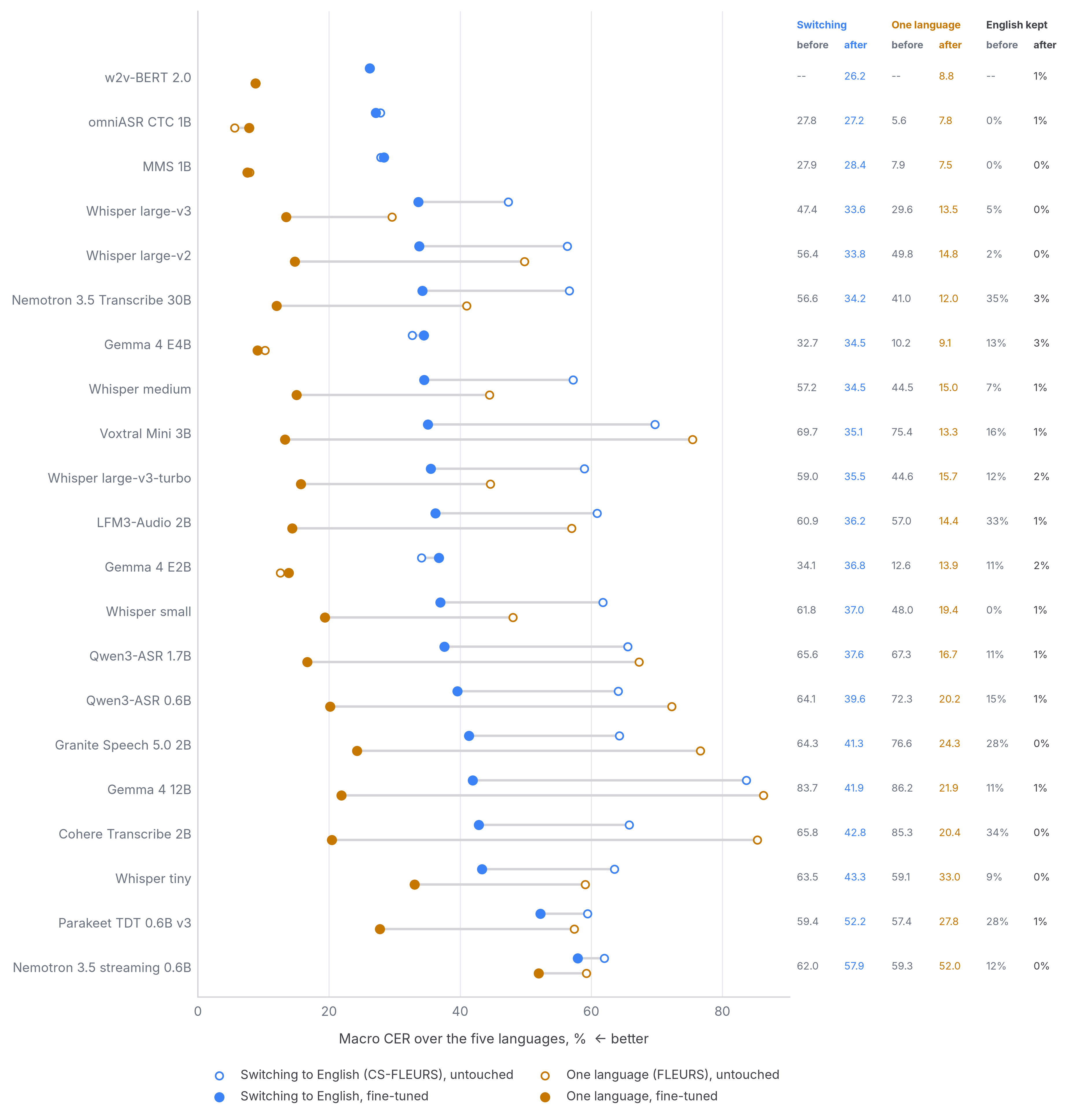}
\caption{Five of the study's languages with English mixed in (CS-FLEURS) and the same languages alone (FLEURS), as downloaded and fine-tuned: macro CER over the five. English kept: of the English words in the Gujarati, Burmese and Tajik clips, where English has a script of its own, the share the output writes.}
\label{fig:cs}
\end{figure}

Fine-tuning helps with mixed speech too: \val{cs/better-n} of the \val{cs/with-before-n} models we can compare end lower, by a median of \val{cs/gain-median} points. It stays far harder than one language, though: the median fine-tuned model is at \val{cs/after-median} on the mixed clips against \val{cs/mono-after-median} on the same languages alone. Much of the gap is the English. In Gujarati, Burmese and Tajik, where English is \val{cs/en-chars-min} to \val{cs/en-chars-max}\% of a reference's characters, the median model writes \val{cs/en-before-median}\% of the English words as downloaded and \val{cs/en-after-median}\% fine-tuned: told the other language, it writes the English in that language's script or drops it.

Malay with English is a switch the Malaysian arm trains on: in the Scicom telephony dataset of Section~\ref{sec:malay}, \val{mycs/train-en-share}\% of the training words are English and \val{mycs/train-mixed-share}\% of the clips have two or more. On CS-FLEURS's Malay pair, against the same model as downloaded and told Malay, that training helps only \val{csms/better-n} of the \val{csms/n} models, by a median of \val{csms/gain-median} points. The public Malay benchmark has switching of its own: \val{mycs/mixed} of its \val{mycs/n} clips have two or more English words, \val{mycs/top-cat-n} of them among its \val{mycs/top-cat-of} telephony clips. Within each category those clips gain about as much from the training as the clips with none (median difference \val{mycs/strat-extra-median} points), so what the training adds there is the domain, not the switching.

\section{Adapters and a language model}

\label{sec:peft}

Two cheap additions to the main recipe: training a small part of the model instead of all of it, and decoding a CTC model with a word language model.

\subsection{Adapters}

Tuning every weight means a full copy of the model for every set of new languages. Low-rank adapters of rank 32 on every attention and feed-forward projection \citep{hu2021lora} train instead, the rest frozen except what the new languages add (Whisper's new token rows, MMS's new CTC head), and for MMS also its own adapters, one small one in every layer, with a head \citep{pratap2024scaling}. MMS keeps one such set for each language and swaps in the set of the language a user names. We train it both ways: one set for all 50 languages, as in every other arm, and one set for each language, trained on that language alone and swapped in for it at test, which needs the language named. Each arm has its own rate sweep and then trains ten epochs on the main data (Table~\ref{tab:adapters}). The sets for one language train longer, since a language alone gets a fiftieth of the updates: each sees 20,000 clips, so a language with fewer than 1,000 trains for more epochs, up to \val{ad/mms-1b/mmsadl/longer-max} for the scarcest.

\begin{table}[H]
\centering
\caption{Tuning every weight against adapters on a frozen model, ten epochs on the same data: the parameters trained, the rate the model's own sweep chose, and test macro CER at the epoch dev chose, with a 95\% interval from resampling speakers. MMS's adapters one set a language: one language's set, trained on 20,000 clips of that language and named at test.}
\label{tab:adapters}
\scriptsize
\input{table_adapters}
\end{table}

LoRA comes close to tuning everything. On MMS, training \val{ad/mms-1b/lora/share}\% of the weights costs \val{ad/mms-1b/lora/cost} points of CER (\val{ad/mms-1b/lora/cost-ci}), and MMS's adapters, one set for all 50 languages and \val{ad/mms-1b/mmsad/share}\% of the weights, cost \val{ad/mms-1b/mmsad/cost}. MMS's own way, one set for each language, does worse still, \val{ad/mms-1b/mmsadl/behind-shared} points behind one set for all (\val{ad/mms-1b/mmsadl/behind-shared-ci}) although each is told its language, and \val{ad/mms-1b/mmsadl/cer-e10} after the same ten epochs as the shared set. On the \val{an/mms/n} languages MMS already had, the sets trained here reach \val{ad/mms-1b/mmsadl/on-own}, against \val{ad/mms-1b/mmsadl/own} for its own; what MMS gains from a set of its own per language is the languages it knew, which keep theirs untouched. On Qwen3-ASR 0.6B LoRA is level with tuning everything, \val{ad/qwen3-asr-0.6b/lora/cost} points behind (\val{ad/qwen3-asr-0.6b/lora/cost-ci}), and on Whisper large-v3 it costs \val{ad/whisper-large-v3/lora/cost} (\val{ad/whisper-large-v3/lora/cost-ci}). On two of the three, what they buy is the languages the model knew. On English, Spanish, French and German, Qwen3-ASR tuned in full loses \val{fk/qwen3-asr-0.6b/loss} points of CER and with LoRA \val{fk/qwen3-asr-0.6b/lora/loss}; MMS loses \val{fk/mms-1b/loss} tuned in full, \val{fk/mms-1b/lora/loss} with LoRA and \val{fk/mms-1b/mms/loss} with one set of its adapters for all 50. \val{fk/lora/more-names} loses more with LoRA, \val{fk/whisper-large-v3/lora/loss} against \val{fk/whisper-large-v3/loss} tuned in full (Figure~\ref{fig:forget}).

\subsection{A language model}

A CTC model writes each frame without knowing which words exist. We searched the three CTC models' output with a word 4-gram model of each language \citep{heafield2011kenlm}, built from the transcripts of the whole training pool, up to 4,000 clips a language where the models heard at most 1,000, less every sentence that dev or test also has (median \val{lm/sentences-median} sentences and \val{lm/words-median} words a language, the fewest \val{lm/sentences-min}). The beam is 64 wide \citep{kensho2021pyctcdecode}, and one weight for the language model and one bonus per word serve every language, chosen on dev (Table~\ref{tab:lm}).

\begin{table}[H]
\centering
\caption{The CTC models decoded greedily and with a word 4-gram, from the same saved output of the epoch dev chose. The interval of the WER gain resamples the same speakers on both sides; the weight and bonus were chosen on dev.}
\label{tab:lm}
\scriptsize
\input{table_lm}
\end{table}

It helps all three by about the same: \val{lm/cer-gain-min} to \val{lm/cer-gain-max} points of CER and \val{lm/wer-gain-min} to \val{lm/wer-gain-max} of WER, every interval clear of zero, and dev chose the same weight for each. With it \val{lm/best/name} reaches \val{lm/best/cer}, the lowest macro CER on our test set, yet its WER is still \val{lm/best/wer}: in the median language \val{lm/oov-median}\% of the test words never occur in the text the language model was built from, and up to \val{lm/oov-max}\% in one.

\section{Malaysian}

\label{sec:malay}

Malay is not a language with little to train on, but the question of what a model brings to it is the same, and one thing is different: the training data and the test data do not come from the same kind of speech. We train on the Scicom telephony dataset, \val{my/train-clips} private clips and \val{my/train-hours} hours of human-transcribed call-centre speech in Malay and English with some Mandarin, cut at the speakers' pauses, and test on the public split of the Revolab ASR Benchmark \citep{revolab2026}: \val{my/bench-clips} clips and \val{my/bench-hours} hours of Malaysian Malay and English in \val{my/bench-cats} categories, from telephone calls to singing, each counting once as each language did. Dev is \val{my/dev-clips} held-out clips of the telephony dataset and alone chooses the epoch; each model trains at its main run's rate, w2v-BERT excepted (below). As the benchmark's card asks, listed word variants count as correct, and a number may be spelled out or written in digits.

A model with Malay on its list can be run told the language or left to detect it; the two differ enough that Figure~\ref{fig:malay} and Table~\ref{tab:malay} have both, and a gain is counted from the better. Trained on telephone calls alone, Whisper large-v3 goes from \val{my/whisper-large-v3/base} to \val{my/whisper-large-v3/after}, a gain of \val{my/whisper-large-v3/gain} points (\val{my/whisper-large-v3/gain-ci}) and the lowest score of any model. Of the \val{my/n} models, \val{my/up-n} gain half a point or more and \val{my/down-names} \val{my/down-verb} worse by as much; a model that starts well has little left for these calls to give. w2v-BERT's new CTC head stalls at its main rate, writing the same thing for every clip for \val{mytry/my_w2v-bert/stuck} and \val{mytry/my3_w2v-bert/stuck} epochs on two seeds, while at a third of the rate it is out by epoch \val{mytry/mylow_w2v-bert/out} and ends \val{my/w2v-bert/rank-word}, at \val{my/w2v-bert/after}. MMS's new head beats the Malay head it came with, \val{my/mms-1b-all/after} against \val{my/mms-1b-all/told}.

\begin{figure}[H]
\centering
\includegraphics[width=0.9\linewidth]{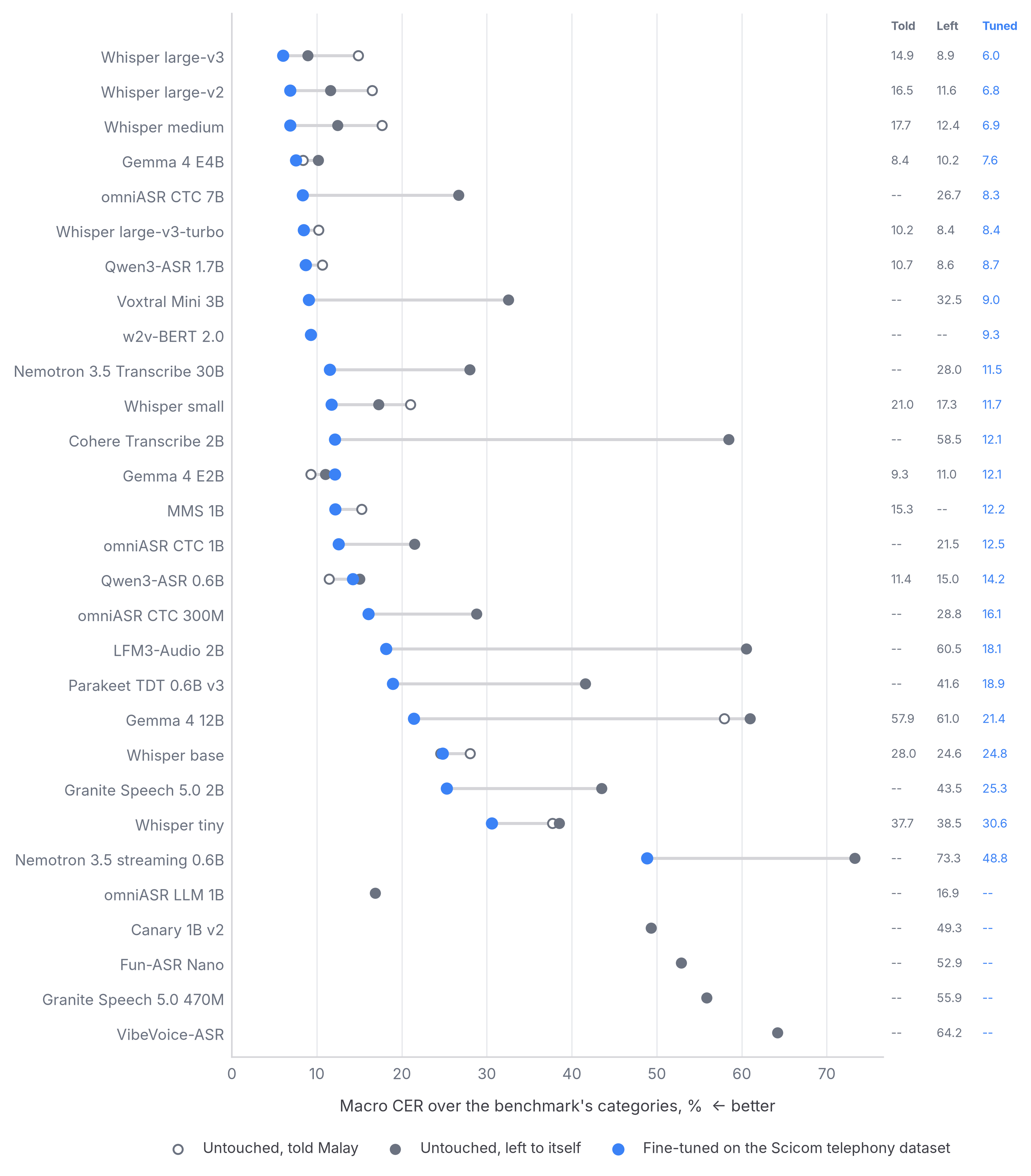}
\caption{Every model on the public Malay benchmark: as downloaded, told Malay where its list has it and left to itself, and after training on the Scicom telephony dataset. Macro over the benchmark's categories. Dashes were not run.}
\label{fig:malay}
\end{figure}

Most of the gap between the two untouched columns is translation (Figure~\ref{fig:malaygain}): told that a clip is Malay, an untouched Whisper translates the English in it. Of the \val{my/en-clips} clips whose reference is mostly English by a count of common words, large-v3 told Malay answers \val{my/whisper-large-v3/en-as-ms} in Malay, and they hold \val{my/whisper-large-v3/en-as-ms-share}\% of its errors; left to detect the language it does so on \val{my/whisper-large-v3/detect-en-as-ms}, and fine-tuned, still told Malay, on \val{my/whisper-large-v3/en-as-ms-after-word}. Counted from the model told Malay, its gain would read \val{my/whisper-large-v3/told-gain} points instead of \val{my/whisper-large-v3/gain}. Detection is not free either: Qwen3-ASR 0.6B is better told, \val{my/qwen3-asr-0.6b/told} against \val{my/qwen3-asr-0.6b/left}.

\begin{table}[H]
\centering
\caption{The Malaysian arm, best first. The test set is the Revolab benchmark, macro over its categories, capped per clip. MMS untouched is told Malay through its Malay adapter and head. The last columns, WER and the benchmark's telephony category, go from the better untouched score to the fine-tuned one.}
\label{tab:malay}
\scriptsize
\input{table_malay}
\end{table}

Beyond translation, the calls help large-v3 most on short inputs, parliament, telephony and street interviews, and cost it a little on singing, news and FLEURS (Figure~\ref{fig:malaygain}; every model is in Figure~\ref{fig:malaycats}). They do not even win the benchmark's own calls: the best fine-tuned telephony score is \val{my/tel/tuned} (\val{my/tel/tuned-name}), and \val{my/tel/untouched-name} scores \val{my/tel/untouched} there untouched. These are not the same calls, and a model tuned on one call centre writes what that call centre's transcribers write.

\begin{figure}[H]
\centering
\includegraphics[width=0.9\linewidth]{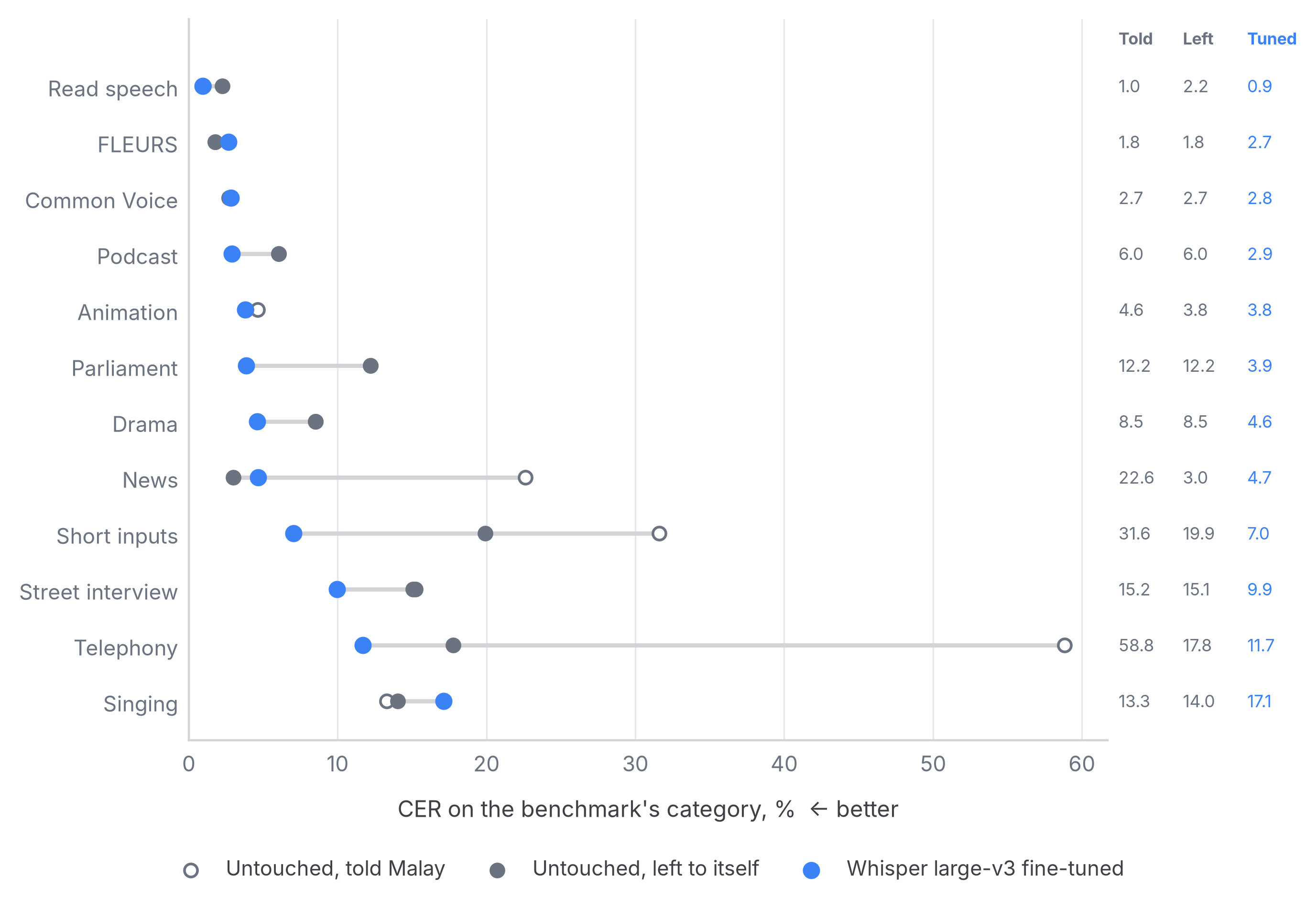}
\caption{The best fine-tuned model on each of the benchmark's categories: as downloaded both ways, and after training on telephone speech alone.}
\label{fig:malaygain}
\end{figure}

\begin{figure}[H]
\centering
\includegraphics[width=0.95\linewidth]{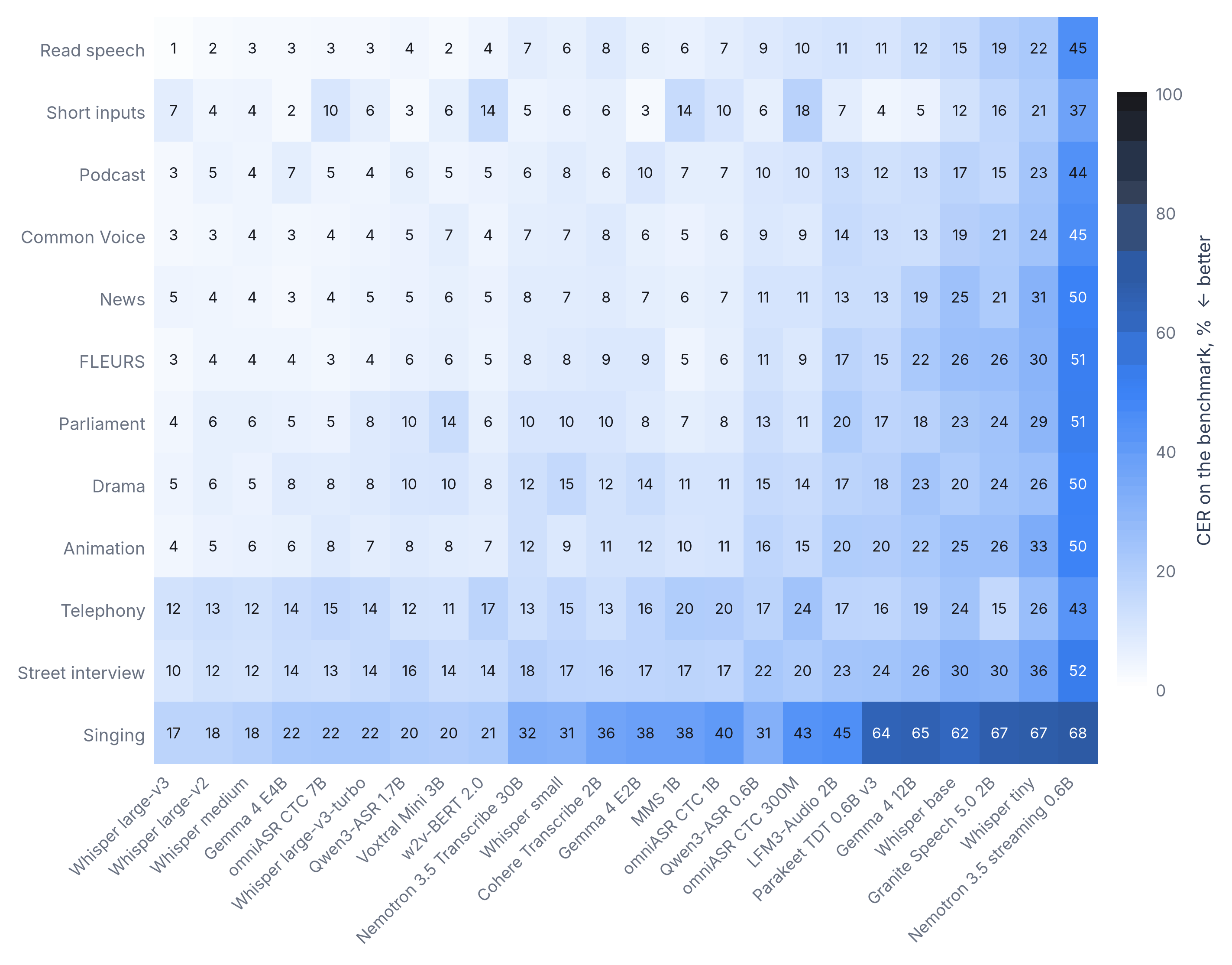}
\caption{Every fine-tuned model on every category of the benchmark.}
\label{fig:malaycats}
\end{figure}

More epochs can hurt here, and dev does not warn of it (Figure~\ref{fig:malaycurve}): \val{my/curve-early} of the \val{my/curve-n} models reach their lowest benchmark score within three epochs, and Whisper tiny climbs from \val{my/whisper-tiny/curve-low} at epoch \val{my/whisper-tiny/curve-low-epoch} to \val{my/whisper-tiny/curve-last} by epoch \val{my/whisper-tiny/curve-last-epoch} while its telephony dev improves from \val{my/whisper-tiny/curve-dev-first} to \val{my/whisper-tiny/curve-dev-last}. It is becoming a better recogniser of these calls and a worse one of everything else.

\begin{figure}[H]
\centering
\includegraphics[width=0.95\linewidth]{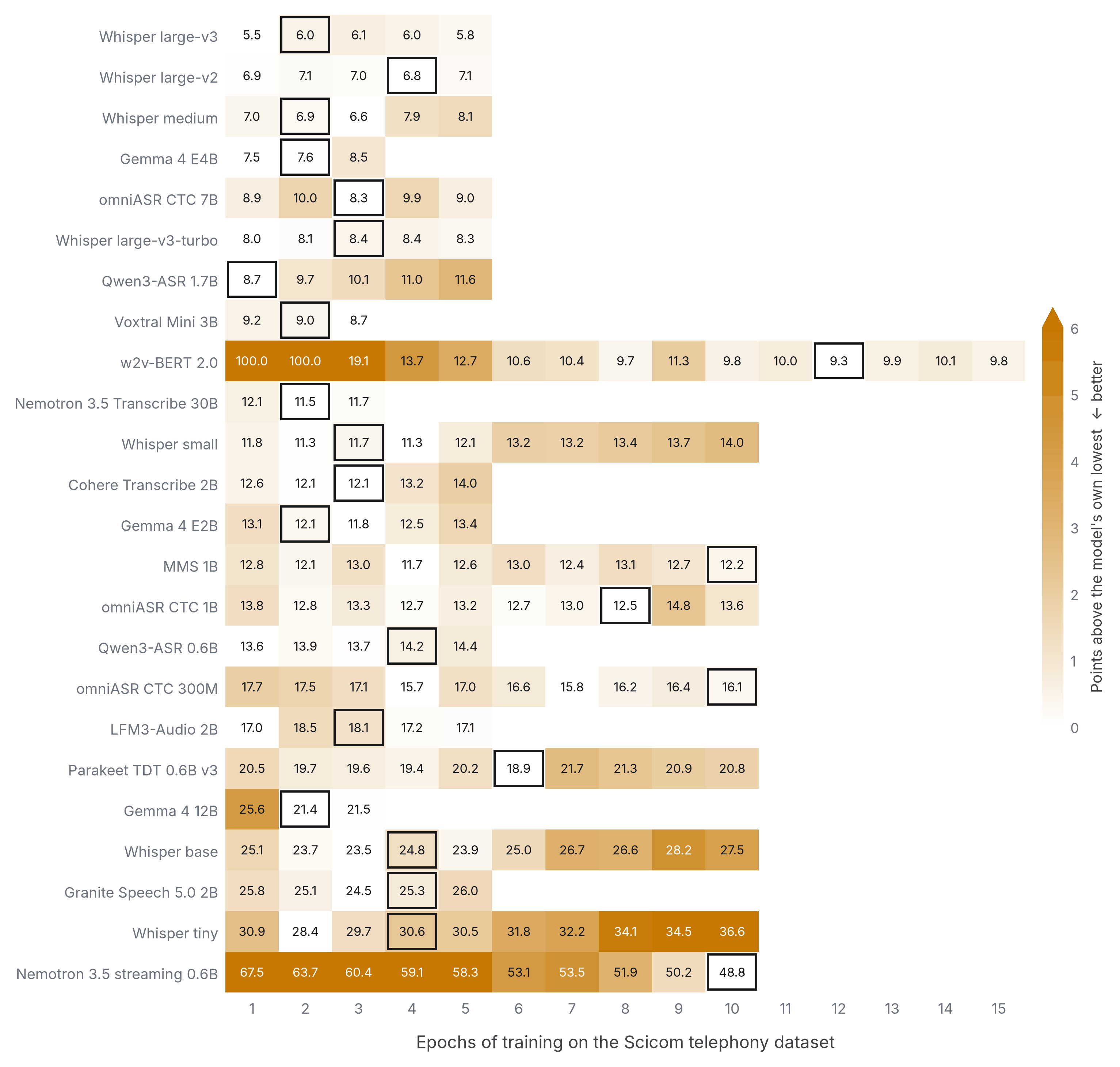}
\caption{The benchmark after every epoch of training on the Scicom telephony dataset, one row a model. The colour is how far a cell is above that model's own lowest. A box marks the epoch telephony dev chose. Blank cells were not run.}
\label{fig:malaycurve}
\end{figure}

Nor does more telephony help the benchmark (Figure~\ref{fig:malaysize}): Whisper small ends at \val{my/size/whisper-small/1000} after a thousand clips, \val{my/size/1000/hours} hours, and at \val{my/size/whisper-small/all} after all \val{my/size/all/hours}, while telephony dev improves the whole way. Whatever the benchmark gains from telephone calls, a small sample of them already has it.

\begin{figure}[H]
\centering
\includegraphics[width=0.84\linewidth]{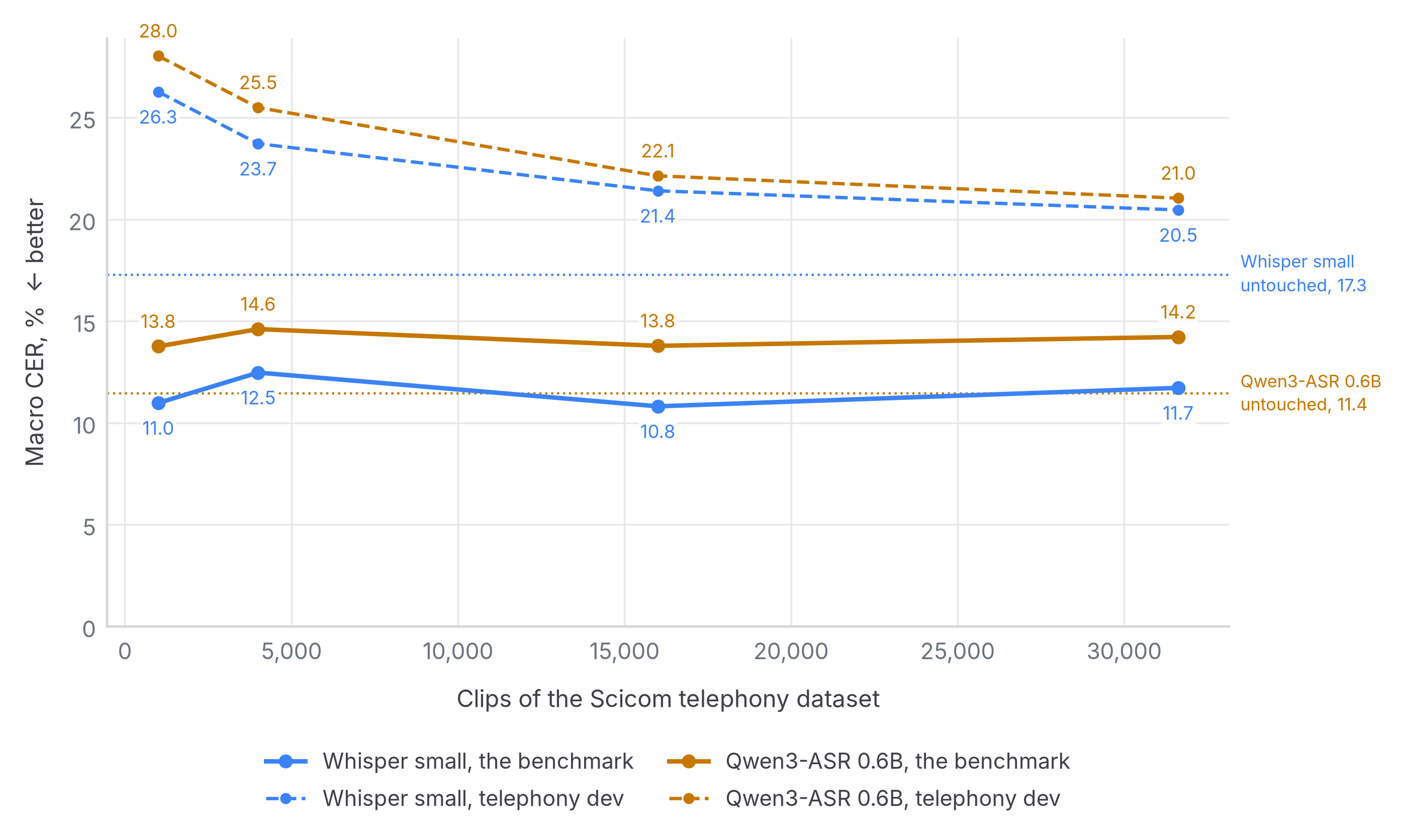}
\caption{More of the Scicom telephony dataset, at the epoch telephony dev chose. Solid: the public benchmark. Dashed: telephony dev. Dotted: the benchmark, the model's better untouched score.}
\label{fig:malaysize}
\end{figure}

\FloatBarrier

\section{Limitations}

\label{sec:limits}

The test set is 25 clips a language: a language's own CER is a rough number, and the macro mean is what we trust. Its intervals resample speakers and take the 50 languages as given; with the languages drawn again, \val{order/both/loose} of the main result's \val{order/pairs} pairs of models could change places (Section~\ref{sec:order}). What a model had heard before is known only from its card or its table, and a language on no list may still have been in a model's training audio. Its labels come from a language detector applied to transcripts, verified by confidence, so a language here means what GlotLID says \citep{kargaran2023glotlid}; our Teso and Karamojong share a corpus labelled Turkana. For \val{data/test-one-corpus} languages every test clip comes from one corpus, and in all \val{data/test-corpus-trains} a test corpus also supplies training clips, from other speakers where it can: the score says how well a model learns a language as that corpus records it, not how it will do on other speech in that language. For \val{data/fleurs-test-langs} languages that corpus is partly or wholly FLEURS, and FLEURS can test other speech in only \val{ooc/n-langs} of the languages it shares with us (Section~\ref{sec:ooc}). \val{data/shared-names} have no other speakers, so their test clips share speakers with training, and every model scores them at \val{an/shared-pct-min} to \val{an/shared-pct-max}\% of its CER on the rest (Figure~\ref{fig:shared}). Of the test clips, \val{data/bad-test-clips} pair a long text with very short audio. Some corpora are readings of the Bible, as Goan Konkani's, Malagasy's and Wayuu's are, and such readings repeat their vocabulary.

The test set's rows were verified one by one, by language and by length; training and dev rows carry the parent corpus's labels, gated on length alone. Dev is harder than test for every model, by \val{an/dev-gap-min} to \val{an/dev-gap-max} points (Figure~\ref{fig:devtest}). Of dev's clips, \val{data/dev-rare}\% come from speakers with at most five clips in the corpus, against \val{data/test-rare}\% of test's, and in the median model \val{an/dev-fail}\% fail outright, with three quarters of their characters wrong or worse, against \val{an/test-fail}\% of test's. So a dev score and a test score here are not comparable, and the test score flatters how these models would do on a language's rarer speakers.

Decoding stops at 40 tokens plus 30 for each second of audio, and Whisper's byte-level vocabulary needs more than that on \val{tok/bod_Tibt/over} of the \val{tok/bod_Tibt/clips} Tibetan test clips and \val{tok/lao_Laoo/over} of the Lao ones (Figure~\ref{fig:tokens}); decoding the chosen epoch again at 60 moves the macro CER by at most \val{rd/gain-max} points on the \val{rd/n} models we tried it on.

\begin{figure}[H]
\centering
\includegraphics[width=0.95\linewidth]{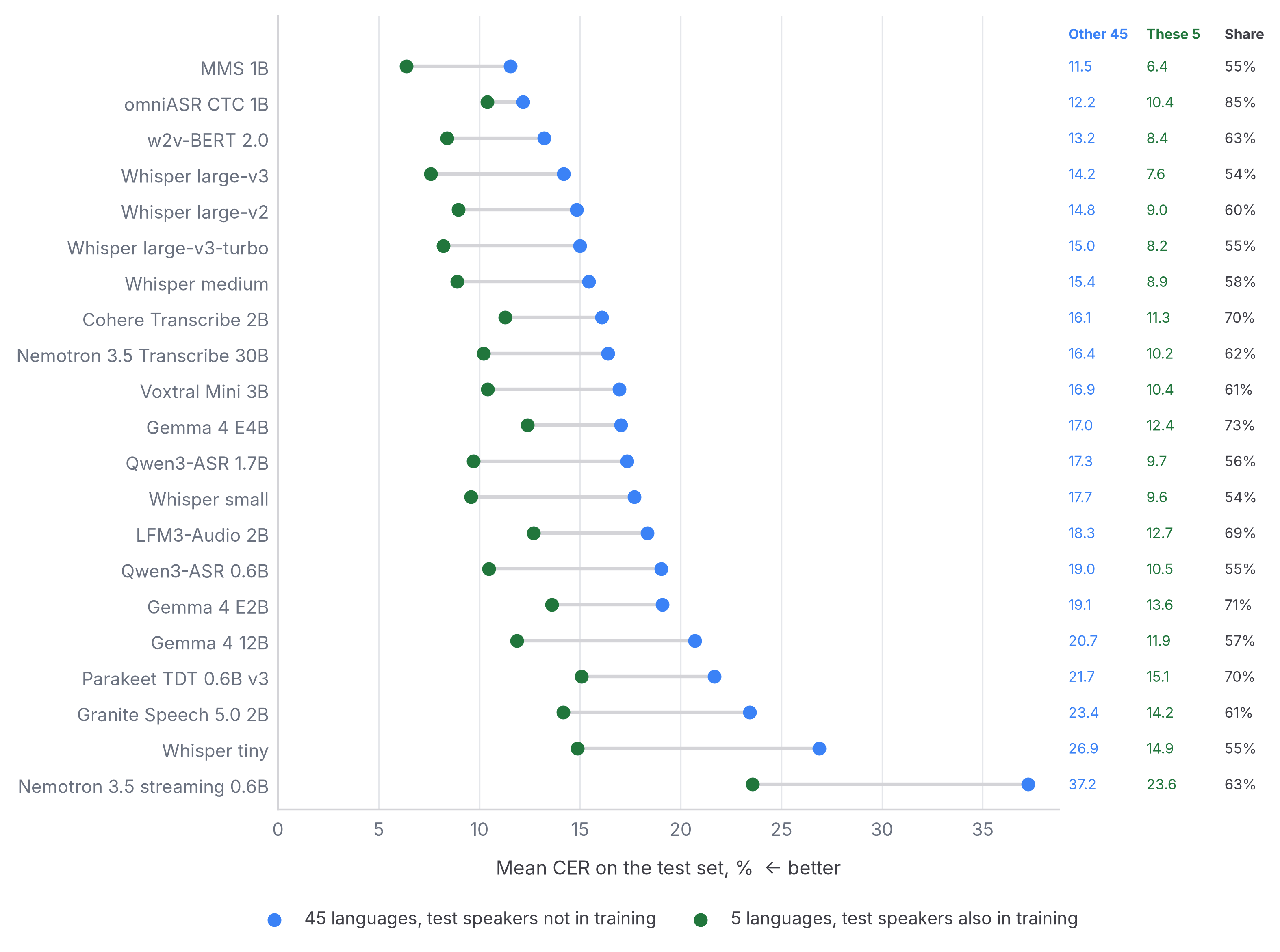}
\caption{Test CER on the \val{data/shared} languages whose test speakers also appear in training, since they have no others, against the rest, each model at the epoch dev chose. Share: the first as a percentage of the second.}
\label{fig:shared}
\end{figure}

\begin{figure}[H]
\centering
\includegraphics[width=0.95\linewidth]{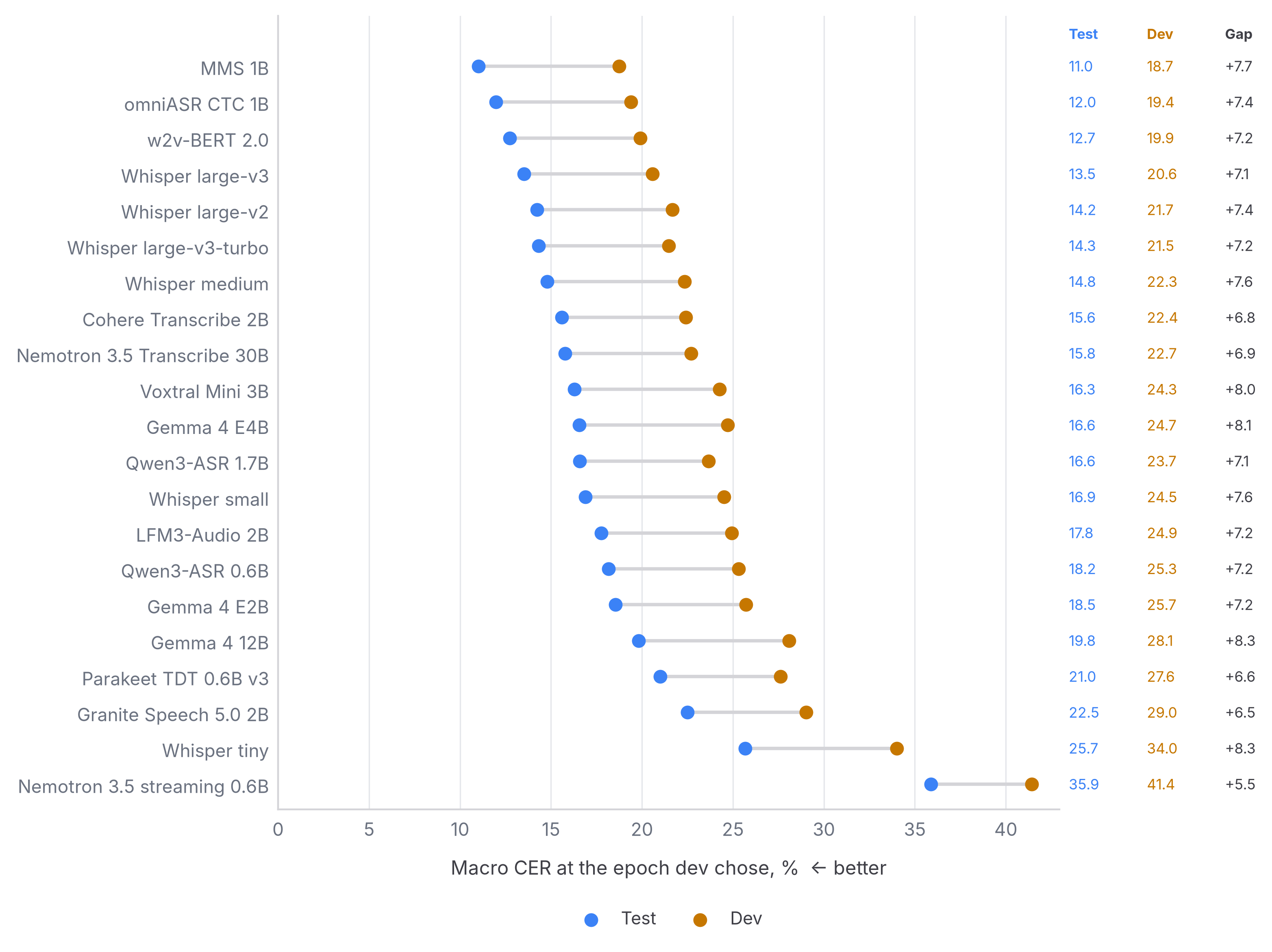}
\caption{Dev and test macro CER of each model at the epoch dev chose.}
\label{fig:devtest}
\end{figure}

We tuned the learning rate per model but no other setting, so a model with a better recipe may rank higher. Granite Speech, MMS and Voxtral were swept late, and the sweep found the rate MMS and Voxtral first trained at too fast (Appendix~\ref{app:lr}). Run again at \val{rerun/mms-1b/main-rate}, MMS went from \val{rerun/mms-1b/first} to \val{rerun/mms-1b/main}, the lowest of any model, and Voxtral at \val{rerun/voxtral-mini/main-rate} from \val{rerun/voxtral-mini/first} to \val{rerun/voxtral-mini/main}. Outside Section~\ref{sec:peft} every score is greedy decoding with no language model. In the Malaysian arm w2v-BERT's new head stalls at its main rate, and we did not find out why; its row there is at a third of that rate, the only row of that arm not at the model's main rate.

\begin{figure}[H]
\centering
\includegraphics[width=0.95\linewidth]{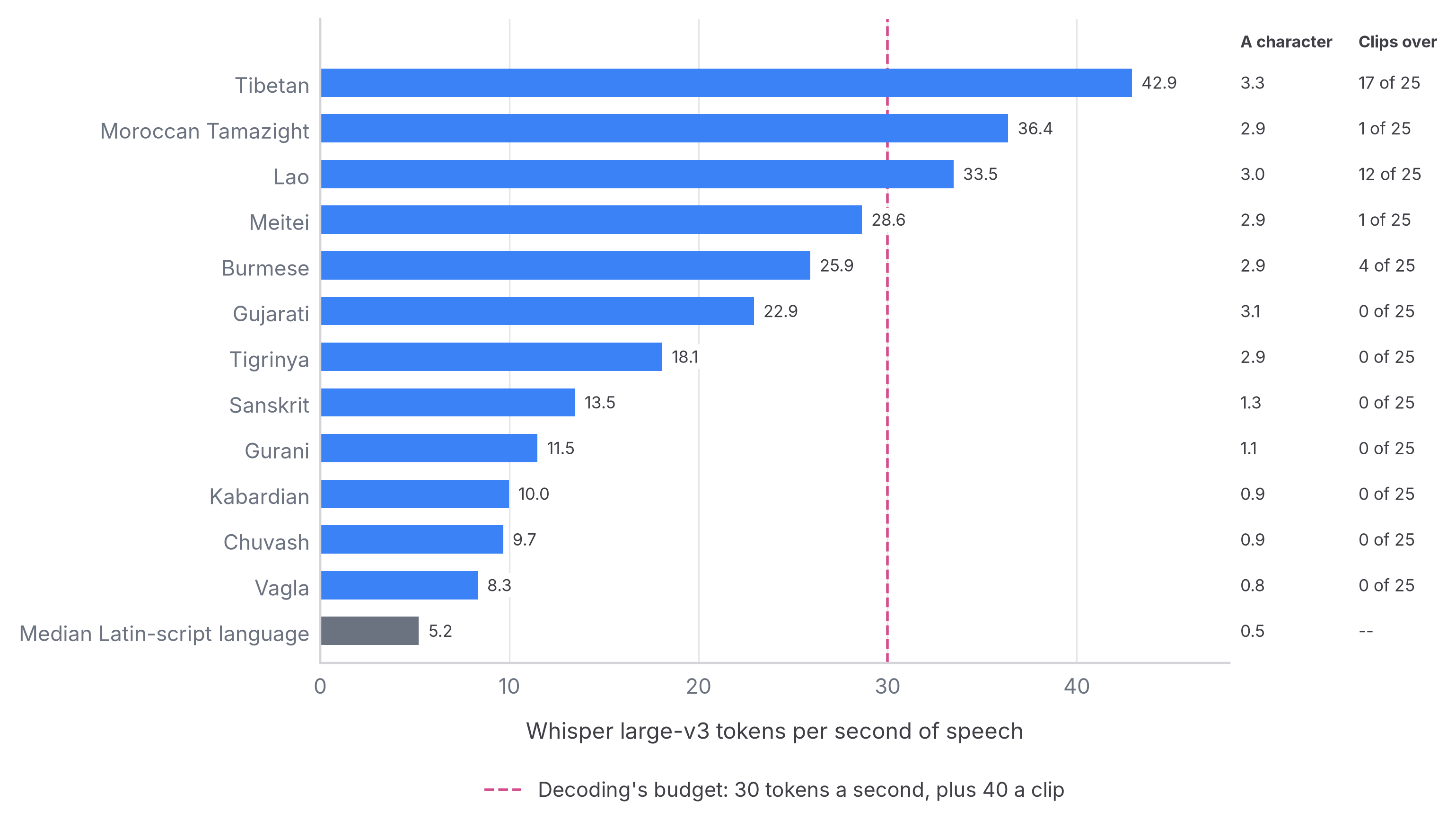}
\caption{Tokens Whisper large-v3 needs per second of speech for each language's test transcripts, as it trains on them, for the twelve languages that need the most and the median Latin-script language. The columns give tokens per character and the test clips whose transcript needs more tokens than decoding allows.}
\label{fig:tokens}
\end{figure}

\section{Conclusion}

\label{sec:conclusion}

A multilingual recogniser fine-tuned on an average of \val{data/k1000-minutes} minutes of a language it has never heard learns to write it. On 50 low-resource languages every model gained, by \val{ft/gain-min} to \val{ft/gain-max} points of macro CER, and the same architectures from random weights barely got started. On the \val{nl/n} that no model lists, every model starts at \val{nl/before-min} or worse and the best ends at \val{nl/best/cer}. What transfers is hearing speech in general: an encoder that never heard these languages already gives a Whisper or Qwen decoder most of what it needs, though not a transducer's, and the script, the thing an untouched model most often gets wrong, is the first thing training puts right. Most of it is the language, not the corpus: on FLEURS, in the languages it shares with ours without sharing recordings, the median model keeps \val{ooc/keep-median}\% of its gain.

With a little of a new language, start from a model with a strong encoder and an output that can spell anything, as large as you can train; a CTC model or a Whisper did best here, and omniASR CTC 7B best of all, at \val{ft/top/cer}. On a seq2seq or language-model decoder, stop after three to five epochs; CTC models and transducers want ten or more. New clips are worth more than further epochs at the same number presented, so spend the time collecting them; where Common Voice has the language, adding it took \val{cv/both-gain} points off. If one model for each language is no burden, train the language alone. On a transducer, check what the new rows write before trusting the loss. On a mixture of experts, train the audio encoder with the rest; the routed experts, most of the weights, were worth under a point, less than letting the rate fall to zero (Section~\ref{sec:nemo-abl}). Before crediting the training, score the untouched model both told the language and left to detect it: told that a clip was Malay, an untouched Whisper translated the English that the speakers mixed in, and lost more to that than training gained. A CTC model should be decoded with a word language model; a few thousand sentences a language took \val{lm/wer-gain-min} to \val{lm/wer-gain-max} points off its WER. Expect to lose some of the languages the model knew, \val{fk/loss-median} points of CER in the median model, and check them before shipping it. Expect a switch into English to go badly: trained on one language a clip, the median model wrote \val{cs/en-after-median}\% of the English words in speech that mixed them into a language with a script of its own. Do not expect a model that has learned forty other low-resource languages to learn the next one faster. Where a full copy of the model for every set of languages is a burden, LoRA on a few percent of the weights does nearly as well, and on \val{fk/lora/less-n} of our \val{fk/lora/n} models it also keeps more of the languages the model knew. Read the output as a draft: even with a language model, the lowest WER here is \val{lm/best/wer}.

We tried one recipe per model, a language model and adapters on three models each, and one seed for most runs, and FLEURS could check our corpus in only \val{ooc/n-langs} languages. The languages we could test are the ones a public corpus happened to record, as it recorded them. What the paper does show is that the first hour or two of a new language is worth far more on top of a multilingual model than on its own, and that it is enough to start.

%% file: table_recipe.tex
\begin{adjustbox}{max width=\linewidth}
\begin{tabular}{lrllll}
\toprule
\schead \scth{Model} & \scth{Rate} & \scth{Added, its rate} & \scth{Language input} & \scth{Batch} & \scth{Its own configuration} \\
\midrule
Whisper tiny & 1e-4 & 33 language tokens & language token & 64 clips & none \\
Whisper base & 1e-4 & 33 language tokens & language token & 64 clips & none \\
Whisper small & 3e-5 & 33 language tokens & language token & 48 clips & none \\
Whisper medium & 1e-5 & 33 language tokens & language token & 32 clips & none \\
Whisper large-v3-turbo & 1e-5 & 33 language tokens & language token & 24 clips & none \\
Whisper large-v2 & 1e-5 & 33 language tokens & language token & 24 clips & none \\
Whisper large-v3 & 1e-5 & 33 language tokens & language token & 24 clips & none \\
Cohere Transcribe 2B & 3e-5 & none (50 unused tags) & language token & 180 s & none \\
Qwen3-ASR 0.6B & 1e-5 & -- & name in prompt & 240 s & none \\
Qwen3-ASR 1.7B & 1e-5 & -- & name in prompt & 180 s & none \\
Granite Speech 5.0 2B & 1e-5 & -- & name in prompt & 90 s & dropout 0.1 \\
Voxtral Mini 3B & 3e-6 & -- & code in prompt & 8 clips & none \\
Nemotron 3.5 Transcribe 30B & 1e-5 & -- & name in instruction & 45 s & dropout 0.1 \\
LFM3-Audio 2B & 1e-5 & -- & name in prompt & 180 s & none \\
Gemma 4 E2B & 1e-5 & -- & name in prompt & 90 s & none \\
Gemma 4 E4B & 3e-6 & -- & name in prompt & 60 s & none \\
Gemma 4 12B & 3e-6 & -- & name in prompt & 60 s & none \\
Parakeet TDT 0.6B v3 & 3e-4 & 746 character rows, 3e-3 & -- & 240 s & dropout 0.1, layer drop 0.1 \\
Nemotron 3.5 streaming 0.6B & 1e-4 & 660 character rows, 1e-3; slots & prompt slot & 240 s & dropout 0.1, layer drop 0.1 \\
w2v-BERT 2.0 & 3e-5 & head of 825 characters + LayerNorm, 3e-5 & -- & 240 s & dropout 0.1, layer drop 0.1 \\
MMS 1B & 1e-5 & head of 825 characters, 1e-3 & -- & 240 s & dropout 0.05, layer drop 0.05, masking 5\% in spans of 10 \\
omniASR CTC 300M & 3e-5 & 75 character rows & -- & 240 s & dropout 0.1, layer drop 0.1 \\
omniASR CTC 1B & 3e-5 & 75 character rows & -- & 240 s & dropout 0.1, layer drop 0.1 \\
omniASR CTC 7B & 1e-5 & 75 character rows & -- & 20 s $\times$ 6 & dropout 0.1, layer drop 0.1 \\
\bottomrule
\end{tabular}
\end{adjustbox}

%% file: table_main.tex
\begin{adjustbox}{max width=\linewidth}
\begin{tabular}{lrrrrrrrr}
\toprule
\schead \scth{Model} & \scth{Parameters} & \scth{Untouched} & \scth{Fine-tuned} & \scth{95\% CI} & \scth{WER} & \scth{Best epoch} & \scth{Rate} & \scth{Hours / epoch} \\
\midrule
MMS 1B & 0.97B & 30.6 & \textbf{11.0} & 10.6--11.6 & 39.3 & 10 of 10 & 1e-5 & 0.52 \\
omniASR CTC 1B & 0.98B & 28.5 & \textbf{12.0} & 11.5--12.6 & 43.1 & 7 of 10 & 3e-5 & 0.51 \\
w2v-BERT 2.0 & 0.58B & -- & \textbf{12.7} & 12.3--13.3 & 44.9 & 9 of 10 & 3e-5 & 0.30 \\
Whisper large-v3 & 1.54B & 50.3 & \textbf{13.5} & 13.0--14.2 & 41.5 & 8 of 10 & 1e-5 & 1.11 \\
Whisper large-v2 & 1.54B & 57.8 & \textbf{14.2} & 13.7--14.9 & 43.2 & 4 of 5 & 1e-5 & 1.09 \\
Whisper large-v3-turbo & 0.81B & 54.6 & \textbf{14.3} & 13.6--15.1 & 43.0 & 9 of 10 & 1e-5 & 0.82 \\
Whisper medium & 0.76B & 63.7 & \textbf{14.8} & 14.2--15.5 & 45.0 & 10 of 10 & 1e-5 & 0.59 \\
Cohere Transcribe 2B & 2.07B & 68.0 & \textbf{15.6} & 15.0--16.3 & 45.8 & 10 of 10 & 3e-5 & 0.34 \\
Nemotron 3.5 Transcribe 30B & 32.36B & 58.1 & \textbf{15.8} & 15.2--16.5 & 45.3 & 2 of 4 & 1e-5 & -- \\
Voxtral Mini 3B & 4.68B & 61.8 & \textbf{16.3} & 15.7--17.1 & 45.7 & 3 of 4 & 3e-6 & 2.93 \\
Gemma 4 E4B & 7.94B & 39.8 & \textbf{16.6} & 16.0--17.3 & 47.1 & 2 of 5 & 3e-6 & 1.90 \\
Qwen3-ASR 1.7B & 2.04B & 59.3 & \textbf{16.6} & 16.0--17.3 & 47.9 & 4 of 10 & 1e-5 & 0.47 \\
Whisper small & 0.24B & 65.7 & \textbf{16.9} & 16.4--17.6 & 49.2 & 9 of 10 & 3e-5 & 0.20 \\
LFM3-Audio 2B & 1.98B & 64.3 & \textbf{17.8} & 17.2--18.6 & 48.7 & 3 of 10 & 1e-5 & 0.44 \\
Qwen3-ASR 0.6B & 0.78B & 60.9 & \textbf{18.2} & 17.5--18.9 & 51.9 & 4 of 10 & 1e-5 & 0.26 \\
Gemma 4 E2B & 5.10B & 43.5 & \textbf{18.5} & 17.8--19.4 & 49.2 & 5 of 5 & 1e-5 & 0.81 \\
Gemma 4 12B & 11.96B & 72.9 & \textbf{19.8} & 19.1--20.6 & 51.8 & 2 of 4 & 3e-6 & 4.03 \\
Parakeet TDT 0.6B v3 & 0.63B & 55.7 & \textbf{21.0} & 20.2--21.9 & 52.6 & 9 of 10 & 3e-4 & 0.80 \\
Granite Speech 5.0 2B & 2.31B & 62.1 & \textbf{22.5} & 21.8--23.4 & 58.1 & 7 of 10 & 1e-5 & 0.68 \\
Whisper tiny & 0.04B & 66.6 & \textbf{25.7} & 24.9--26.5 & 63.9 & 5 of 10 & 1e-4 & 0.04 \\
Nemotron 3.5 streaming 0.6B & 0.64B & 68.7 & \textbf{35.9} & 35.3--36.5 & 78.2 & 10 of 10 & 1e-4 & 0.35 \\
\bottomrule
\end{tabular}
\end{adjustbox}

%% file: table_zero.tex
\begin{adjustbox}{max width=\linewidth}
\begin{tabular}{lrrr}
\toprule
\schead \scth{Model} & \scth{Untouched} & \scth{Told the language, untrained} & \scth{Fine-tuned} \\
\midrule
Whisper large-v3 & 50.3 & 45.1 & 13.5 \\
Whisper large-v3-turbo & 54.6 & 49.1 & 14.3 \\
Whisper large-v2 & 57.8 & 56.4 & 14.2 \\
Whisper medium & 63.7 & 76.4 & 14.8 \\
Whisper small & 65.7 & 61.9 & 16.9 \\
Whisper tiny & 66.6 & 60.9 & 25.7 \\
Qwen3-ASR 1.7B & 59.3 & 50.9 & 16.6 \\
Qwen3-ASR 0.6B & 60.9 & 51.5 & 18.2 \\
Nemotron 3.5 Transcribe 30B & 58.1 & 54.4 & 15.8 \\
Voxtral Mini 3B & 61.8 & 57.3 & 16.3 \\
Granite Speech 5.0 2B & 62.1 & 62.6 & 22.5 \\
Cohere Transcribe 2B & 68.0 & 79.1 & 15.6 \\
LFM3-Audio 2B & 64.3 & 57.5 & 17.8 \\
Gemma 4 E2B & 52.2$^\dagger$ & 43.5 & 18.5 \\
Gemma 4 E4B & 47.7$^\dagger$ & 39.8 & 16.6 \\
Gemma 4 12B & 75.8$^\dagger$ & 72.9 & 19.8 \\
\bottomrule
\end{tabular}
\end{adjustbox}

%% file: table_kin.tex
\begin{adjustbox}{max width=\linewidth}
\begin{tabular}{lrrrrrrrrr}
\toprule
\schead \scth{Model} & \multicolumn{3}{c}{\scth{On its list}} & \multicolumn{3}{c}{\scth{A relative on its list}} & \multicolumn{3}{c}{\scth{No relative}} \\
\schead & \scth{n} & \scth{Untouched} & \scth{Tuned} & \scth{n} & \scth{Untouched} & \scth{Tuned} & \scth{n} & \scth{Untouched} & \scth{Tuned} \\
\midrule
Whisper large-v3 & 11 & 18.1 & 8.5 & 21 & 56.5 & 14.6 & 4 & 55.2 & 10.2 \\
Whisper large-v3-turbo & 11 & 20.3 & 10.8 & 21 & 58.3 & 14.6 & 4 & 61.7 & 10.2 \\
Whisper medium & 11 & 35.0 & 10.5 & 21 & 66.9 & 15.6 & 4 & 74.0 & 10.9 \\
Whisper small & 11 & 35.0 & 13.4 & 21 & 70.2 & 17.5 & 4 & 68.5 & 11.4 \\
Whisper tiny & 11 & 41.3 & 22.9 & 21 & 65.8 & 26.5 & 4 & 68.1 & 17.1 \\
Qwen3-ASR 1.7B & 1 & 8.6 & 15.1 & 23 & 52.0 & 15.3 & 12 & 51.9 & 16.7 \\
Qwen3-ASR 0.6B & 1 & 13.0 & 18.4 & 23 & 51.5 & 16.8 & 12 & 59.2 & 18.2 \\
Parakeet TDT 0.6B v3 & 4 & 5.3 & 5.4 & 10 & 36.0 & 16.2 & 22 & 53.2 & 18.3 \\
Nemotron 3.5 streaming 0.6B & 4 & 9.7 & 20.8 & 18 & 65.3 & 35.0 & 14 & 76.2 & 33.8 \\
Nemotron 3.5 Transcribe 30B & 4 & 3.7 & 4.0 & 20 & 50.7 & 15.0 & 12 & 61.2 & 16.7 \\
\bottomrule
\end{tabular}
\end{adjustbox}

%% file: table_epochs.tex
\begin{adjustbox}{max width=\linewidth}
\begin{tabular}{lrrrrrrrr}
\toprule
\schead \scth{Model} & \scth{Epoch 1} & \scth{Epoch 2} & \scth{Epoch 3} & \scth{Epoch 5} & \scth{Epoch 10} & \scth{Best (dev)} & \scth{Within 1 pt at} & \scth{Last gains on epoch 3 (95\% CI)} \\
\midrule
Whisper large-v3 & 17.1 & 14.8 & 13.8 & 13.6 & 13.4 & 8 & 3 & +0.5 (+0.2 to +0.7) \\
Whisper large-v3-turbo & 21.4 & 16.5 & 15.1 & 14.2 & 14.3 & 9 & 3 & +0.8 (+0.4 to +1.2) \\
Whisper large-v2 & 18.5 & 15.4 & 14.7 & 14.3 & -- & 4 & 3 & +0.4 (-0.0 to +0.8) \\
Whisper medium & 19.2 & 16.2 & 15.6 & 15.2 & 14.8 & 10 & 3 & +0.8 (+0.5 to +1.2) \\
Whisper small & 22.2 & 18.9 & 17.6 & 16.9 & 17.4 & 9 & 3 & +0.2 (-0.3 to +0.6) \\
Whisper tiny & 34.3 & 30.1 & 28.5 & 25.7 & 26.9 & 5 & 5 & +1.6 (+0.8 to +2.3) \\
Qwen3-ASR 1.7B & 19.4 & 18.0 & 17.3 & 16.6 & 17.3 & 4 & 3 & +0.1 (-0.4 to +0.6) \\
Qwen3-ASR 0.6B & 22.3 & 19.4 & 18.7 & 18.2 & 18.3 & 4 & 3 & +0.4 (-0.1 to +0.9) \\
Nemotron 3.5 Transcribe 30B & 16.7 & 15.8 & 16.3 & -- & -- & 2 & 1 & +0.0 (-0.4 to +0.4) \\
Parakeet TDT 0.6B v3 & 35.3 & 30.8 & 31.8 & 24.5 & 20.6 & 9 & 8 & +11.2 (+10.2 to +12.3) \\
Nemotron 3.5 streaming 0.6B & 45.7 & 45.7 & 45.1 & 41.1 & 35.9 & 10 & 10 & +9.2 (+8.6 to +9.9) \\
w2v-BERT 2.0 & 26.6 & 18.7 & 16.5 & 14.6 & 12.7 & 9 & 8 & +3.8 (+3.4 to +4.1) \\
MMS 1B & 16.4 & 14.4 & 14.5 & 12.3 & 11.0 & 10 & 6 & +3.5 (+3.1 to +3.9) \\
omniASR CTC 1B & 16.6 & 15.6 & 14.9 & 14.2 & 11.7 & 7 & 7 & +3.2 (+3.0 to +3.4) \\
Voxtral Mini 3B & 18.1 & 16.7 & 16.3 & -- & -- & 3 & 2 & +0.2 (-0.4 to +0.7) \\
Granite Speech 5.0 2B & 26.9 & 24.4 & 22.8 & 23.1 & 22.5 & 7 & 4 & +0.3 (-0.2 to +1.0) \\
Cohere Transcribe 2B & 22.0 & 17.1 & 16.3 & 15.5 & 15.6 & 10 & 4 & +0.7 (+0.3 to +1.2) \\
LFM3-Audio 2B & 21.2 & 18.2 & 17.8 & 18.1 & 18.2 & 3 & 2 & -0.4 (-0.9 to +0.1) \\
Gemma 4 E2B & 18.9 & 18.1 & 18.4 & 18.5 & -- & 5 & 3 & -0.1 (-0.8 to +0.5) \\
Gemma 4 E4B & 17.4 & 16.6 & 17.4 & 18.3 & -- & 2 & 2 & -0.9 (-1.5 to -0.3) \\
Gemma 4 12B & 22.6 & 19.8 & 21.4 & -- & -- & 2 & 2 & +0.4 (-0.4 to +1.1) \\
\bottomrule
\end{tabular}
\end{adjustbox}

%% file: table_scale.tex
\begin{adjustbox}{max width=\linewidth}
\begin{tabular}{rrrrrrrrr}
\toprule
\schead \scth{Clips a language, at most} & \scth{Clips} & \scth{Hours} & \scth{Epochs} & \scth{Clips presented} & \scth{Qwen3-ASR 0.6B} & \scth{Whisper small} & \scth{Whisper large-v3-turbo} & \scth{Parakeet TDT 0.6B v3} \\
\midrule
25 & 1,209 & 2 & 60 & 72,540 & 39.3 & 38.4 & 32.3 & 45.6 \\
50 & 2,419 & 4 & 40 & 96,760 & 33.4 & 31.7 & 26.7 & 43.2 \\
100 & 4,820 & 8 & 24 & 115,680 & 28.4 & 26.7 & 22.4 & 34.7 \\
200 & 9,542 & 17 & 15 & 143,130 & 24.4 & 22.5 & 19.1 & 32.8 \\
500 & 22,896 & 42 & 10 & 228,960 & 20.4 & 18.7 & 15.9 & 27.3 \\
1,000 & 42,816 & 82 & 10 & 428,160 & 18.2 & 16.9 & 14.3 & 21.7 \\
2,000 & 77,278 & 159 & 5 & 386,390 & 16.1 & 15.4 & 12.7 & 21.4 \\
4,000 & 129,236 & 279 & 3 & 387,708 & 15.3 & 14.4 & 12.8 & 20.7 \\
\bottomrule
\end{tabular}
\end{adjustbox}

%% file: table_adapters.tex
\begin{adjustbox}{max width=\linewidth}
\begin{tabular}{llrrrr}
\toprule
\schead \scth{Model} & \scth{Trained} & \scth{Parameters trained} & \scth{Rate} & \scth{Test CER} & \scth{95\% CI} \\
\midrule
Whisper large-v3 & everything & 1,543.5M (100.0\%) & 1e-5 & 13.5 & 13.0--14.2 \\
Whisper large-v3 & LoRA, rank 32 & 57.7M (3.6\%) & 3e-4 & 14.6 & 14.0--15.4 \\
Qwen3-ASR 0.6B & everything & 782.4M (100.0\%) & 1e-5 & 18.2 & 17.5--18.9 \\
Qwen3-ASR 0.6B & LoRA, rank 32 & 29.5M (3.6\%) & 1e-4 & 18.4 & 17.8--19.2 \\
MMS 1B & everything & 965.7M (100.0\%) & 1e-5 & 11.0 & 10.6--11.6 \\
MMS 1B & LoRA, rank 32 & 36.4M (3.6\%) & 1e-4 & 11.7 & 11.3--12.4 \\
MMS 1B & MMS's adapters, one set for all 50 & 3.2M (0.3\%) & 3e-4 & 14.0 & 13.6--14.6 \\
MMS 1B & MMS's adapters, one set a language & 2.2M (0.2\%) & 3e-4 & 16.8 & 16.4--17.3 \\
\bottomrule
\end{tabular}
\end{adjustbox}

%% file: table_lm.tex
\begin{adjustbox}{max width=\linewidth}
\begin{tabular}{lrrrrrr}
\toprule
\schead \scth{Model} & \scth{CER, greedy} & \scth{CER, LM} & \scth{WER, greedy} & \scth{WER, LM} & \scth{WER gain, 95\% CI} & \scth{Weight, bonus} \\
\midrule
MMS 1B & 11.1 & \textbf{9.8} & 40.1 & \textbf{34.4} & 5.7 (4.8 to 6.4) & 0.25, 2 \\
w2v-BERT 2.0 & 12.7 & \textbf{11.3} & 44.8 & \textbf{38.8} & 6.0 (5.0 to 6.7) & 0.25, 2 \\
omniASR CTC 1B & 12.0 & \textbf{10.8} & 43.1 & \textbf{38.3} & 4.8 (4.1 to 5.4) & 0.25, 1 \\
\bottomrule
\end{tabular}
\end{adjustbox}

%% file: table_malay.tex
\begin{adjustbox}{max width=\linewidth}
\begin{tabular}{lrrrrrrr}
\toprule
\schead \scth{Model} & \scth{Told Malay} & \scth{Left to itself} & \scth{Fine-tuned} & \scth{95\% CI} & \scth{WER} & \scth{Best epoch} & \scth{Telephony} \\
\midrule
Whisper large-v3 & 14.9 & 8.9 & \textbf{6.0} & 5.0--7.2 & 14.9 to 13.3 & 2 & 17.8 to 11.7 \\
Whisper large-v2 & 16.5 & 11.6 & \textbf{6.8} & 6.2--7.6 & 18.7 to 16.2 & 4 & 26.6 to 12.8 \\
Whisper medium & 17.7 & 12.4 & \textbf{6.9} & 6.0--8.0 & 22.0 to 16.1 & 2 & 30.4 to 12.1 \\
Gemma 4 E4B & 8.4 & 10.2 & \textbf{7.6} & 6.8--8.4 & 17.1 to 16.2 & 2 & 20.9 to 13.8 \\
omniASR CTC 7B & -- & 26.7 & \textbf{8.3} & 7.6--9.1 & 43.1 to 24.8 & 3 & 48.5 to 15.5 \\
Whisper large-v3-turbo & 10.2 & 8.4 & \textbf{8.4} & 7.5--9.4 & 14.9 to 19.4 & 3 & 17.3 to 14.1 \\
Qwen3-ASR 1.7B & 10.7 & 8.6 & \textbf{8.7} & 7.9--9.5 & 17.6 to 20.7 & 1 & 7.3 to 11.5 \\
Voxtral Mini 3B & -- & 32.5 & \textbf{9.0} & 8.0--10.3 & 44.4 to 19.6 & 2 & 25.0 to 11.2 \\
w2v-BERT 2.0 & -- & -- & \textbf{9.3} & 8.4--10.3 & 27.9 & 12 & -- \\
Nemotron 3.5 Transcribe 30B & -- & 28.0 & \textbf{11.5} & 10.6--12.5 & 48.7 to 26.7 & 2 & 17.1 to 12.8 \\
Whisper small & 21.0 & 17.3 & \textbf{11.7} & 10.6--13.1 & 30.8 to 27.1 & 3 & 28.0 to 15.4 \\
Cohere Transcribe 2B & -- & 58.5 & \textbf{12.1} & 11.1--13.1 & 82.2 to 29.6 & 3 & 23.1 to 12.7 \\
Gemma 4 E2B & 9.3 & 11.0 & \textbf{12.1} & 11.1--13.3 & 20.0 to 24.3 & 2 & 21.4 to 16.5 \\
MMS 1B & 15.3 & -- & \textbf{12.2} & 11.3--13.1 & 40.2 to 32.3 & 10 & 36.7 to 20.1 \\
omniASR CTC 1B & -- & 21.5 & \textbf{12.5} & 11.5--13.6 & 39.9 to 34.2 & 8 & 36.5 to 19.9 \\
Qwen3-ASR 0.6B & 11.4 & 15.0 & \textbf{14.2} & 13.2--15.4 & 25.2 to 31.4 & 4 & 13.1 to 17.0 \\
omniASR CTC 300M & -- & 28.8 & \textbf{16.1} & 15.0--17.2 & 53.5 to 43.6 & 10 & 46.6 to 24.2 \\
LFM3-Audio 2B & -- & 60.5 & \textbf{18.1} & 16.9--19.4 & 82.0 to 37.9 & 3 & 24.6 to 16.6 \\
Parakeet TDT 0.6B v3 & -- & 41.6 & \textbf{18.9} & 17.9--20.1 & 77.8 to 41.2 & 6 & 15.1 to 16.3 \\
Gemma 4 12B & 57.9 & 61.0 & \textbf{21.4} & 19.9--23.0 & 73.6 to 39.5 & 2 & 40.1 to 19.3 \\
Whisper base & 28.0 & 24.6 & \textbf{24.8} & 23.5--26.1 & 47.8 to 48.7 & 4 & 33.4 to 23.5 \\
Granite Speech 5.0 2B & -- & 43.5 & \textbf{25.3} & 24.0--26.7 & 79.6 to 49.2 & 4 & 21.5 to 15.4 \\
Whisper tiny & 37.7 & 38.5 & \textbf{30.6} & 29.2--32.0 & 65.8 to 56.8 & 4 & 69.5 to 26.1 \\
Nemotron 3.5 streaming 0.6B & -- & 73.3 & \textbf{48.8} & 47.5--50.1 & 88.4 to 77.3 & 10 & 33.5 to 42.9 \\
omniASR LLM 1B & -- & 16.9 & -- & -- & -- & -- & -- \\
Canary 1B v2 & -- & 49.3 & -- & -- & -- & -- & -- \\
Fun-ASR Nano & -- & 52.9 & -- & -- & -- & -- & -- \\
Granite Speech 5.0 470M & -- & 55.9 & -- & -- & -- & -- & -- \\
VibeVoice-ASR & -- & 64.2 & -- & -- & -- & -- & -- \\
\bottomrule
\end{tabular}
\end{adjustbox}

%% file: appendix.tex
\FloatBarrier

\section{Untouched scores}

\label{app:baseline}

\begin{figure}[H]
\centering
\includegraphics[width=0.95\linewidth]{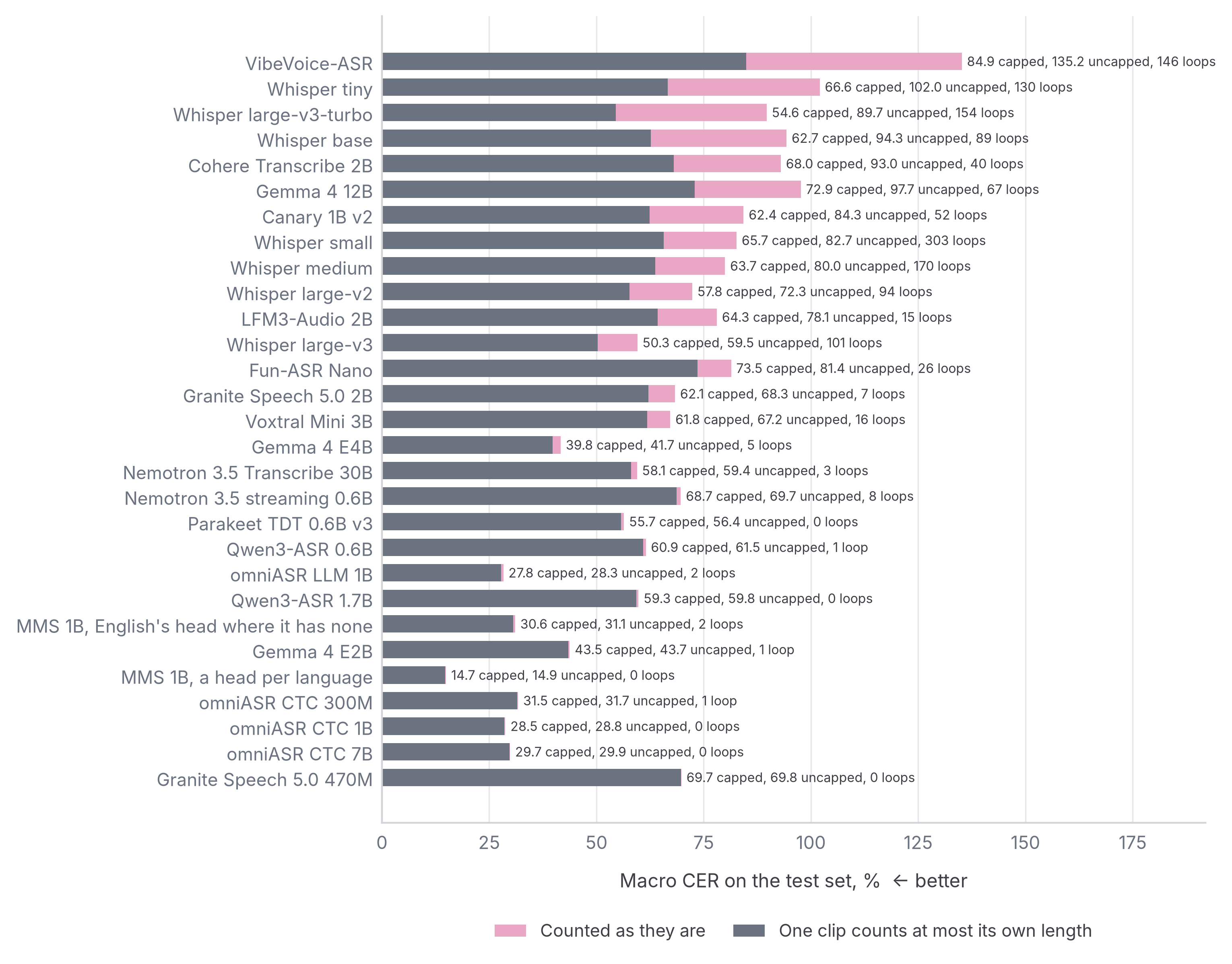}
\caption{Every model as downloaded, with and without the cap on a clip's errors. The pink part is what clips add beyond their own length, which is mostly decoders stuck in a loop.}
\label{fig:loops}
\end{figure}

Table~\ref{tab:baseline} has every model as it was downloaded, and Figure~\ref{fig:loops} what the cap takes away: the part of each model's error that comes from clips writing more than their own length, mostly decoders stuck in a loop.

\begin{table}[H]
\centering
\caption{Every model as downloaded, on the test set. CER is capped per clip; the uncapped column counts every character of a looping decoder. Loops: clips that end by repeating a piece of themselves four times or more.}
\label{tab:baseline}
\scriptsize
\input{table_baseline}
\end{table}

\FloatBarrier

\section{The 50 languages}

\label{app:langs}

Table~\ref{tab:languages} lists the study's languages, ordered by how little the parent corpus holds.

\begin{table}[H]
\centering
\caption{The study's languages. Corpus rows are from the Low-Language-TTS card. Training clips are those left after every rule of Section~\ref{sec:data}, at most 4,000 a language.
Disjoint: no test speaker in training.}
\label{tab:languages}
\scriptsize
\input{table_languages}
\end{table}

\FloatBarrier

\section{Learning rates}

\label{app:lr}

Table~\ref{tab:lr} has the sweep that chose each model's rate.

The rows a transducer adds have a rate of their own, since they start with nothing to keep. Table~\ref{tab:lrr} shows it hardly matters: dev moves by a median of \val{lrr/spread-median} points across three rates for the rows at a given rate for the rest of the model, against more than ten points between the rest of the model's rates in Table~\ref{tab:lr}. The main runs give the rows ten times the rate of the rest of the model, and at least 1e-3.

\begin{table}[H]
\centering
\caption{The learning-rate sweep: dev macro CER on 200 clips a language after three epochs. Bold is the rate the main run used. A sweep this short favours a fast rate, so where the lowest score was not clearly lowest, or the rate was the fastest tried, a full-length run at a lower rate decided on dev; Qwen3-ASR 0.6B, which every ablation starts from, kept the faster of its two tied rates, and Nemotron streaming's lower-rate run slowed only its new rows (Table~\ref{tab:lrr}). Where the lowest score sat at an end of the grid, the sweep tried one rate past that end. Each did worse, except on Parakeet, where 3e-4 beat 1e-4 and 1e-3 did not, so 3e-4 also ran the full length, and dev chose it. Over the full length, 3e-4 silences Nemotron streaming (Section~\ref{sec:start}).\ifval{lrlow/gemma/both}{ Gemma 4 E4B and 12B also ran the next rate down for the full length, and dev chose \val{lrlow/gemma4-e4b/chosen} for E4B (\val{lrlow/gemma4-e4b/dev-chosen} against \val{lrlow/gemma4-e4b/dev-other} at \val{lrlow/gemma4-e4b/other}) and \val{lrlow/gemma4-12b/chosen} for the 12B (\val{lrlow/gemma4-12b/dev-chosen} against \val{lrlow/gemma4-12b/dev-other} at \val{lrlow/gemma4-12b/other}).}{} w2v-BERT at 1e-4 stayed at \val{lrrep/w2v-bert/1e-4}, writing the same output for every clip, when run a second time too. omniASR CTC 7B takes the same 120 seconds of audio per update as the rest, in pieces of 40 seconds in the sweep and 20 in its full run, because larger pieces did not fit. Dashes are rates not run.}
\label{tab:lr}
\scriptsize
\input{table_lr}
\end{table}

\begin{table}[H]
\centering
\caption{The transducers' new rows, on the sweep of Table~\ref{tab:lr}: dev macro CER, by the rate of the pretrained weights (Rate) and of the new rows.}
\label{tab:lrr}
\scriptsize
\input{table_lrr}
\end{table}

%% file: table_baseline.tex
\begin{adjustbox}{max width=\linewidth}
\begin{tabular}{llrrrrrr}
\toprule
\schead \scth{Model} & \scth{Decoder} & \scth{Languages} & \scth{CER} & \scth{95\% CI} & \scth{CER, uncapped} & \scth{WER} & \scth{Loops} \\
\midrule
MMS 1B, a head per language & CTC & 26 & 14.7 & 14.0--15.7 & 14.9 & 42.7 & 0 \\
omniASR LLM 1B & language model & 50 & 27.8 & 26.9--28.8 & 28.3 & 56.4 & 2 \\
omniASR CTC 1B & CTC & 50 & 28.5 & 27.8--29.6 & 28.8 & 63.1 & 0 \\
omniASR CTC 7B & CTC & 50 & 29.7 & 28.9--30.7 & 29.9 & 59.2 & 0 \\
MMS 1B, English's head where it has none & CTC & 50 & 30.6 & 30.1--31.2 & 31.1 & 62.9 & 2 \\
omniASR CTC 300M & CTC & 50 & 31.5 & 30.8--32.4 & 31.7 & 69.5 & 1 \\
Gemma 4 E4B & language model & 50 & 39.8 & 39.1--40.7 & 41.7 & 77.0 & 5 \\
Gemma 4 E2B & language model & 50 & 43.5 & 42.7--44.4 & 43.7 & 80.2 & 1 \\
Whisper large-v3 & seq2seq & 50 & 50.3 & 49.5--51.4 & 59.5 & 83.7 & 101 \\
Whisper large-v3-turbo & seq2seq & 50 & 54.6 & 53.6--55.8 & 89.7 & 84.8 & 154 \\
Parakeet TDT 0.6B v3 & transducer & 50 & 55.7 & 55.0--56.7 & 56.4 & 88.0 & 0 \\
Whisper large-v2 & seq2seq & 50 & 57.8 & 56.7--58.9 & 72.3 & 86.8 & 94 \\
Nemotron 3.5 Transcribe 30B & language model & 50 & 58.1 & 57.2--59.3 & 59.4 & 86.3 & 3 \\
Qwen3-ASR 1.7B & language model & 50 & 59.3 & 58.5--60.2 & 59.8 & 92.9 & 0 \\
Qwen3-ASR 0.6B & language model & 50 & 60.9 & 60.1--61.9 & 61.5 & 94.2 & 1 \\
Voxtral Mini 3B & language model & 50 & 61.8 & 60.8--62.9 & 67.2 & 91.2 & 16 \\
Granite Speech 5.0 2B & language model & 50 & 62.1 & 61.4--63.0 & 68.3 & 95.9 & 7 \\
Canary 1B v2 & seq2seq & 50 & 62.4 & 61.5--63.4 & 84.3 & 90.4 & 52 \\
Whisper base & seq2seq & 50 & 62.7 & 61.8--63.8 & 94.3 & 94.5 & 89 \\
Whisper medium & seq2seq & 50 & 63.7 & 62.7--64.8 & 80.0 & 89.3 & 170 \\
LFM3-Audio 2B & language model & 50 & 64.3 & 63.5--65.2 & 78.1 & 89.5 & 15 \\
Whisper small & seq2seq & 50 & 65.7 & 64.7--66.7 & 82.7 & 91.5 & 303 \\
Whisper tiny & seq2seq & 50 & 66.6 & 65.6--67.6 & 102.0 & 96.1 & 130 \\
Cohere Transcribe 2B & seq2seq & 50 & 68.0 & 67.0--69.1 & 93.0 & 95.8 & 40 \\
Nemotron 3.5 streaming 0.6B & transducer & 50 & 68.7 & 68.0--69.5 & 69.7 & 92.2 & 8 \\
Granite Speech 5.0 470M & CTC & 50 & 69.7 & 68.8--70.5 & 69.8 & 95.8 & 0 \\
Gemma 4 12B & language model & 50 & 72.9 & 71.9--73.9 & 97.7 & 96.4 & 67 \\
Fun-ASR Nano & language model & 50 & 73.5 & 72.7--74.6 & 81.4 & 97.7 & 26 \\
VibeVoice-ASR & language model & 50 & 84.9 & 84.1--85.9 & 135.2 & 98.3 & 146 \\
\bottomrule
\end{tabular}
\end{adjustbox}

%% file: table_languages.tex
\begin{adjustbox}{max width=\linewidth}
\begin{tabular}{lllrrrrrl}
\toprule
\schead \scth{Code} & \scth{Language} & \scth{Family} & \scth{Corpus rows} & \scth{Train clips} & \scth{Train h} & \scth{Train speakers} & \scth{Test speakers} & \scth{Disjoint} \\
\midrule
gom\_Latn & Goan Konkani & Indo-European & 262 & 212 & 0.6 & 1 & 1 & no \\
njm\_Latn & Angami Naga & Sino-Tibetan & 443 & 96 & 0.1 & 30 & 12 & yes \\
guc\_Latn & Wayuu & Arawakan & 736 & 649 & 0.7 & 48 & 10 & no \\
nre\_Latn & Southern Rengma Naga & Sino-Tibetan & 992 & 310 & 0.4 & 70 & 13 & yes \\
mni\_Mtei & Meitei & Sino-Tibetan & 1,098 & 991 & 1.8 & 4 & 4 & no \\
nri\_Latn & Chokri Naga & Sino-Tibetan & 1,246 & 410 & 0.5 & 79 & 16 & yes \\
njo\_Latn & Ao Naga & Sino-Tibetan & 1,282 & 440 & 0.7 & 80 & 12 & yes \\
zgh\_Tfng & Moroccan Tamazight & Afro-Asiatic & 1,283 & 424 & 0.3 & 95 & 13 & yes \\
bci\_Latn & Baoule & Atlantic-Congo & 1,317 & 1,031 & 1.8 & 53 & 11 & yes \\
qup\_Latn & Southern Pastaza Quechua & Quechuan & 1,529 & 719 & 0.9 & 118 & 19 & yes \\
lad\_Latn & Ladino & Indo-European & 1,635 & 204 & 0.2 & 73 & 10 & yes \\
mjw\_Latn & Karbi & Sino-Tibetan & 1,644 & 605 & 0.7 & 208 & 15 & yes \\
gos\_Latn & Gronings & Indo-European & 1,895 & 1,690 & 3.1 & 1,646 & 19 & yes \\
iba\_Latn & Iban & Austronesian & 2,942 & 2,006 & 5.6 & 36 & 8 & yes \\
teo\_Latn & Teso & Nilotic & 3,195 & 1,186 & 1.0 & 90 & 9 & yes \\
ekk\_Latn & Estonian & Uralic & 3,774 & 1,790 & 5.2 & 4 & 3 & yes \\
lvs\_Latn & Latvian & Indo-European & 3,847 & 1,266 & 3.9 & 5 & 4 & yes \\
tgk\_Cyrl & Tajik & Indo-European & 4,349 & 3,165 & 11.2 & 94 & 10 & yes \\
hac\_Arab & Gurani & Indo-European & 4,448 & 2,272 & 1.8 & 1 & 2 & no \\
oci\_Latn & Occitan & Indo-European & 4,800 & 2,696 & 11.0 & 15 & 7 & yes \\
kdj\_Latn & Karamojong & Nilotic & 5,230 & 1,737 & 1.6 & 94 & 15 & yes \\
trp\_Latn & Kok Borok & Sino-Tibetan & 5,384 & 2,435 & 3.2 & 701 & 19 & yes \\
bre\_Latn & Breton & Indo-European & 5,790 & 3,967 & 2.0 & 2,822 & 20 & yes \\
crh\_Latn & Crimean Tatar & Turkic & 6,784 & 3,237 & 4.6 & 311 & 15 & yes \\
slv\_Latn & Slovenian & Indo-European & 9,769 & 1,384 & 4.2 & 6 & 7 & yes \\
fon\_Latn & Fon & Atlantic-Congo & 11,201 & 3,997 & 11.7 & 923 & 7 & yes \\
san\_Deva & Sanskrit & Indo-European & 12,197 & 3,664 & 10.3 & 80 & 10 & yes \\
suk\_Latn & Sukuma & Atlantic-Congo & 12,330 & 3,453 & 9.9 & 36 & 9 & yes \\
hat\_Latn & Haitian Creole & Indo-European & 12,988 & 3,994 & 10.8 & 3,994 & 25 & yes \\
lin\_Latn & Lingala & Atlantic-Congo & 13,887 & 4,000 & 16.6 & 8 & 11 & yes \\
crh\_Cyrl & Crimean Tatar & Turkic & 14,219 & 3,930 & 4.8 & 1 & 1 & no \\
kha\_Latn & Khasi & Austroasiatic & 14,234 & 1,865 & 4.9 & 21 & 8 & yes \\
smj\_Latn & Lule Sami & Uralic & 14,715 & 3,999 & 4.6 & 503 & 11 & yes \\
sna\_Latn & Shona & Atlantic-Congo & 15,476 & 3,946 & 19.9 & 27 & 15 & yes \\
kbd\_Cyrl & Kabardian & Abkhaz-Adyge & 17,393 & 3,348 & 5.3 & 124 & 10 & yes \\
sdh\_Arab & Southern Kurdish & Indo-European & 17,998 & 3,999 & 6.6 & 71 & 9 & yes \\
fin\_Latn & Finnish & Uralic & 20,414 & 4,000 & 6.8 & 45 & 1 & yes \\
plt\_Latn & Malagasy & Austronesian & 21,705 & 3,997 & 11.7 & 97 & 12 & yes \\
lat\_Latn & Latin & Indo-European & 22,862 & 3,916 & 11.1 & 190 & 1 & yes \\
mya\_Mymr & Burmese & Sino-Tibetan & 24,226 & 3,998 & 6.2 & 275 & 10 & yes \\
chv\_Cyrl & Chuvash & Turkic & 29,796 & 3,998 & 4.5 & 508 & 14 & yes \\
guj\_Gujr & Gujarati & Indo-European & 32,373 & 4,000 & 6.9 & 220 & 11 & yes \\
fry\_Latn & West Frisian & Indo-European & 35,956 & 4,000 & 5.2 & 2,526 & 8 & yes \\
lao\_Laoo & Lao & Tai-Kadai & 38,586 & 3,986 & 13.2 & 124 & 9 & yes \\
fuv\_Latn & Nigerian Fulfulde & Atlantic-Congo & 39,284 & 3,417 & 11.2 & 823 & 12 & yes \\
vag\_Latn & Vagla & Atlantic-Congo & 39,303 & 3,989 & 2.3 & 2,521 & 12 & yes \\
bod\_Tibt & Tibetan & Sino-Tibetan & 48,956 & 3,994 & 2.5 & 3,994 & 12 & yes \\
gsw\_Latn & Swiss German & Indo-European & 51,710 & 3,000 & 5.7 & 3 & 1 & yes \\
tir\_Ethi & Tigrinya & Afro-Asiatic & 53,835 & 3,992 & 16.7 & 583 & 12 & yes \\
xsm\_Latn & Kasem & Atlantic-Congo & 58,196 & 3,832 & 2.4 & 2,883 & 13 & yes \\
\bottomrule
\end{tabular}
\end{adjustbox}

%% file: table_lr.tex
\begin{adjustbox}{max width=\linewidth}
\begin{tabular}{lrrrrrrr}
\toprule
\schead \scth{Model} & \scth{1e-6} & \scth{3e-6} & \scth{1e-5} & \scth{3e-5} & \scth{1e-4} & \scth{3e-4} & \scth{1e-3} \\
\midrule
Cohere Transcribe 2B & -- & 49.9 & 42.0 & \textbf{37.0} & 38.3 & -- & -- \\
Gemma 4 12B & 36.3 & \textbf{35.2} & 41.4 & 100.0 & -- & -- & -- \\
Gemma 4 E2B & -- & 32.2 & \textbf{31.5} & 35.1 & 45.9 & -- & -- \\
Gemma 4 E4B & 30.5 & \textbf{30.2} & 31.2 & 36.4 & -- & -- & -- \\
Granite Speech 5.0 2B & -- & 43.3 & \textbf{40.4} & 42.6 & -- & -- & -- \\
LFM3-Audio 2B & -- & 33.0 & \textbf{32.0} & 33.0 & 41.5 & -- & -- \\
MMS 1B & -- & 24.9 & \textbf{23.5} & 23.8 & 29.0 & -- & -- \\
Nemotron 3.5 streaming 0.6B & -- & 80.3 & 83.1 & 73.1 & \textbf{69.0} & -- & -- \\
omniASR CTC 1B & -- & -- & 30.7 & \textbf{24.4} & 29.2 & -- & -- \\
omniASR CTC 300M & -- & -- & 36.5 & \textbf{28.1} & 31.9 & 49.6 & -- \\
omniASR CTC 7B & -- & 29.0 & \textbf{20.8} & 24.5 & 34.3 & -- & -- \\
Parakeet TDT 0.6B v3 & -- & 58.9 & 53.2 & 50.5 & 47.0 & \textbf{43.6} & 47.8 \\
Qwen3-ASR 0.6B & 39.2 & 34.8 & \textbf{34.8} & 36.7 & 43.2 & -- & -- \\
Qwen3-ASR 1.7B & 36.2 & 30.3 & \textbf{31.4} & 32.2 & -- & -- & -- \\
Voxtral Mini 3B & 30.8 & \textbf{29.5} & 32.9 & 40.1 & -- & -- & -- \\
w2v-BERT 2.0 & -- & -- & 60.3 & \textbf{40.8} & 100.0 & -- & -- \\
Whisper base & -- & -- & -- & 43.4 & \textbf{39.3} & 49.1 & -- \\
Whisper large-v3-turbo & -- & 34.9 & \textbf{32.1} & 35.8 & -- & -- & -- \\
Whisper large-v3 & -- & 30.8 & \textbf{29.3} & 29.6 & -- & -- & -- \\
Whisper medium & -- & -- & \textbf{33.0} & 32.7 & 37.8 & -- & -- \\
Whisper small & -- & -- & 40.3 & \textbf{35.0} & 37.8 & -- & -- \\
Whisper tiny & -- & -- & 53.1 & 48.3 & \textbf{47.1} & 54.6 & -- \\
\bottomrule
\end{tabular}
\end{adjustbox}

%% file: table_lrr.tex
\begin{adjustbox}{max width=\linewidth}
\begin{tabular}{llrrr}
\toprule
\schead \scth{Model} & \scth{Rate} & \scth{Rows 1e-4} & \scth{Rows 3e-4} & \scth{Rows 1e-3} \\
\midrule
Nemotron 3.5 streaming 0.6B & 1e-5 & 83.2 & 83.3 & 83.1 \\
Nemotron 3.5 streaming 0.6B & 3e-5 & 76.7 & 75.8 & 73.1 \\
Nemotron 3.5 streaming 0.6B & 1e-4 & 71.3 & 72.1 & 69.0 \\
Parakeet TDT 0.6B v3 & 3e-5 & 50.3 & 51.2 & 50.5 \\
Parakeet TDT 0.6B v3 & 1e-4 & 46.9 & 47.6 & 47.0 \\
\bottomrule
\end{tabular}
\end{adjustbox}